\documentclass[3p]{elsarticle}

\usepackage[utf8]{inputenc}
\usepackage[T1]{fontenc}
\usepackage{url}
\usepackage{booktabs}
\usepackage{graphicx}
\usepackage{amsmath,amssymb,amsfonts,bbm}
\usepackage{natbib}
\biboptions{numbers,sort&compress} 
\usepackage{bm}
\usepackage{appendix}
\usepackage[maxfloats=256]{morefloats}
 
\DeclareMathOperator{\E}{\mathbb{E}}

\journal{Journal of Biomedical Informatics}

\begin{document}

\begin{frontmatter}

\title{An Empirical Characterization of Fair Machine Learning For Clinical Risk Prediction}

\author[1]{Stephen R. Pfohl\corref{cor1}}
\ead{spfohl@stanford.edu}

\author[1,2]{Agata Foryciarz}
\ead{agataf@stanford.edu}

\author[1]{Nigam H. Shah}
\ead{nigam@stanford.edu}

\cortext[cor1]{Corresponding author}

\address[1]{Stanford Center for Biomedical Informatics Research, Stanford University, 1265 Welch Road, Stanford, CA 94305}
\address[2]{Computer Science Department, Stanford University, 353 Jane Stanford Way, Stanford, CA 94305}

\begin{abstract}
The use of machine learning to guide clinical decision making has the potential to worsen existing health disparities. 
Several recent works frame the problem as that of algorithmic fairness, a framework that has attracted considerable attention and criticism.
However, the appropriateness of this framework is unclear due to both ethical as well as technical considerations, the latter of which include trade-offs between measures of fairness and model performance that are not well-understood for predictive models of clinical outcomes.
To inform the ongoing debate, we conduct an empirical study to characterize the impact of penalizing group fairness violations on an array of measures of model performance and group fairness. We repeat the analysis across multiple observational healthcare databases, clinical outcomes, and sensitive attributes.
We find that procedures that penalize differences between the distributions of predictions across groups induce nearly-universal degradation of multiple performance metrics within groups.
On examining the secondary impact of these procedures, we observe heterogeneity of the effect of these procedures on measures of fairness in calibration and ranking across experimental conditions.
Beyond the reported trade-offs, we emphasize that analyses of algorithmic fairness in healthcare lack the contextual grounding and causal awareness necessary to reason about the mechanisms that lead to health disparities, as well as about the potential of algorithmic fairness methods to counteract those mechanisms.
In light of these limitations, we encourage researchers building predictive models for clinical use to step outside the algorithmic fairness frame and engage critically with the broader sociotechnical context surrounding the use of machine learning in healthcare.
\end{abstract}




\end{frontmatter}

\section{Introduction}
The use of machine learning with observational health data to guide clinical decision making has the potential to introduce and exacerbate health disparities for disadvantaged and underrepresented populations \citep{Rajkomar2018,Goodman2018,Obermeyer2019,Ferryman2018,Nordling2019,Vyas2020}.
This effect can derive from inequity in historical and current patterns of care access and delivery \citep{Rajkomar2018,Chen2020,Gaskin2012,Williams2001,Hall2015,Bailey2017}, underrepresentation in clinical datasets \cite{Larrazabal2020}, the use of biased or mis-specified proxy outcomes during model development \citep{Obermeyer2019,Kallus2018,Jiang2019}, and differences in the accessibility, usability, and effectiveness of predictive models across groups \citep{Veinot2018,Rajkomar2018}.
In response, considerable attention has been devoted to reasoning about the extent to which clinical predictive models may be designed to anticipate and proactively mitigate harms to advance health equity, while upholding ethical standards \cite{Rajkomar2018,Goodman2018,Chen2020,Ferryman2018,McCradden2020,McCradden2020b,Char2018,Parikh2019}.

The role that algorithmic fairness techniques should have in the development of clinical predictive models is actively debated \cite{Rajkomar2018,Goodman2018,McCradden2020,McCradden2020a,McCradden2020b}. While these techniques have been extensively studied and scrutinized in domains such as criminal justice, hiring, and education \cite{Dwork2011,Hardt2016,Chouldechova2018,Green2020,Hutchinson2019}, and have been made accessible by several open source software frameworks \cite{aif360-oct-2018,Dudik2020,google_fairness}, the scope of their examination in the context of clinical predictive models has been relatively limited \cite{Pfohl2019MLHC,Pfohl2019AIES,Zink2019,Zhang2020,Singh2019a,Singh2019b}.
Algorithmic fairness methods specify a mathematical formalization of a fairness criterion representative of an ideal (such as equal error rates between male and female patients), and provide procedures for minimizing violations from the fairness criterion without unduly deteriorating model performance \cite{Hardt2016,Zemel2013,Cotter2018,Cotter2019,Agarwal2018,Song2019}.
Furthermore, these mathematical formalizations can be used as auditing mechanisms to identify problematic characteristics of a predictive model and to promote transparency in the model's output \cite{Mitchell2019,Sun2019,Madaio2020,ChenJohanssonSontag_NIPS18}. 
In the context of this debate, it is essential to recognize that algorithmic fairness techniques enable monitoring and manipulating the output of predictive models, but are generally insufficient by themselves to mitigate the introduction or perpetuation of health disparities resulting from model-guided interventions \cite{Goodman2018,McCradden2020,McCradden2020a,McCradden2020b,Corbett-Davies2018,Fazelpour2020,Herington2020,Green2020,Liu2018,Hanna2020,Obermeyer2019,Jacobs2019}. 

Health disparities arise as a result of structural forms of racism and related inequities in areas such as housing, education, employment, and criminal justice that affect healthcare access, utilization, and quality \cite{Bailey2017,Hicken2018}. These effects are further compounded by under-representation of the elderly, women, and ethnic minorities in clinical trials and cohort studies \cite{vitale2017under, hussain2004ethnic}, as well as warped financial incentives in the healthcare system \cite{dickman2017inequality}.
Typical formulations of algorithmic fairness, particularly \textit{group fairness criteria}, are unaware of this context, because they are primarily defined in terms of a model's predictions, observed outcomes, as well as membership in a pre-specified group of demographic attributes, and evaluated on cohorts derived retrospectively.
As a result, algorithmic fairness criteria may be misleading in several situations, including when the observed outcome is a biased surrogate for the construct of interest \cite{Obermeyer2019,Jacobs2019} and when predictive models are not appropriately contextualized in terms of the heterogeneous and dynamic impact of the complex interventions and policies that they enable, contributing to an erroneous conflation of model performance metrics with an accrual of benefit \cite{Shah2019,Jung2020,Corbett-Davies2018,Liu2018,Creager2019}.

Some researchers argue that techniques from algorithmic fairness still have a role to play in promoting health equity \cite{Rajkomar2018}, particularly if their use is informed by a deeply embedded understanding of the sociotechnical and clinical contexts that surround the use of a predictive model's output.
From this perspective, it is relevant to reason about the trade-offs between properties such as model performance and fairness criteria satisfaction, as long as those properties can be appropriately contextualized.
However, it is generally impossible to satisfy conflicting notions of fairness simultaneously \cite{Kleinberg2016,Chouldechova2017,Corbett-Davies2018,Binns2020} and methods that impose fairness constraints are known to do so at the cost of substantial trade-offs between various measures of model performance and fairness criteria satisfaction, in ways that are sensitive to the properties of the dataset and learning algorithm used \citep{Kleinberg2016,Chouldechova2017,Friedler2016,Kearns2017,Corbett-Davies2018,Khani2019,ChenJohanssonSontag_NIPS18,Friedler2019,Lipton2018}.
As the evidence base that surrounds the use of algorithmic fairness methods in the context of clinical risk prediction remains limited, the extent to which these trade-offs manifest when adopting algorithmic fairness approaches for training clinical predictive models remains unclear.

To inform this discussion, we conduct a large-scale empirical study characterizing the trade-offs between various measures of model performance and fairness criteria satisfaction for clinical predictive models that are penalized to varying degrees against violations of group fairness criteria.
We repeat the analysis across several databases, outcomes and sensitive attributes in an attempt to identify patterns that generalize. We make all code relevant for reproducing these experiments available at \url{https://github.com/som-shahlab/fairness_benchmark}.

The structure of the remainder of the paper is as follows. 
In Section \ref{sec:methods}, we formally define the prediction problem, group fairness criteria, metrics, and learning objectives that we evaluate. We then describe the experimental setup, including datasets used, definitions of cohorts, outcomes and sensitive attributes, as well as procedures for feature extraction, model training and evaluation.
In Section \ref{sec:results}, we report on the results.
In Section \ref{sec:discussion}, we provide recommendations for model developers and policy makers in light of both the trade-offs that we find as well as the limitations inherent to the algorithmic fairness framework. Furthermore, we discuss the appropriateness of typical abstractions used for evaluation of group fairness, including the use of discrete categories for sex, race, and ethnicity.

\section{Methods} \label{sec:methods}
\begin{figure}[!ht]
	\centering
	\includegraphics[width=0.95\linewidth]{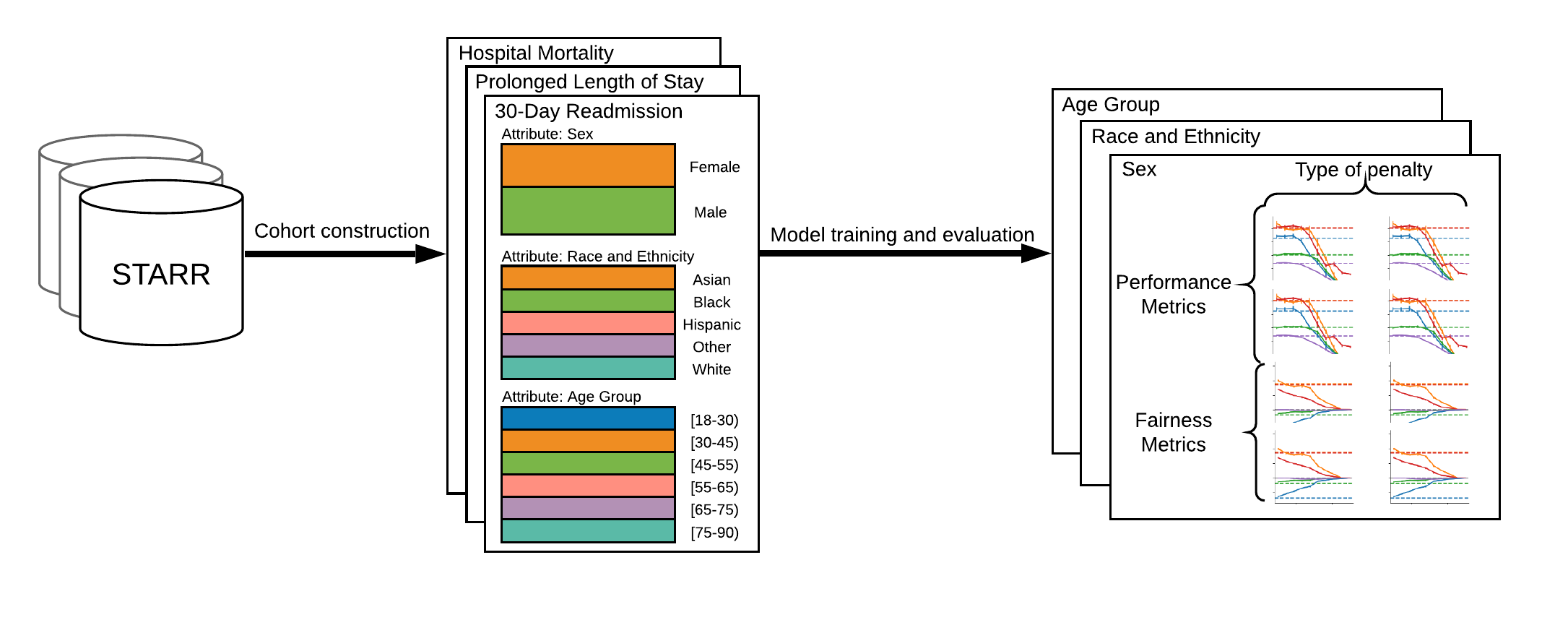}
	\caption{
		An overview of the experimental procedure. We extract cohorts from a collection of databases and label patients on the basis of observed clinical outcomes and membership in groups of multiple demographic attributes. For twenty five combinations of database, outcome, and sensitive attribute, we train a series of predictive models that are penalized to varying degrees against violations of conditional prediction parity. We report on the effect of such penalties on measures of model performance and fairness criteria satisfaction. Shown here is process for the cohort drawn from the STARR database.
	}
	\label{fig:graphical_abstract}
\end{figure}

Across twenty five combinations of datasets, clinical outcomes, and sensitive attributes, we train a series of predictive models that are penalized to varying degrees against violations of fairness criteria, and report on model performance and group fairness metrics. Figure \ref{fig:graphical_abstract} outlines the experimental procedure.

We first define the mathematical formulation of group fairness from which fairness metrics and training objectives are derived (Section \ref{subsec:form_fairness}).
We present three classes of group fairness criteria -- conditional prediction parity (Section \ref{subsubsec:cond_pred_parity}), calibration (Section \ref{subsubsec:calibration}), and cross-group ranking (Section \ref{subsubsec:cross_group_ranking}) -- each of which is operationalized by one or more fairness metrics that quantify the extent to which the associated criterion is violated.
We evaluate each of these metrics for models developed with six regularization strategies that penalize violation of conditional prediction parity (Section \ref{subsec:optimization}). The computational experiments that examine the behavior of these constructs are described in Section \ref{subsec:experiments}, which includes a description of the datasets (Section \ref{subsec:datasets}), definitions of the cohort inclusion criteria, outcomes, and sensitive attributes (Section \ref{subsec:cohort_attributes}), as well as details of feature extraction (Section \ref{subsec:features}), model training (Section \ref{subsec:training}), and evaluation (Section \ref{subsec:model_evaluation}).

\subsection{Formulations of Group Fairness Criteria} \label{subsec:form_fairness}

\subsubsection{Notation and Problem Formulation} \label{subsec:notation}
Let $X \in \mathcal{X} = \mathbb{R}^m$ be a variable designating a vector of covariates; $Y \in \mathcal{Y} = \{0, 1\}$ be a binary indicator of an outcome; and $A \in \mathcal{A}$ be a discrete indicator for a protected or sensitive attribute, such as race, ethnicity, gender, sex, or age with $K$ groups. 
The objective of supervised learning with binary outcomes is to use data $\mathcal{D} = \{(x_i, y_i, a_i)\}_{i = 1}^N \sim P(X, Y, A)$ to learn a function $f_{\theta}(X) : \mathbb{R}^{m} \rightarrow [0,1]$, parameterized by $\theta$, to estimate $\E[Y \mid X] = P(Y = 1 \mid X)$. The output of the predictor may be further thresholded at $T$ to produce binarized predictions $\hat{Y} = \mathbbm{1}[f_{\theta}(X) \geq T] \in \{0, 1\}$. 
For notational convenience, we refer to an empirical mean defined over a dataset as an expectation involving $\mathcal{D}$. For example, $\E_{x \sim \mathcal{D}\mid Y=1} f_{\theta}(x)$ refers to the empirical mean of the predicted probability of the outcome over the set of patients for whom the outcome is observed.

Many of the group fairness criteria that we evaluate can be stated as a form of conditional independence involving a function of some or all of the variables $\{X, Y, f_{\theta}, \hat{Y}$\} with $A$.
Given access to a dataset and model predictions, quantifying the extent to which a group fairness criterion is violated relies on a fairness metric that compares the empirical distribution of the relevant function across groups.
The objective of supervised learning subject to group fairness is to learn a model $f_{\theta}$ such that the violation of one or more fairness criteria is small. 
This framing does not preclude the use of the group indicator $A$ as a component of the vector of covariates $X$.

We consider fairness criteria belonging to three general categories: conditional prediction parity, calibration and cross-group ranking. The first - conditional prediction parity - will later be incorporated into our training objective. Below, we describe the motivation behind considering each class, as well as mathematically define particular metrics which will be used to evaluate our experiments.

\subsubsection{Fairness Criteria Based on Conditional Prediction Parity} \label{subsubsec:cond_pred_parity}
We refer to a class of group fairness criteria that assess conditional independence between model predictions $\hat{Y}$ or $f_{\theta}$ and the categorical group indicator $A$ as \textit{conditional prediction parity}.
When an instance of a fairness criterion of this class depends on a binarized prediction $\hat{Y}$, we say it is a \textit{threshold-based criterion}, and when it depends on a probabilistic prediction $f_{\theta}$, we say it is a \textit{threshold-free criterion}.
This form captures \textit{demographic parity} ($\hat{Y} \perp A$ or $f_{\theta} \perp A$) \citep{Calders2009,Dwork2011,Zemel2013}, \textit{equalized odds} ($\hat{Y} \perp A \mid Y$ or $f_{\theta} \perp A \mid Y$) \cite{Hardt2016}, and \textit{equal opportunity} ($\hat{Y} \perp A \mid Y = 1$ or $f_{\theta} \perp A \mid Y = 1$) \cite{Hardt2016}, among others \cite{Celis2018}.

These criteria have natural interpretations that can motivate their use in some contexts.
For instance, violations of demographic parity can imply allocation discrepancies across groups.
Equal opportunity and equalized odds can be understood as either a notion of demographic parity within strata defined by the outcome $Y$ or as a notion of error rate balance.
For binarized predictions $\hat{Y}$, the equal opportunity criterion implies equal true positive rates across groups, and the equalized odds criterion implies equal opportunity as well as equal false positive rates across groups.
Satisfying the threshold-free variants of these criteria implies satisfaction of the threshold-criteria at any threshold.
An implication of this is that the threshold-free variant of the equalized odds criteria corresponds to the condition of identical ROC curves across groups \cite{Hardt2016}.

We now present metrics that assess the extent to which a predictive model violates notions of threshold-free conditional prediction parity, starting with demographic parity ($M_{\textrm{DP}}$) before extending it to equal opportunity ($M_{\textrm{EqOpp}}$) and equalized odds ($M_{\textrm{EqOdds}}$).
Threshold-free demographic parity is satisfied if $P(f_{\theta} \mid A=A_k)$ matches $P(f_{\theta})$, for each group $A_k$.
This motivates the use of a metric of the following form:
\begin{equation} \label{eq:div_objective_dp}
    M_{\textrm{DP}} = \sum_{A_k \in \mathcal{A}} D(P(f_{\theta} \mid A=A_k) \mid \mid P(f_{\theta})),
\end{equation}
where $D$ is a function that measures a notion of distance between probability distributions.
As equalized odds and equal opportunity may be interpreted as demographic parity within strata defined by the outcome, the associated metrics can be constructed analogously, with
\begin{equation} \label{eq:div_objective_eqopp}
    M_{\textrm{EqOpp}} = \sum_{A_k \in \mathcal{A}} D(P(f_{\theta} \mid A=A_k, Y=1) \mid \mid P(f_{\theta} \mid Y=1))
\end{equation}
for equal opportunity, and
\begin{equation} \label{eq:div_objective_eqodds}
    M_{\textrm{EqOdds}} = \sum_{Y_j \in \mathcal{Y}} \sum_{A_k \in \mathcal{A}} D(P(f_{\theta} \mid A=A_k, Y = Y_j) \mid \mid P(f_{\theta} \mid Y=Y_j))
\end{equation}
for equalized odds.

With this construction, the form of function $D$ determines the properties of the metric used to assess fairness criteria violation.
If $D$ is chosen to be a divergence or integral probability metric \cite{Sriperumbudur2009}, then the fairness criteria is satisfied if the associated metric $M$ is equal to zero, and is positive otherwise.
In our experiments, we evaluate these metrics using the Earth Mover's Distance (EMD) \cite{Ramdas2017}, an example of an integral probability metric.

These metrics may be further decomposed into group- and outcome- specific components. 
For instance, if
\begin{equation} \label{eq:div_group_dp}
    M_k = D(P(f_{\theta} \mid A=A_k) \mid \mid P(f_{\theta})),
\end{equation}
then $M_{\textrm{DP}} = \sum_{k=1}^{K} M_k$.
Similarly, if 
\begin{equation} \label{eq:div_group_eqopp}
    M_k^1 = D(P(f_{\theta} \mid A=A_k, Y=1) \mid \mid P(f_{\theta} \mid Y=1)),
\end{equation}
and 
\begin{equation} \label{eq:div_group_eqodds}
    M_k^0 = D(P(f_{\theta} \mid A=A_k, Y=0) \mid \mid P(f_{\theta} \mid Y=0)),
\end{equation}
then $M_{\textrm{EqOpp}} = \sum_{k=1}^{K} M_k^1$, and $M_{\textrm{EqOdds}} = M_{\textrm{EqOpp}} + \sum_{k=1}^{K} M_k^0$.

As an alternative, we also consider metrics based on the difference in means between the distributions of interest.
As before, these metrics decompose into group-specific components.
For instance, if we define the difference in the mean predicted probability of the outcome for patients in group $k$ versus the marginal distribution ($M_k^{\textrm{mean}}$) as
\begin{equation} \label{eq:mean_group_dp}
    M_k^{\textrm{mean}} = \E_{x \sim \mathcal{D} \mid A=A_k} [f_{\theta}(x)] - \E_{x \sim \mathcal{D}} [f_{\theta}(x)],
\end{equation}
then an aggregate metric of demographic parity violation can be constructed as
\begin{equation} \label{eq:mean_objective_dp}
    M_{\textrm{DP}}^{\textrm{mean}} = \sum_{k=1}^K (M_k^{\textrm{mean}})^2.
\end{equation}
Expressions for equal opportunity and equalized odds can be constructed analogously by nesting these comparisons within strata defined by the observed outcomes. If 
\begin{equation} \label{eq:mean_group_eqopp}
    M_k^{1,\textrm{mean}} = \E_{x \sim \mathcal{D} \mid A=A_k, Y=1} [f_{\theta}(x)] - \E_{x \sim \mathcal{D}\mid Y=1} [f_{\theta}(x)],
\end{equation}
and 
\begin{equation} \label{eq:mean_group_eqodds}
    M_k^{0,\textrm{mean}} = \E_{x \sim \mathcal{D} \mid A=A_k, Y=0} [f_{\theta}(x)] - \E_{x \sim \mathcal{D}\mid Y=0} [f_{\theta}(x)],
\end{equation}
then
\begin{equation} \label{eq:mean_objective_eqopp}
    M_{\textrm{EqOpp}}^{\textrm{mean}} = \sum_{k=1}^K (M_k^{1,\textrm{mean}})^2
\end{equation}
and
\begin{equation} \label{eq:mean_objective_eqodds}
    M_{\textrm{EqOdds}}^{\textrm{mean}} =  M_{\textrm{EqOpp}}^{\textrm{mean}} + \sum_{k=1}^K (M_k^{0,\textrm{mean}})^2.
\end{equation}

\subsubsection{Fairness Criteria Based on Calibration} \label{subsubsec:calibration}
We construct a measure of fairness in calibration to estimate the extent to which observed event rates conditioned on the predicted risk differ across groups. 
We do so by first presenting a measure of absolute calibration, which can be understood as a performance metric that assesses overall model calibration, before extending that measure to account for relative differences in calibration across groups.

Calibration in the context of clinical risk prediction typically refers to the extent to which probabilistic predictions are faithful estimates of the observed event rates, such that a model is well-calibrated if $P(Y = 1 \mid f_{\theta} = p)$ = $p$ for $p \in [0, 1]$.
In other words, a model is well-calibrated if the outcome is observed $p*100\%$ of the time among patients whose prediction is $p$.

The absolute calibration error (ACE) assesses deviations from perfect calibration, relying on an auxiliary estimator $g_{\phi}: [0, 1] \rightarrow [0, 1]$ that provides an approximation of $P(Y = 1 \mid f_{\theta})$:
\begin{equation} \label{eq:ACE}
    \textrm{ACE} = \E_{x \sim \mathcal{D}} \Big[\Big(g_{\phi}\big(f_{\theta}(x)\big) - f_{\theta}(x)\Big)^2\Big].
\end{equation}
This measure may be interpreted as an instance of the weighted mean squared calibration error proposed in Yadlowsky \textit{et al.} \cite{Yadlowsky2019}, used here in the context of uncensored binary outcomes, and is also similar to the integrated calibration index proposed in Austin \textit{et al.} \cite{Austin2019}.
In addition, we consider a signed variant of the metric:
\begin{equation} \label{eq:ACE_signed}
    \textrm{ACE}^{\textrm{signed}} = \E_{x \sim \mathcal{D}} \Big[g_{\phi}\big(f_{\theta}(x)\big) - f_{\theta}(x)\Big].
\end{equation}
The signed measure can assess the directionality of mis-calibration, but may be misleading when positive deviations offset negative ones. For clarity, a positive value for the signed measure corresponds to \textit{under-prediction} of risk, as it corresponds to an excess number of observed outcomes given the risk estimates.

In the context of group fairness, it is relevant to reason about model calibration in a \textit{relative} sense.
The \textit{matching conditional frequencies} \cite{Hardt2016,Chouldechova2017} (MCF) or \textit{sufficiency} \cite{Liu2019} condition requires that the probability of the outcome not differ across groups conditioned on the value of the risk score \textit{i.e.} $Y \perp A \mid f_{\theta}$.
An alternative (but not equivalent) formulation is to require the risk score to be well-calibrated within groups \cite{Kleinberg2016,pleiss2017fairness}.
We present a measure of relative calibration to estimate the extent to which observed event rates conditioned on the predicted risk differ across groups, thus assessing the violation of the MCF criterion.
Given an auxiliary estimator $g_{\phi}$ that estimates the marginal density $P(Y = 1 \mid f_{\theta})$ for the entire population and an estimator $g_k$ that estimates $P(Y = 1 \mid f_{\theta}, A=A_k)$, the relative calibration error (RCE) for group $A_k$ is defined as
\begin{equation} \label{eq:RCE}
    \textrm{RCE}_k = \E_{x \sim \mathcal{D}\mid A=A_k} \Big[\Big(g_k\big(f_{\theta}(x)\big) - g_{\phi}\big(f_{\theta}(x)\big)\Big)^2\Big].
\end{equation}
with the corresponding signed metric defined as
\begin{equation} \label{eq:RCE_signed}
    \textrm{RCE}_k^{\textrm{signed}} = \E_{x \sim \mathcal{D}\mid A=A_k}\Big[g_{k}(f_{\theta}(x)) - g_{\phi}(f_{\theta}(x))\Big].
\end{equation}
The measures that we present here depend on the form of the estimators $g_{\phi}$ and $g_k$. In our experiments, we use logistic regression as an estimator. Alternatives include loess regression \cite{Austin2019}, kernel density estimates over $P(f_{\theta} \mid Y)$ combined with simple conditional probability rules, or binning estimators.

\subsubsection{Cross-Group Ranking Measures} \label{subsubsec:cross_group_ranking}
We now consider notions of cross-group ranking performance \cite{Kallus2019,beutel2019fairness}. These measures can be understood as variations on the area under the receiver operating characteristic curve (AUROC). The AUROC is a typical metric for assessing the performance of a clinical predictive model. This metric has an equivalent interpretation as the probability with which predictions for members of the positive class are ranked above those for members of the negative class:
\begin{equation}
    \textrm{AUROC} = \E_{x^1 \sim \mathcal{D}\mid Y=1} \E_{x^0 \sim \mathcal{D}\mid Y=0} \mathbbm{1}(f_{\theta}(x^1) > f_{\theta}(x^0)].
\end{equation}
A comparison of the AUROC across groups can be misleading as a comparison of relative model quality and as a fairness criterion since that comparison does not account for the degree with which positive examples of a group are ranked above negative examples of other groups. In order to gain insight into this phenomenon, we consider multi-group extensions to measures defined in the xAUC framework \cite{Kallus2019,beutel2019fairness} and consider deviations in these measures across groups as a violation of a measure of group fairness.
We define $\textrm{xAUC}_k^1$ as the probability with which positive instances of group $A_k$ are ranked above negative instances of all other groups:
\begin{equation}
\label{eq:xauc_1}
    \textrm{xAUC}_k^1 = \E_{x^1 \sim \mathcal{D}\mid Y=1, A=A_k} \E_{x^0 \sim \mathcal{D}\mid Y=0, A \neq A_k} \mathbbm{1}[f_{\theta}(x^1) > f_{\theta}(x^0)].
\end{equation}
Similarly, we define $\textrm{xAUC}_k^0$ as the probability with which negative instances of group $A_k$ are ranked below positive instances of all other groups:
\begin{equation}
\label{eq:xauc_0}
    \textrm{xAUC}_k^0 = \E_{x^1 \sim \mathcal{D}\mid Y=1, A \neq A_k} \E_{x^0 \sim \mathcal{D}\mid Y=0, A = A_k} \mathbbm{1}[f_{\theta}(x^1) > f_{\theta}(x^0)].
\end{equation}

\subsubsection{Regularized Objectives for Conditional Prediction Parity} \label{subsec:optimization}
Several techniques for constructing models that satisfy measures of group fairness have been proposed, including representation learning \cite{Zemel2013,Louizos2015,Madras2018,Ilvento2020}, post-processing \cite{Hardt2016}, and constrained or regularized learning objectives \cite{Agarwal2018,pmlr-v54-zafar17a,Cotter2018,Cotter2019,Celis2018}.
We focus our attention on regularized learning objectives that penalize violations of measures of threshold-free conditional prediction parity.
Given a loss function $\mathcal{L}$, such as the negative log-likelihood, the general form of this objective is as follows:
\begin{equation} \label{eq:reg_objective}
    \min_{\theta} \E_{(x, y) \sim \mathcal{D}} \mathcal{L}(y, f_{\theta}(x)) + \lambda R,
\end{equation}
where $R$ is a non-negative regularizer indicative of the extent to which the fairness criterion is violated, and $\lambda$ is a non-negative scalar. In general, tuning the value of $\lambda$ allows for exploring the trade-offs between measures of model performance and fairness.

In our experiments, we evaluate six regularization strategies to penalize the violation of threshold-free conditional prediction parity measures. These strategies correspond to regularizers that penalize violation of demographic parity, equal opportunity, and equalized odds, either with regularizers of the form of Equations (\ref{eq:div_objective_dp}), (\ref{eq:div_objective_eqopp}), or (\ref{eq:div_objective_eqodds}), respectively, using an empirical maximum mean discrepancy (MMD) \cite{gretton2012kernel} with a Gaussian kernel to estimate $D$, or with regularizers in the form of Equations (\ref{eq:mean_objective_dp}), (\ref{eq:mean_objective_eqopp}), or (\ref{eq:mean_objective_eqodds}), respectively, that penalize the sum of the squared difference in the mean prediction in the relevant outcome strata.
During training, the value of the regularizer $R$ is computed on the basis of comparisons between the distributions of $\log f_{\theta}$, computed via a log-softmax transformation of the model logits, rather than on the basis on $f_{\theta}$ directly.
In some instances, we refer to regularizers that penalize violation of demographic parity, equalized odds, and equal opportunity as \textit{unconditional}, \textit{conditional}, and \textit{positive conditional}, respectively, corresponding to the outcome strata used to evaluate the regularizer, and include a modifier for either \textit{MMD} or \textit{mean} to indicate how the relevant distributions are compared.

\subsection{Experiments} \label{subsec:experiments}

\subsubsection{Datasets} \label{subsec:datasets}
\textbf{STARR}
The Stanford Medicine Research Data Repository (STARR) \citep{Datta2020} is a clinical data warehouse containing records from approximately three million patients from Stanford Hospitals and Clinics and the Lucile Packard Children's Hospital for inpatient and outpatient clinical encounters that occurred between 1990 and 2020. This database contains structured longitudinal data in the form of diagnoses, procedures, medications, and laboratory tests that have been mapped to standard concept identifiers in the Observational Medical Outcomes Partnership Common Data Model (OMOP CDM) version 5.3.1 \cite{Hripcsak2015,Overhage2012,Reps2018}. De-identified clinical notes with clinical concepts extracted and annotated with negation and family-history detection are also made available \cite{Datta2020}.
\\ \\
\textbf{Optum©’s de-identifed Clinformatics® Data Mart Database}
Optum©’s de-identifed Clinformatics® Data Mart Database (Optum CDM) is a statistically 
de-identified large commercial and medicare advantage claims database.
The database includes approximately 17-19 million annual covered lives, for a total of over 57 million unique lives over a 9 year period (1/2007 through 12/2017). 
We utilize a variant of the database that makes available the month and date of death, sourced from internal and external sources including the Death Master File maintained by the Social
Security Office, as well as records from the Center for Medicare and Medicaid Services.
However, this version of the data does not provide access to detailed socioeconomic, demographic, or geographic variables.
We utilize version 7.1 of the data mapped to OMOP CDM v5.3.1.
\\ \\
\textbf{MIMIC-III}
The Medical Information Mart for Intensive Care-III (MIMIC-III) \cite{Johnson2016} is a widely studied database containing comprehensive physiologic information from approximately fifty thousand intensive care unit (ICU) patients admitted to the Beth Israel Deaconess Medical Center between 2001 and 2012. We utilize MIMIC-OMOP\footnote{https://github.com/MIT-LCP/mimic-omop/tree/fa5113c3f0777e74d2a6b302322477e6fe666910}, a variant of the database that has been mapped to OMOP CDM v5.3.1.

\subsubsection{Cohort, Outcome, and Attribute Definitions} \label{subsec:cohort_attributes}

\begin{table}[!t]
\centering
\caption{Cohort characteristics for patients drawn from STARR. Data are grouped on the basis of age, sex, and the race and ethnicity category. Shown, for each group, is the number of patients extracted and the incidence of hospital mortality, prolonged length of stay, and 30-day readmission}
\label{tab:cohort_starr}
\begin{tabular}{lrrrr}
\toprule
{} & {} & \multicolumn{3}{c}{Outcome Incidence} \\
\cmidrule{3-5}
Group & Count &  Hospital Mortality &  Prolonged Length of Stay &  30-Day Readmission \\
\midrule
\lbrack18-30)                   &   23,042 &             0.00681 &  0.175 &          0.0461 \\
\lbrack30-45)                   &   43,432 &             0.00596 &   0.130 &          0.0396 \\
\lbrack45-55)                   &   27,394 &              0.0178 &  0.205 &          0.0527 \\
\lbrack55-65)                   &   35,703 &              0.0251 &  0.227 &          0.0558 \\
\lbrack65-75)                   &   36,084 &              0.0284 &  0.234 &          0.0548 \\
\lbrack75-90)                   &   32,989 &                0.0400 &  0.238 &          0.0545 \\
\midrule
Female                    &  112,713 &              0.0161 &  0.166 &          0.0452 \\
Male                     &   85,923 &              0.0271 &  0.244 &          0.0571 \\
\midrule
Asian                     &   29,460 &              0.0209 &  0.171 &          0.0536 \\
Black &    7,813 &              0.0198 &   0.240 &          0.0581 \\
Hispanic        &   33,742 &               0.0180 &  0.193 &          0.0544 \\
Other                     &   20,270 &              0.0327 &  0.226 &          0.0442 \\
White                     &  107,359 &              0.0196 &  0.202 &          0.0488 \\
\bottomrule
\end{tabular}
\end{table}

We define a set of analogous cohort, outcome, attribute and group definitions for each database that differ on the basis of differential availability of data elements. For the STARR and Optum CDM databases, we consider the set of inpatient hospital admissions that span at least two distinct calendar dates for which patients were of age 18 years or older at the time of admission. 
For admissions in the STARR database, the start of the admission is variably defined as the date and time of admission to the emergency department, if available, and admission to the hospital otherwise.
For patients with more than one admission meeting this criteria, we randomly sample one admission.
In each case, we consider the start of an admission as the index date and time.
For each admission, we derive binary outcome labels for prolonged length of stay (defined as a hospital length of stay greater than or equal to seven days) and 30-day readmission (defined as a subsequent admission within thirty days of discharge of the considered admission).
For admissions derived from the STARR database, we also derive a binary outcome label for in-hospital mortality.
This procedure returns admissions from 198,644 patients in STARR (Table \ref{tab:cohort_starr}) and 8,074,571 patients in Optum CDM (Supplementary Table \ref{tab:cohort_optum}).

For MIMIC-III, we replicate in MIMIC-OMOP the cohort and outcome definitions defined as benchmarks in the \texttt{MIMIC-Extract} project \cite{Wang2020}. In particular, we consider hospital admissions associated with each patient's first ICU stay, further restricting the set of allowable ICU stays to those lasting between twelve hours and ten days. We set the index date and time to be twenty-four hours after hospital admission, and further restrict the set of admissions to those for which the index time is at least six hours prior to ICU discharge and for which patients are between 15 and 89 years old at the index time. We define four binary outcomes, corresponding to ICU length of stay greater than three and seven days and mortality over the course of the ICU stay and hospital admission, following the approach of Wang \textit{et al.} \cite{Wang2020}. This procedure returns 26,170 patients (Supplementary Table \ref{tab:cohort_mimic}).

For the purposes of evaluating the extent to which a model satisfies group fairness criteria, we define discrete groups of the population on the basis of demographic attributes. 
We consider (1) a combined race and ethnicity variable based on self-reported racial and ethnic categories, (2) sex\footnote{We note that ``gender'' is the term used in the OMOP CDM, but the stated definition of the underlying concept in each of the data sources refers to biological sex. This field is almost always recorded as either male or female. We observe eight occurrences that do not map to male or female in the derived STARR cohort and 1,176 occurrences in the Optum CDM cohort. We exclude these patients from model training and evaluation when sex is considered as the sensitive attribute, but include them otherwise.}, and (3) age at the index date, discretized into categories at 18-29, 30-44, 45-54, 55-64, 65-74, 75-89 years for STARR and Optum cohorts and 15-29, 30-44, 45-54, 55-64, 65-74, 75-89 years for MIMIC-III. 

The race and ethnicity attribute is constructed by assigning Hispanic if the ethnicity is recorded as Hispanic, and the value of the recorded racial category otherwise. In line with the categories provided by the upper level of the OMOP CDM vocabulary, we consider groups corresponding to ``Asian'', ``American Indian or Alaska Native'', ``Black or African American'', ``Native Hawaiian or Other Pacific Islander'', ``Other'', and ``White''. We additionally aggregate groups of the race and ethnicity attribute with the ``Other'' category when few observed outcomes are available within a group in a database specific manner. This reduces the categorization to ``Asian'', ``Black or African American'', ``Hispanic'', ``Other'', and ``White'' for STARR, and to ``Other'' and ``White'' for MIMIC-III. Race and ethnicity data is not available for the version of the Optum CDM database that we use. We further contextualize this operationalization of race and ethnicity in Section \ref{sec:discussion}.

\subsubsection{Feature Extraction} \label{subsec:features}
We extract clinical features with a generic procedure that operates on OMOP CDM databases, similar to Reps \textit{et al.} \cite{Reps2018}.
A graphical depiction of this procedure is provided in Supplementary Figure \ref{fig:supplement/feature_extraction}.
The process returns 539,823, 438,369, and 21,026 unique features in the STARR, Optum CDM, and MIMIC-III cohorts, respectively.
In brief, we consider a set of binary features based on the occurrence of both unique OMOP CDM concepts and derived elements prior to the index date and time.
We extract all diagnoses, medication orders, procedures, device exposures, encounters, and labs, assigning a unique feature identifier for the presence of each OMOP CDM concept, in time intervals defined relative to the index date and time.
For lab test results, we construct additional features for lab results above and below the corresponding reference range, if available, and indicators for whether the result belongs to the empirical quintiles for the corresponding lab result observed in the time interval.
For STARR data, we include additional features for the presence of clinical concepts derived from clinical notes, modified by an indicator for whether the concept corresponds to a present and positive mention \cite{Datta2020}.
For MIMIC-III, we do not use diagnoses or clinical notes to derive features.
We include time-agnostic demographic features, including race, ethnicity, gender, and age group (discretized in five year intervals), using the OMOP CDM concept identifiers directly rather than the attribute definitions described previously.

We repeat the feature extraction procedure for a set of time-intervals defined relative to the index date and time in a database-specific manner. For the STARR and Optum CDM cohorts, we define intervals at 29 to 1 days prior to the index date, 89 to 30 days prior, 179 to 90 days prior, 364 to 180 days prior, and any time prior. 
For STARR and MIMIC-III data, we include additional time-intervals defined only over the subset of the data elements recorded with date and time resolution, repeating each extraction procedure described for which that set is not empty.
For STARR, these intervals correspond to 4 hours prior to the index time, 12 hours to 4 hours prior to the index time, 24 to 12 hours prior, three days to 24 hours prior, and seven days to three days prior. 
For MIMIC-III, these intervals correspond to 4 hours prior to the index time, 12 hours to 4 hours prior to the index time, 24 to 12 hours prior, and any time prior.

\subsubsection{Model Training} \label{subsec:training}
We conduct a modified cross-validation procedure designed to enable robust model selection and evaluation.
For each cohort, a randomly sampled partition of 10\% of the patients is set aside as a test set for final evaluation.
For STARR and MIMIC-III, we further partition the remaining data into ten equally-sized folds, each of which can be considered as a validation set corresponding to a training set composed of the remainder of the folds. 
Due to the computational constraints imposed by the large size of the Optum CDM cohort, we do not perform cross-validation and instead randomly partition the data such that 81\% of the patients are used for training, 9\% of the patients are used for validation, and 10\% used for testing.

In all cases, we leverage fully-connected feedforward neural networks for prediction, as they enable scalable and flexible learning for differentiable objectives in the form of equation (\ref{eq:reg_objective}). Tuning of the unpenalized models begins with a random sample of fifty hyperparameter configurations from a grid of architectural and training-dynamic hyperparameters (Supplementary Table \ref{tab:hyperparameter_grid}).
For each combination of dataset, outcome, training-validation partition, and hyperparameter configuration, we train a model with the Adam \cite{kingma2014adam} optimizer for up to 150 iterations of 100 batches, terminating early if the cross-entropy loss on the validation set does not improve for 10 iterations. 
For each combination of dataset and outcome, we select model hyperparameters on the basis of the mean validation log-loss across folds (the selected hyperparameters are provided in Supplementary Tables \ref{tab:hyperparameter_starr}, \ref{tab:hyperparameter_mimic}, and \ref{tab:hyperparameter_optum}).
For models trained on the Optum CDM cohort, we select model hyperparameters on the basis of the log-loss measured on the validation set.
We then train models with regularization that penalizes fairness criteria violation using an objective in the form of Equation (\ref{eq:reg_objective}) for ten values of $\lambda$ distributed log-uniformly on the interval $10^{-3}$ to $10$, repeating the process separately for each dataset, task, attribute, and form of regularizer (of which there are six, as discussed in Section \ref{subsec:optimization}), holding the model hyperparameters fixed to those selected for the corresponding unpenalized model. 
As before, we train each model for up to 150 iterations of 100 batches, but perform early stopping and model selection on the basis of the penalized loss that incorporates the regularization term.
Pytorch version 1.5.0 \cite{pytorch} is used to define all models and training procedures.

\subsubsection{Model Evaluation} \label{subsec:model_evaluation}
We report model performance and fairness metrics as the mean $\pm$ the standard deviation (SD) for metrics evaluated on the held-out test set for the set of ten models derived from the procedure described in Section \ref{subsec:training}.
For models trained on the Optum CDM cohort, we report results on the test set for the selected model without an associated standard deviation.
To assess within-group model performance, we report the AUROC, average precision, and cross entropy loss at baseline and each value of $\lambda$.
To assess conditional prediction parity violation, we report on the decomposed group-specific components of the metrics presented in Section \ref{subsec:form_fairness}.
In particular, we report the EMD and difference in means between the distribution of predictions for each group with the marginal distribution constructed via aggregation of predictions from all groups (Equations (\ref{eq:div_group_dp}) and (\ref{eq:mean_group_dp})), respectively, as metrics that assess violations of demographic parity.
We repeat the process in the strata of the population for which the outcome is and is not observed, as metrics that assess violations of equalized odds and equal opportunity (Equations (\ref{eq:div_group_eqodds}), (\ref{eq:mean_group_eqodds}), (\ref{eq:div_group_eqopp}), and (\ref{eq:mean_group_eqopp})).
We report on cross-group ranking discrepancies in the form of Equations (\ref{eq:xauc_1}) and (\ref{eq:xauc_0}) for each group.

To assess absolute and relative model calibration, we estimate ACE and RCE post-hoc on the test set. To do so, for each predictive model $f_{\theta}$, we train an auxiliary logistic regression model to estimate $P(Y \mid f_{\theta})$ on the basis of $\log(f_{\theta})$ for the aggregate test set and for each group. The resulting group-level estimates of ACE and RCE are constructed by plugging-in the data to the relevant estimators using Equations (\ref{eq:ACE}), (\ref{eq:ACE_signed}), (\ref{eq:RCE}), and (\ref{eq:RCE_signed}). The logistic regression models are fit using the LBFGS \cite{fletcher2013practical} algorithm implemented in Scikit-Learn \cite{scikit-learn}. We report ACE alongside model performance measures and RCE alongside other fairness metrics.

\section{Results} \label{sec:results}
\begin{figure*}[!ht]
	\centering
	\includegraphics[width=0.85\linewidth]{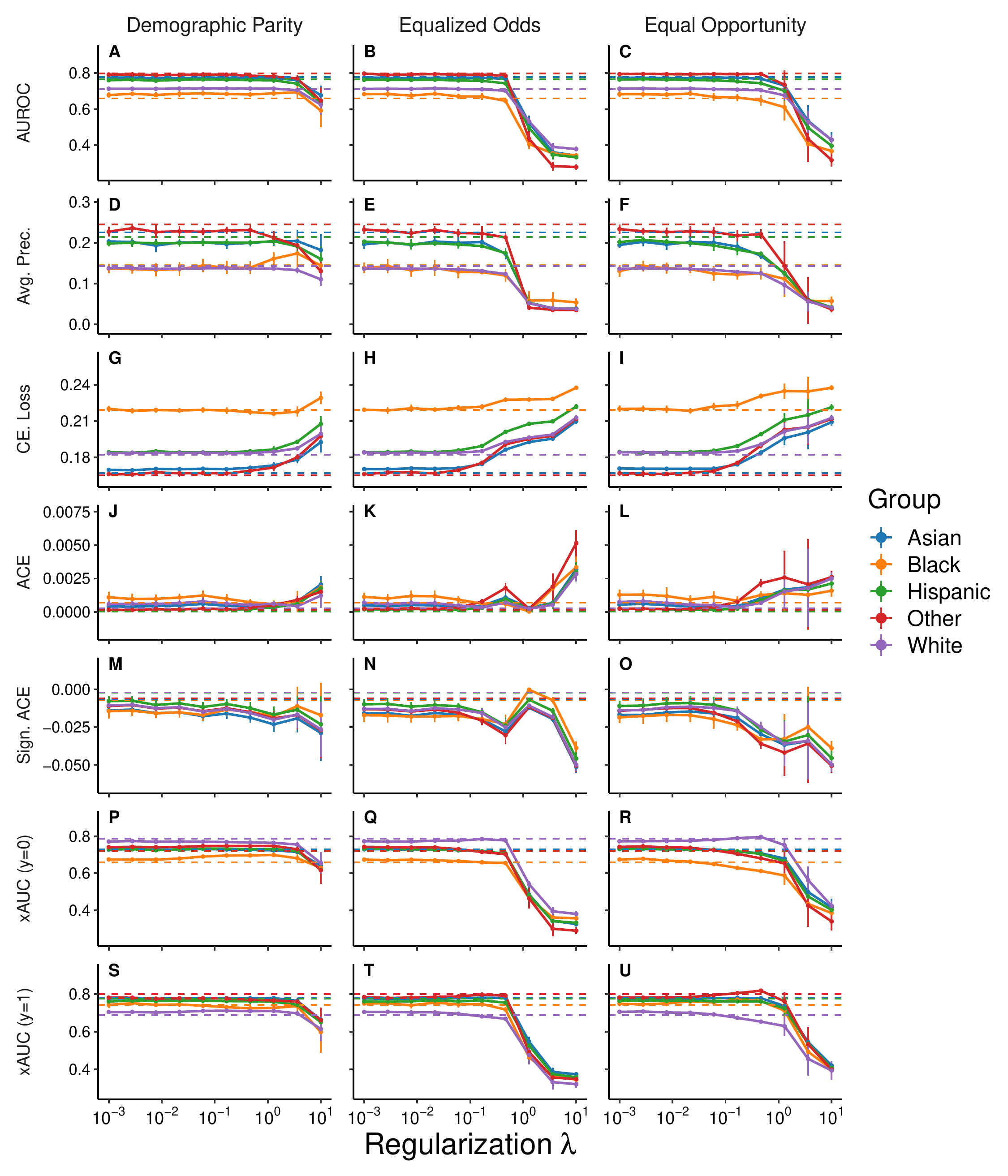}
	\caption{
	    Group-level model performance measures as a function of the extent $\lambda$ that violation of the fairness criterion is penalized when the race and ethnicity category is considered as the sensitive attribute for prediction of 30-day readmission in the STARR database. Results shown are the mean $\pm$ SD for the area under the ROC curve (AUROC), average precision (Avg. Prec), the cross entropy loss (CE Loss), the absolute calibration error (ACE), the signed absolute calibration error (Sign. ACE), and cross group ranking performance (xAUC; $\textrm{xAUC}_k^1$ is indicated by (y=1) and $\textrm{xAUC}_k^0$ by (y=0)) for each group for objectives that penalize violation of threshold-free Demographic Parity, Equalized Odds, and Equal Opportunity with MMD-based penalties. Dashed lines correspond to the mean result for the unpenalized training procedure.
	}
	\label{fig:main_text/starr/all_performance/readmission_30/race_eth}
\end{figure*}

\begin{figure*}[!htb]
	\centering
	\includegraphics[width=0.85\linewidth]{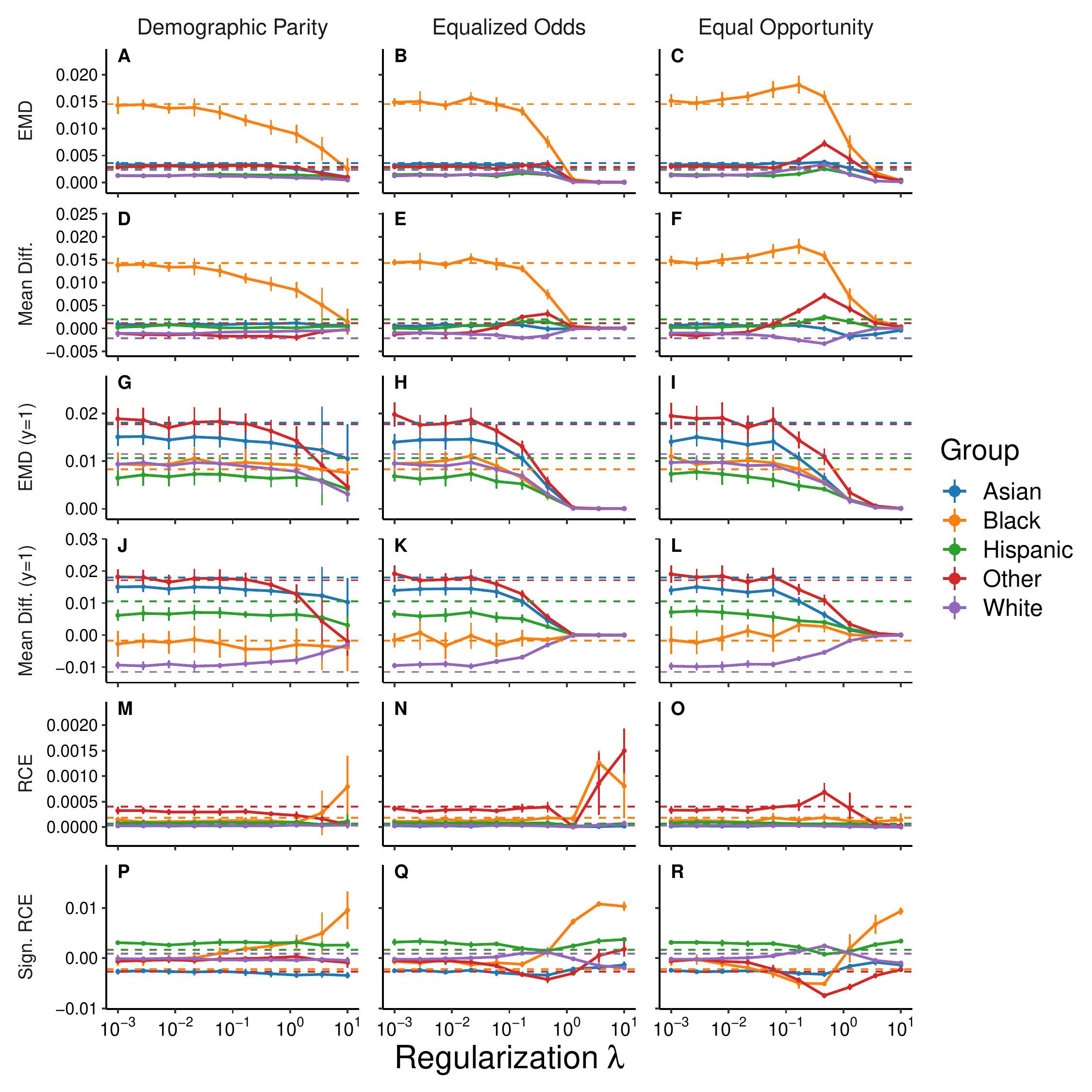}
	\caption{
	    Fairness metrics as a function of the extent $\lambda$ that violation of the fairness criterion is penalized when the race and ethnicity category is considered as the sensitive attribute for prediction of 30-day readmission in the STARR database. Results shown are the mean $\pm$ SD for decomposed group-level metrics that assess conditional prediction parity (EMD and Mean Diff.) and relative calibration error (RCE and Sign. RCE) for objectives that penalize violation of threshold-free Demographic Parity, Equalized Odds, and Equal Opportunity on the basis of MMD-based penalties.  Measures of conditional prediction parity are separately assessed in the whole population and in the strata for which the outcome is observed (suffixed with (y=1)). Dashed lines correspond to the mean result for the unpenalized training procedure.
	}
	\label{fig:main_text/starr/all_fairness/readmission_30/race_eth}
\end{figure*}

\begin{figure*}[!ht]
	\centering
	\includegraphics[width=0.85\linewidth]{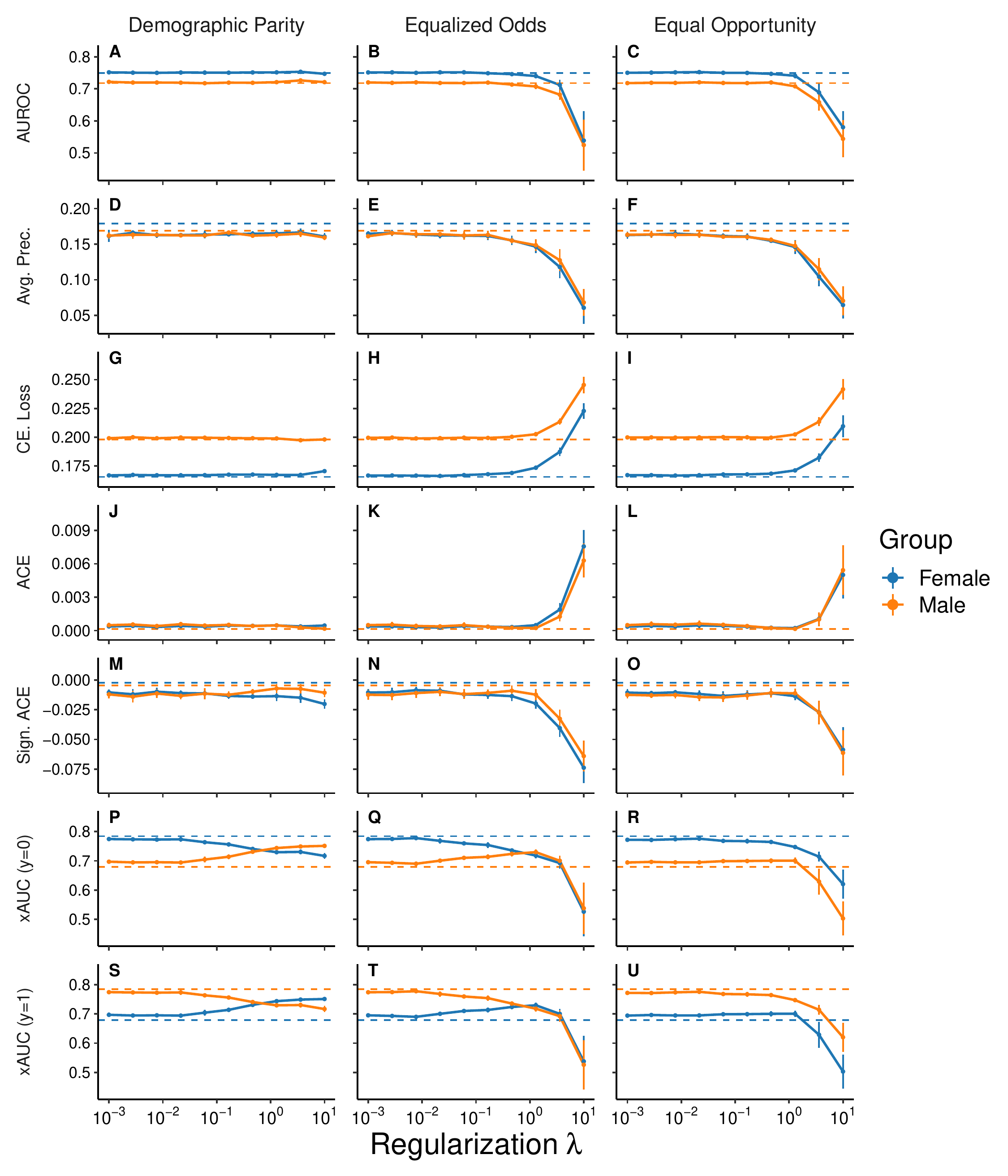}
	\caption{
	    Group-level model performance measures as a function of the extent $\lambda$ that violation of the fairness criterion is penalized when sex is considered as the sensitive attribute for prediction of 30-day readmission in the STARR database. Results shown are the mean $\pm$ SD for the area under the ROC curve (AUROC), average precision (Avg. Prec), the cross entropy loss (CE Loss), the absolute calibration error (ACE), the signed absolute calibration error (Sign. ACE), and cross group ranking performance (xAUC; $\textrm{xAUC}_k^1$ is indicated by (y=1) and $\textrm{xAUC}_k^0$ by (y=0)) for each group for objectives that penalize violation of threshold-free Demographic Parity, Equalized Odds, and Equal Opportunity with MMD-based penalties. Dashed lines correspond to the mean result for the unpenalized training procedure.
	}
	\label{fig:main_text/starr/all_performance/readmission_30/gender_concept_name}
\end{figure*}

\begin{figure*}[!ht]
	\centering
	\includegraphics[width=0.85\linewidth]{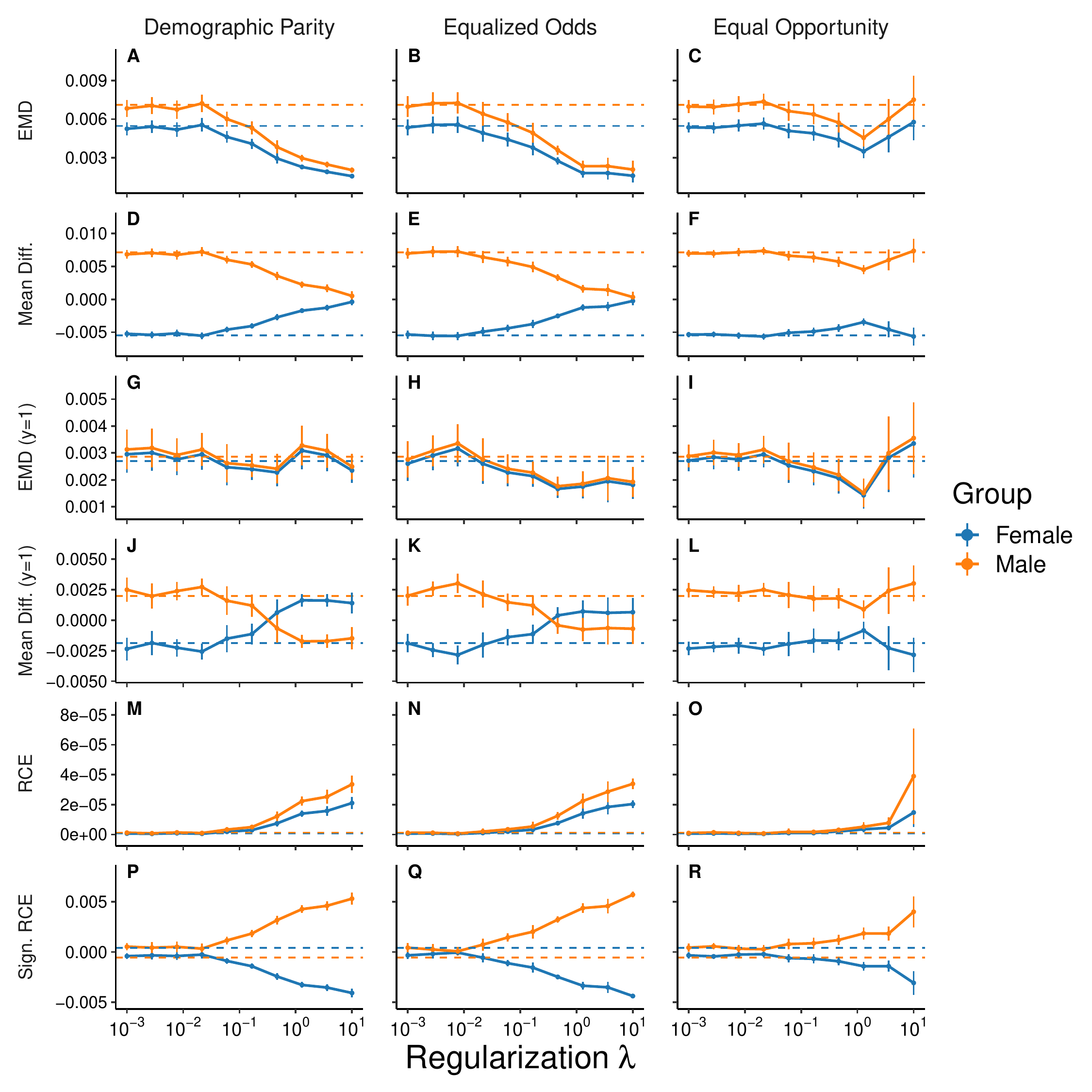}
	\caption{
	    Fairness metrics as a function of the extent $\lambda$ that violation of the fairness criterion is penalized when sex is considered as the sensitive attribute for prediction of 30-day readmission in the STARR database. Results shown are the mean $\pm$ SD for decomposed group-level metrics that assess conditional prediction parity (EMD and Mean Diff.) and relative calibration error (RCE and Sign. RCE) for objectives that penalize violation of threshold-free Demographic Parity, Equalized Odds, and Equal Opportunity on the basis of MMD-based penalties.  Measures of conditional prediction parity are separately assessed in the whole population and in the strata for which the outcome is observed (suffixed with (y=1)). Dashed lines correspond to the mean result for the unpenalized training procedure.
	}
	\label{fig:main_text/starr/all_fairness/readmission_30/gender_concept_name}
\end{figure*}

\begin{figure*}[!ht]
	\centering
	\includegraphics[width=0.85\linewidth]{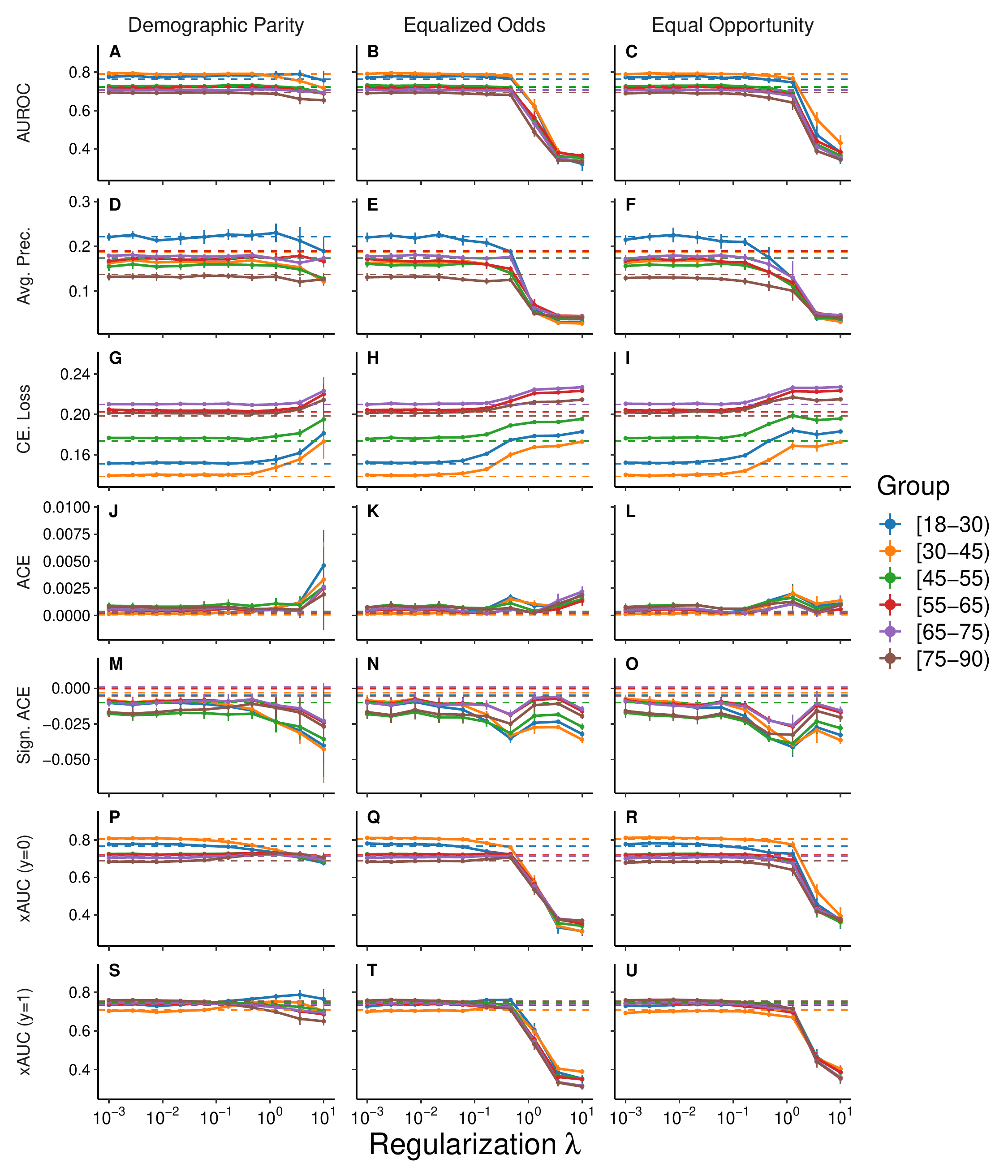}
	\caption{
	    Group-level model performance measures as a function of the extent $\lambda$ that violation of the fairness criterion is penalized when the age group is considered as the sensitive attribute for prediction of 30-day readmission in the STARR database. Results shown are the mean $\pm$ SD for the area under the ROC curve (AUROC), average precision (Avg. Prec), the cross entropy loss (CE Loss), the absolute calibration error (ACE), the signed absolute calibration error (Sign. ACE), and cross group ranking performance (xAUC; $\textrm{xAUC}_k^1$ is indicated by (y=1) and $\textrm{xAUC}_k^0$ by (y=0)) for each group for objectives that penalize violation of threshold-free Demographic Parity, Equalized Odds, and Equal Opportunity with MMD-based penalties. Dashed lines correspond to the mean result for the unpenalized training procedure.
	}
	\label{fig:main_text/starr/all_performance/readmission_30/age_group}
\end{figure*}

\begin{figure*}[!ht]
	\centering
	\includegraphics[width=0.85\linewidth]{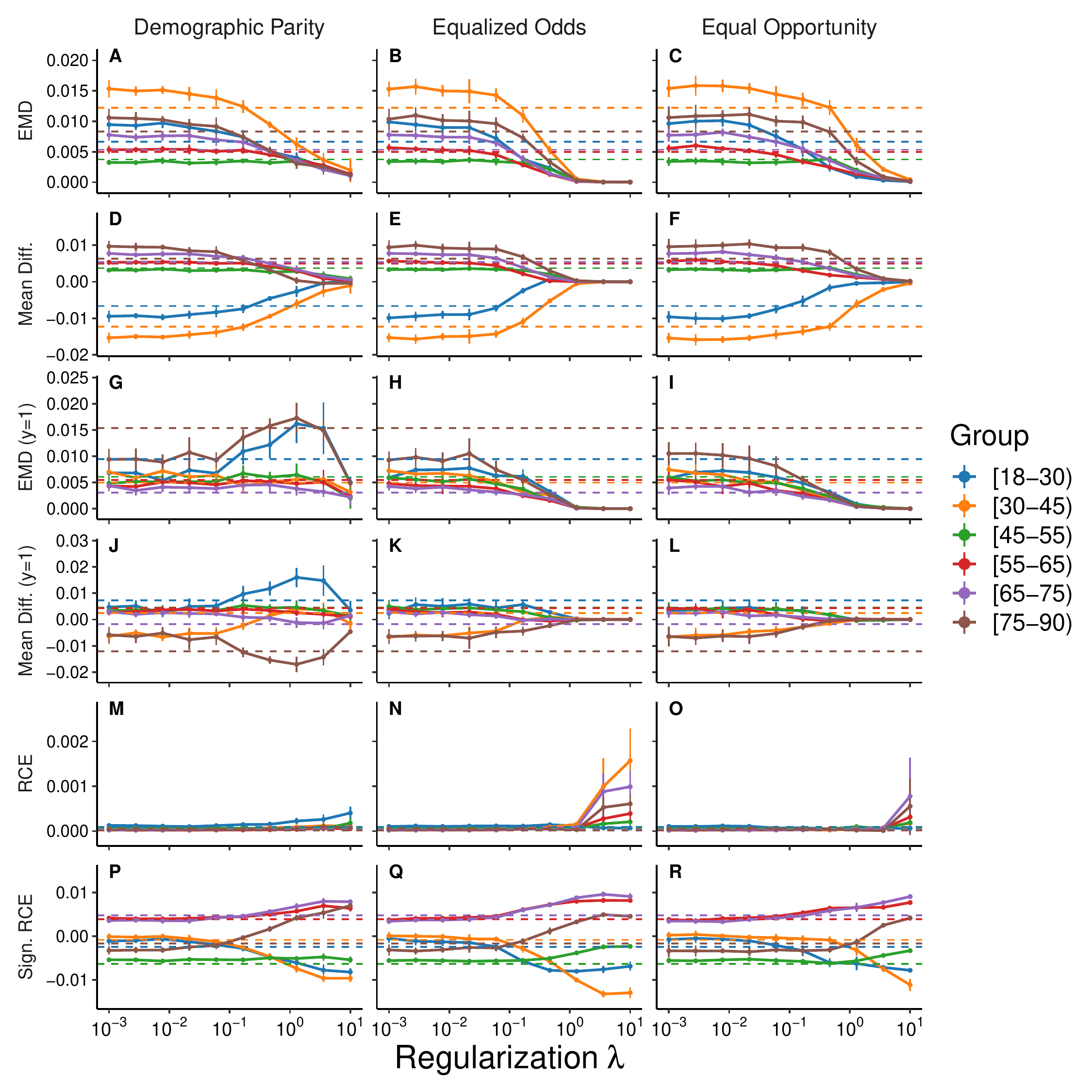}
	\caption{
	    Fairness metrics as a function of the extent $\lambda$ that violation of the fairness criterion is penalized when the age group considered as the sensitive attribute for prediction of 30-day readmission in the STARR database. Results shown are the mean $\pm$ SD for decomposed group-level metrics that assess conditional prediction parity (EMD and Mean Diff.) and relative calibration error (RCE and Sign. RCE) for objectives that penalize violation of threshold-free Demographic Parity, Equalized Odds, and Equal Opportunity on the basis of MMD-based penalties. Measures of conditional prediction parity are separately assessed in the whole population and in the strata for which the outcome is observed (suffixed with (y=1)). Dashed lines correspond to the mean result for the unpenalized training procedure.
	}
	\label{fig:main_text/starr/all_fairness/readmission_30/age_group}
\end{figure*}

Given the breadth of experimentation conducted, and in the interest of brevity, we focus our reporting on general trends that replicate across experimental conditions and on notable exceptions to those trends.
In the main text, we focus on the results for the models derived on STARR, and provide, as examples, figures for models that predict 30-day readmission in the STARR cohort using MMD-based penalties, and exclude results for decomposed fairness metrics computed on the strata of the population for which the relevant outcome is not observed.
Analogous figures for the complete set of experimental conditions and evaluation metrics are provided in the Appendix.

We observe that, in the absence of fairness-promoting regularization, models exhibit substantial differences in group-level model performance measures (AUROC, average precision, and cross entropy loss), and show clear violation of measures of conditional prediction parity as well as differences in cross-group ranking performance (Figures \ref{fig:main_text/starr/all_performance/readmission_30/race_eth}, \ref{fig:main_text/starr/all_fairness/readmission_30/race_eth}, \ref{fig:main_text/starr/all_performance/readmission_30/gender_concept_name}, \ref{fig:main_text/starr/all_fairness/readmission_30/gender_concept_name}, \ref{fig:main_text/starr/all_performance/readmission_30/age_group}, and \ref{fig:main_text/starr/all_fairness/readmission_30/age_group}).
These baseline models tend to have low absolute calibration error for each group, with small differences in relative calibration error across groups.
For example, the model that predicts 30-day readmission in the STARR cohort shows a large degree of variability in AUROC across groups of the race and ethnicity attribute (Mean AUROC: 0.78, 0.66, 0.77, 0.80, 0.71 for the Asian, Black, Hispanic, Other, and White groups, respectively; Figure \ref{fig:main_text/starr/all_performance/readmission_30/race_eth}) and further shows a difference in the mean predicted probability across groups that disproportionately affects the Black population (Figure \ref{fig:main_text/starr/all_fairness/readmission_30/race_eth}). However, the signed absolute and relative calibration errors by group are small (Mean $\textrm{ACE}_k^{\textrm{signed}}$: -0.0067, -0.0074, -0.0021, -0.0059, -0.0021; Mean $\textrm{RCE}_k^{\textrm{signed}}$: -0.0027, -0.0022, 0.0017, -0.0027, 0.00090 for the Asian, Black, Hispanic, Other, and White groups, respectively; Figures \ref{fig:main_text/starr/all_performance/readmission_30/race_eth} and \ref{fig:main_text/starr/all_fairness/readmission_30/race_eth}).

As expected, training with an objective that penalizes violation of a measure of conditional prediction parity typically leads to better satisfaction of the fairness criterion that corresponds to the form of the regularizer used in the objective.
For instance, training with an unconditional penalty to encourage demographic parity typically minimizes the EMD and difference in means between the distribution of predictions for each group and the corresponding marginal distribution constructed via aggregation of the data from all groups.
In some cases, the regularization strategy is less successful at minimizing violation of the targeted fairness criteria, such when using a conditional penalty in the outcome-positive strata to encourage equal opportunity on the basis of sex for the 30-day readmission model in the STARR cohort (Figure  \ref{fig:main_text/starr/all_fairness/readmission_30/gender_concept_name}I).
In this case, the relevant fairness criteria is actually violated to a greater extent when $\lambda = 10$ ($M_{\textrm{EqOpp}}$ = 0.0069) than at baseline ($M_{\textrm{EqOpp}}$ = 0.0056).

We observe a tension between equal opportunity and demographic parity, but these trade-offs are not consistent across experimental conditions.
For instance, conditional penalties in the strata for which hospital mortality is observed in the STARR cohort lead to further violation of demographic parity, relative to baseline, across all three of the sensitive attributes that we test (Supplementary Figures \ref{fig:supplement/starr/all_fairness/hospital_mortality/race_eth}, \ref{fig:supplement/starr/all_fairness/hospital_mortality/gender_concept_name}, and \ref{fig:supplement/starr/all_fairness/hospital_mortality/age_group}). 
In other cases, this penalty actually leads to improved satisfaction of demographic parity, such as in the case of 30-day readmission prediction in the STARR cohort when age is the sensitive attribute ($M_{\textrm{DP}}$: 0.041 for $\lambda=0$ vs. $M_{\textrm{DP}}$: 0.0010 for $\lambda=10$; Figure \ref{fig:main_text/starr/all_fairness/readmission_30/age_group}C), but at the cost of a major reduction in AUROC and Average Precision for all groups (Figure \ref{fig:main_text/starr/all_fairness/readmission_30/age_group}C and \ref{fig:main_text/starr/all_fairness/readmission_30/age_group}F).
A similar phenomenon is observed when considering the impact of unconditional penalties that target demographic parity on metrics that assess equal opportunity in that two criteria appear to coincide in some cases, such as for the model that predicts hospital mortality in the STARR cohort when race and ethnicity is considered the sensitive attribute (Supplementary Figures \ref{fig:supplement/starr/all_fairness/hospital_mortality/race_eth}) and conflict in others, such when sex is considered to be the sensitive attribute for models that predict prolonged length of stay (Supplementary Figure \ref{fig:supplement/optum/all_fairness/LOS_7/gender_concept_name}) or 30-day readmission (Supplementary Figure \ref{fig:supplement/optum/all_fairness/readmission_30/gender_concept_name}) in the Optum CDM cohort.

With few exceptions, the effect of increasing the weight on the conditional regularization penalties that target equalized odds or equal opportunity is a monotonic reduction in group-level model performance measures for all groups. 
In contrast, the effects of unconditional penalties that encourage demographic parity are more heterogeneous.
For instance, while the effect of unconditional penalties on model performance measures that we observe over the trajectory of $\lambda$ are often similar to those that we observe for conditional penalties, we note that unconditional penalties can result in little change in model performance measures (Figures \ref{fig:main_text/starr/all_performance/readmission_30/gender_concept_name} and  \ref{fig:main_text/starr/all_performance/readmission_30/age_group}), and in some cases, actually results in improved model performance for one or more groups relative to baseline
(Figure \ref{fig:main_text/starr/all_performance/readmission_30/race_eth} and Supplementary Figures \ref{fig:supplement/starr/all_performance/hospital_mortality/race_eth} and \ref{fig:supplement/starr/all_performance/hospital_mortality/gender_concept_name}).
For example, penalizing the violation of demographic parity increases the performance of the model that predicts 30-day readmission for the Black population when $\lambda = 3.6$ compared to baseline (AUROC: 0.69 vs. 0.66; Average Precision: 0.17 vs. 0.15) (Figures \ref{fig:main_text/starr/all_performance/readmission_30/race_eth}A and \ref{fig:main_text/starr/all_performance/readmission_30/race_eth}D).

In general, models become less well-calibrated, in the absolute sense, at the group level, as the weight on either conditional penalty increases, as measured by changes in the ACE or signed ACE relative to baseline.
In many cases, unconditional penalties seem to have little impact on group-level model calibration relative to that which is observed for conditional penalties (Figure \ref{fig:main_text/starr/all_performance/readmission_30/gender_concept_name}), whereas in other cases the effect of the unconditional penalty is similar to that of the conditional penalties (Figure \ref{fig:main_text/starr/all_performance/readmission_30/age_group}).
However, both unconditional and conditional penalties can, but do not always, introduce relative calibration error across groups, and in a way that appears unrelated to the changes to absolute calibration.
For example, for models that predict hospital mortality in the STARR cohort when age is the sensitive attribute, the magnitude of the effect on absolute calibration differs substantially on the basis of the type of regularization applied (Supplementary Figure \ref{fig:supplement/starr/all_performance/hospital_mortality/age_group}), while the effect on relative calibration is similar across all penalties (Supplementary Figure \ref{fig:supplement/starr/all_fairness/hospital_mortality/age_group}).
In cases where the effects on relative calibration are large, such as this set of models, the impact of the effect on relative calibration concentrates in relatively few groups (Supplementary Figure \ref{fig:supplement/starr/all_fairness/hospital_mortality/age_group}).

We observe heterogeneity in the manner in which training with fairness-promoting objectives impacts measures of cross-group ranking across the combinations of regularizer, dataset, outcome, and sensitive attribute.
In many respects, the trajectories of xAUC measures are similar to those that we generally observe for the AUROC, in that primary effect that we observe is a decline in cross-group ranking accuracy as a function of $\lambda$, regardless of the type of penalty selected (Figures \ref{fig:main_text/starr/all_performance/readmission_30/race_eth}Q, \ref{fig:main_text/starr/all_performance/readmission_30/race_eth}R, \ref{fig:main_text/starr/all_performance/readmission_30/race_eth}T,
and \ref{fig:main_text/starr/all_performance/readmission_30/race_eth}U).
However, in some cases, the trajectories of xAUC measures are convergent in a way that both improves fairness and allows for an improvement in the measure for at least one group at the expense of one or more other groups. 
In some cases, we observe this effect only for unconditional penalties (Supplementary Figures \ref{fig:supplement/starr/all_performance/hospital_mortality/race_eth}, \ref{fig:supplement/starr/all_performance/hospital_mortality/gender_concept_name}, and \ref{fig:supplement/starr/all_performance/hospital_mortality/age_group}) and in others, we observe it for both unconditional and conditional penalties (Supplementary Figures \ref{fig:supplement/starr/all_performance/LOS_7/gender_concept_name} and \ref{fig:supplement/starr/all_performance/LOS_7/age_group}).

\section{Discussion} \label{sec:discussion}
Our experiments aim to provide a comprehensive empirical evaluation of the effect of penalizing group fairness criteria violations on measures of model performance and group fairness for clinical predictive models.
Our results reveal substantial heterogeneity in the effect of imposing measures of group fairness across datasets, outcomes, sensitive attribute and group definitions, and regularization strategies.
These results quantify the extent to which the trade-offs among measures of model performance and group fairness described by ``impossibility theorems'' \cite{Chouldechova2017,Kleinberg2016,pleiss2017fairness,Corbett-Davies:2017:ADM:3097983.3098095} manifest when learning clinical predictive models from real-world databases.

We acknowledge technical limitations of our work that may limit the generalizability of our results. First, the regularizers and metrics used to quantify conditional prediction parity and relative calibration are the result of ``one vs. marginal'' comparisons where a measure computed for one group is compared to the measure computed for the aggregate population.
This choice is one of several ways to construct fairness metrics, including ``one vs. other'' comparisons between one group and all other groups and pairwise comparisons across all pairs of groups.
An effect of this choice is that metrics that assess violation of conditional prediction parity for an over-represented group are more likely to be small since the over-represented group comprises a larger fraction of the population than under-represented groups do.
Furthermore, our use of penalized objectives could exaggerate the extent of the reported trade-offs, relative to the alternative of a Lagrangian formulation that directly encodes the fairness criteria as a constraint \cite{Agarwal2018,pmlr-v54-zafar17a,Cotter2018,Cotter2019,Celis2018}.
While the constrained approach typically only provides guarantees of constraint satisfaction in the case of a convex objective, recent work has demonstrated empirical success with a modified proxy-Lagrangian formulation that is effective for non-convex constrained optimization problems \cite{Cotter2018}.
It remains to be seen whether reformulating the problem as constrained optimization allows for satisfaction of fairness constraints with less severe trade-offs than those reported here.

Our work inherits the fundamental limitations of the group fairness framework and of algorithmic fairness more broadly. The group fairness framework, which arose from legal notions of anti-discrimination, reinforces a perspective that groups based on categorical attributes are well-defined constructs that correspond to a set of homogeneous populations -- a perspective that has several problematic implications.
For example, the definitions of racial categories are entangled with historical and on-going patterns of structural racism, and their continued use reinforces the idea of race as an accurate way to describe human variability, rather than a socially constructed taxonomy \cite{Hanna2020,Sewell2016,Bailey2017,Hicken2018,Vyas2020,VanderWeele2014,Duster2005,Braun2007}.
This framework further marginalizes groups that are not well-represented by the attributes used to assess group fairness, including intersectional identities \cite{crenshaw1989demarginalizing,Hoffmann2019,Hanna2020,Kearns2017,pmlr-v80-hebert-johnson18a}.
Furthermore, in addressing each attribute independently, the group fairness framework treats various sensitive attributes as abstract, interchangeable constructs, without awareness of meaningful contextual differences between them.
For example, while observed differences on the basis of race should be primarily interpreted as deriving from systemic and structurally racist factors \cite{Hanna2020,Sewell2016,Bailey2017,Hicken2018,Vyas2020,VanderWeele2014}, those observed for sex could potentially be attributed to clinically meaningful differences in human physiology as well as sociological factors \cite{Cirillo2020}.

Alternative forms of algorithmic fairness raise additional normative questions that require context-specific judgement and domain knowledge. Individual fairness measures require robustness over a metric space that encodes domain-specific norms, which concern how outputs of an algorithm may change over the space of observed covariates \cite{Dwork2011}. Counterfactual fairness provides a particular instantiation of individual fairness, defining closeness in the domain-specific metric in terms of counterfactuals with respect to a sensitive attribute \cite{Kusner2017,Pfohl2019MLHC}. 
However, this requires specification of the causal pathways between the sensitive attribute, outcomes, and discrimination. It further assumes manipulability of sensitive attributes within the context of a well-defined structural equation model, which is particularly unrealistic for complex high-dimensional data and contestable whenever race is considered to be the sensitive attribute \cite{VanderWeele2014,Hanna2020}.

\subsection{Recommendations} \label{subsec:recommendations}
Striving for health equity requires designing policies that directly counteract the systemic factors that contribute to health disparities, primarily structural forms of racism and economic inequality \cite{Ford2010,Bailey2017}.
By considering only changes to observable properties of a model, evaluations of models using the group fairness framework ignore the inequities in the data generating and measurement processes. They also miss the decision-theoretic and causal framing necessary to connect predictions to the interventions they trigger, as well as the resulting downstream effect on health disparities \cite{Goodman2018,Fazelpour2020,kilbertus2017avoiding}.
In the absence of this context, a requirement that a predictive model satisfy a notion of group fairness provides little more than a ``veneer of neutrality'' \cite{Benjamin421,McCradden2020b}.
Overall, constraining a model such that group fairness is achieved is insufficient for, and may actively work against, the goal of promoting health equity using machine learning guided interventions. 
To be sure, this lack of sufficiency does not mean that explicitly optimizing for fairness criteria satisfaction can not be useful. However, the value of achieving algorithmic fairness should be defined in terms of the impact of an algorithm-guided intervention on individuals, groups, and on status quo power structures that directly or indirectly perpetuate health disparities \cite{Kalluri2020}.

In light of these limitations, model developers in healthcare should engage in participatory design practices that explicitly incorporate perspectives from a diverse set of stakeholders, including patient advocacy groups and civil society organizations. 
Doing so is necessary to identify mechanisms through which measurement error, bias, and historical inequities affect data collection, measurement, and problem formulation, as well as to reason about the mechanisms by which the intervention triggered by the model's prediction interacts with those factors  \cite{Sendak2020,Martin2020,Martin2020a,McCradden2020a}.
However, it is important to allow that the conclusion derived from this process may be to abstain from algorithm-aided decision making entirely if it is not practical to do so responsibly \cite{selbst2019fairness,Baumer2011}.

\section{Conclusion}
The debate on the use of algorithmic fairness techniques in healthcare has largely proceeded without empirical characterization of the effects of these techniques on the properties of predictive models derived from large-scale clinical data.
We explicitly measure and comprehensively report on the extent of the empirical trade-offs between measures of model performance and notions of group fairness such as conditional prediction parity, relative calibration, and cross-group ranking. 
These constructs are generally well-understood in theoretical contexts, but under-explored in the context of clinical predictive models.
Given our results, and the known limitations of the algorithmic fairness framework, we recommend that the use of algorithmic fairness for proactive monitoring and auditing of a clinical predictive model proceed only if measures of model performance and fairness can be appropriately contextualized in terms of the impact of the complex intervention that the model enables.

\section{Acknowledgements and Funding}
We thank Conor Corbin, Jonathan Lu, Jennifer Wilson, Alison Callahan, Ethan Steinberg, Jason Fries, Erin Craig, Scott Fleming, Mila Hardt, and Steve Yadlowsky for insightful feedback and discussion. 
Data access for this project was provided by the Stanford Center for Population Health Sciences Data Core, the Stanford School of Medicine Research Office, and the Stanford Medicine Research IT team.
We further thank the Stanford Medicine Research IT team for supporting STARR-OMOP, with special thanks to Jose Posada and Priya Desai, and the Stanford Research Computing Center  for supporting the Nero computing platform.
Approval for this non-human subjects research study is provided by the Stanford Institutional Review Board, protocol 24883.
This work is supported by the National Science Foundation Graduate Research Fellowship Program DGE-1656518 and the Stanford Medicine Program for AI in Healthcare.
Any opinions, findings, and conclusions or recommendations expressed in this material are those of the authors and do not necessarily reflect the views of the funding bodies.

\bibliographystyle{elsarticle-num}
\bibliography{references}

\begin{thebibliography}{100}
\expandafter\ifx\csname url\endcsname\relax
  \def\url#1{\texttt{#1}}\fi
\expandafter\ifx\csname urlprefix\endcsname\relax\def\urlprefix{URL }\fi
\expandafter\ifx\csname href\endcsname\relax
  \def\href#1#2{#2} \def\path#1{#1}\fi

\bibitem{Rajkomar2018}
A.~Rajkomar, M.~Hardt, M.~D. Howell, G.~Corrado, M.~H. Chin,
  \href{http://annals.org/article.aspx?doi=10.7326/M18-1990}{{Ensuring Fairness
  in Machine Learning to Advance Health Equity}}, Annals of Internal
  Medicine\href {http://dx.doi.org/10.7326/M18-1990}
  {\path{doi:10.7326/M18-1990}}.
\newline\urlprefix\url{http://annals.org/article.aspx?doi=10.7326/M18-1990}

\bibitem{Goodman2018}
S.~N. Goodman, S.~Goel, M.~R. Cullen,
  \href{http://annals.org/article.aspx?doi=10.7326/M18-3297}{{Machine Learning,
  Health Disparities, and Causal Reasoning}}, Annals of Internal Medicine
  169~(12) (2018) 883.
\newblock \href {http://dx.doi.org/10.7326/M18-3297}
  {\path{doi:10.7326/M18-3297}}.
\newline\urlprefix\url{http://annals.org/article.aspx?doi=10.7326/M18-3297}

\bibitem{Obermeyer2019}
Z.~Obermeyer, B.~Powers, C.~Vogeli, S.~Mullainathan, {Dissecting racial bias in
  an algorithm used to manage the health of populations}, Science 366~(6464)
  (2019) 447--453.
\newblock \href {http://dx.doi.org/10.1126/SCIENCE.AAX2342}
  {\path{doi:10.1126/SCIENCE.AAX2342}}.

\bibitem{Ferryman2018}
K.~Ferryman, M.~Pitcan, Fairness in precision medicine, Data \& Society.

\bibitem{Nordling2019}
L.~Nordling, {A fairer way forward for AI in health care.}, Nature 573~(7775)
  (2019) S103.

\bibitem{Vyas2020}
D.~A. Vyas, L.~G. Eisenstein, D.~S. Jones,
  \href{http://www.nejm.org/doi/10.1056/NEJMms2004740}{{Hidden in Plain Sight
  — Reconsidering the Use of Race Correction in Clinical Algorithms}}, New
  England Journal of Medicine (2020) NEJMms2004740\href
  {http://dx.doi.org/10.1056/NEJMms2004740} {\path{doi:10.1056/NEJMms2004740}}.
\newline\urlprefix\url{http://www.nejm.org/doi/10.1056/NEJMms2004740}

\bibitem{Chen2020}
I.~Y. Chen, S.~Joshi, M.~Ghassemi,
  \href{http://www.nature.com/articles/s41591-019-0649-2}{{Treating health
  disparities with artificial intelligence}}, Nature Medicine 26~(1) (2020)
  16--17.
\newblock \href {http://dx.doi.org/10.1038/s41591-019-0649-2}
  {\path{doi:10.1038/s41591-019-0649-2}}.
\newline\urlprefix\url{http://www.nature.com/articles/s41591-019-0649-2}

\bibitem{Gaskin2012}
D.~J. Gaskin, G.~Y. Dinwiddie, K.~S. Chan, R.~McCleary, {Residential
  segregation and disparities in health care services utilization.}, Medical
  care research and review : MCRR 69~(2) (2012) 158--75.
\newblock \href {http://dx.doi.org/10.1177/1077558711420263}
  {\path{doi:10.1177/1077558711420263}}.

\bibitem{Williams2001}
D.~R. Williams, C.~Collins, {Racial residential segregation: a fundamental
  cause of racial disparities in health.}, Public Health Reports 116~(5) (2001)
  404.
\newblock \href {http://dx.doi.org/10.1093/PHR/116.5.404}
  {\path{doi:10.1093/PHR/116.5.404}}.

\bibitem{Hall2015}
W.~J. Hall, M.~V. Chapman, K.~M. Lee, Y.~M. Merino, T.~W. Thomas, B.~K. Payne,
  E.~Eng, S.~H. Day, T.~Coyne-Beasley, {Implicit Racial/Ethnic Bias Among
  Health Care Professionals and Its Influence on Health Care Outcomes: A
  Systematic Review.}, American journal of public health 105~(12) (2015)
  e60--76.
\newblock \href {http://dx.doi.org/10.2105/AJPH.2015.302903}
  {\path{doi:10.2105/AJPH.2015.302903}}.

\bibitem{Bailey2017}
Z.~D. Bailey, N.~Krieger, M.~Ag{\'{e}}nor, J.~Graves, N.~Linos, M.~T. Bassett,
  \href{https://www-thelancet-com/series/america-equity-equality-in-health}{{Structural
  racism and health inequities in the USA: evidence and interventions}}, The
  Lancet 389~(10077) (2017) 1453--1463.
\newblock \href {http://dx.doi.org/10.1016/S0140-6736(17)30569-X}
  {\path{doi:10.1016/S0140-6736(17)30569-X}}.
\newline\urlprefix\url{https://www-thelancet-com/series/america-equity-equality-in-health}

\bibitem{Larrazabal2020}
A.~J. Larrazabal, N.~Nieto, V.~Peterson, D.~H. Milone, E.~Ferrante, {Gender
  imbalance in medical imaging datasets produces biased classifiers for
  computer-aided diagnosis}, Proceedings of the National Academy of
  Sciences\href {http://dx.doi.org/10.1073/PNAS.1919012117}
  {\path{doi:10.1073/PNAS.1919012117}}.

\bibitem{Kallus2018}
N.~Kallus, A.~Zhou, {Residual unfairness in fair machine learning from
  prejudiced data}, 35th International Conference on Machine Learning, ICML
  2018 6 (2018) 3821--3834.
\newblock \href {http://arxiv.org/abs/1806.02887} {\path{arXiv:1806.02887}}.

\bibitem{Jiang2019}
H.~Jiang, O.~Nachum, {Identifying and Correcting Label Bias in Machine
  Learning}, International Conference on Artificial Intelligence and Statistics
  (2020) 702--712\href {http://arxiv.org/abs/1901.04966}
  {\path{arXiv:1901.04966}}.

\bibitem{Veinot2018}
T.~C. Veinot, H.~Mitchell, J.~S. Ancker,
  \href{https://academic.oup.com/jamia/advance-article/doi/10.1093/jamia/ocy052/4996916}{{Perspective
  Good intentions are not enough: how informatics interventions can worsen
  inequality}}, Journal of the American Medical Informatics Association 25~(8)
  (2018) 1080--1088.
\newblock \href {http://dx.doi.org/10.1093/jamia/ocy052}
  {\path{doi:10.1093/jamia/ocy052}}.
\newline\urlprefix\url{https://academic.oup.com/jamia/advance-article/doi/10.1093/jamia/ocy052/4996916}

\bibitem{McCradden2020}
M.~McCradden, M.~Mazwi, S.~Joshi, J.~A. Anderson,
  \href{http://dl.acm.org/doi/10.1145/3375627.3375824}{{When Your Only Tool Is
  A Hammer}}, in: Proceedings of the AAAI/ACM Conference on AI, Ethics, and
  Society, ACM, New York, NY, USA, 2020, pp. 109--109.
\newblock \href {http://dx.doi.org/10.1145/3375627.3375824}
  {\path{doi:10.1145/3375627.3375824}}.
\newline\urlprefix\url{http://dl.acm.org/doi/10.1145/3375627.3375824}

\bibitem{McCradden2020b}
M.~D. McCradden, S.~Joshi, J.~A. Anderson, M.~Mazwi, A.~Goldenberg, R.~{Zlotnik
  Shaul}, \href{https://doi.org/10.1093/jamia/ocaa085}{{Patient safety and
  quality improvement: Ethical principles for a regulatory approach to bias in
  healthcare machine learning}}, Journal of the American Medical Informatics
  Association\href {http://dx.doi.org/10.1093/jamia/ocaa085}
  {\path{doi:10.1093/jamia/ocaa085}}.
\newline\urlprefix\url{https://doi.org/10.1093/jamia/ocaa085}

\bibitem{Char2018}
D.~S. Char, N.~H. Shah, D.~Magnus,
  \href{http://www.nejm.org/doi/10.1056/NEJMp1714229}{{Implementing Machine
  Learning in Health Care — Addressing Ethical Challenges}}, New England
  Journal of Medicine 378~(11) (2018) 981--983.
\newblock \href {http://dx.doi.org/10.1056/NEJMp1714229}
  {\path{doi:10.1056/NEJMp1714229}}.
\newline\urlprefix\url{http://www.nejm.org/doi/10.1056/NEJMp1714229}

\bibitem{Parikh2019}
R.~B. Parikh, S.~Teeple, A.~S. Navathe,
  \href{https://jamanetwork.com/journals/jama/fullarticle/2756196}{{Addressing
  Bias in Artificial Intelligence in Health Care}}, JAMA 170~(1) (2019) 51--58.
\newblock \href {http://dx.doi.org/10.1001/jama.2019.18058}
  {\path{doi:10.1001/jama.2019.18058}}.
\newline\urlprefix\url{https://jamanetwork.com/journals/jama/fullarticle/2756196}

\bibitem{McCradden2020a}
M.~D. McCradden, S.~Joshi, M.~Mazwi, J.~A. Anderson, {Ethical limitations of
  algorithmic fairness solutions in health care machine learning}, The Lancet
  Digital Health 2~(5) (2020) e221--e223.
\newblock \href {http://dx.doi.org/10.1016/S2589-7500(20)30065-0}
  {\path{doi:10.1016/S2589-7500(20)30065-0}}.

\bibitem{Dwork2011}
C.~Dwork, M.~Hardt, T.~Pitassi, O.~Reingold, R.~Zemel, {Fairness Through
  Awareness}, Proceedings of the 3rd innovations in theoretical computer
  science conference (2011) 214--226\href {http://arxiv.org/abs/1104.3913}
  {\path{arXiv:1104.3913}}.

\bibitem{Hardt2016}
M.~Hardt, E.~Price, N.~Srebro, {Equality of Opportunity in Supervised
  Learning}, Advances in Neural Information Processing Systems (2016)
  3315--3323\href {http://arxiv.org/abs/1610.02413} {\path{arXiv:1610.02413}},
  \href {http://dx.doi.org/10.1109/ICCV.2015.169}
  {\path{doi:10.1109/ICCV.2015.169}}.

\bibitem{Chouldechova2018}
A.~Chouldechova, A.~Roth, {The Frontiers of Fairness in Machine Learning},
  arXiv preprint arXiv:1810.08810\href {http://arxiv.org/abs/1810.08810}
  {\path{arXiv:1810.08810}}, \href {http://dx.doi.org/10.17226/25021}
  {\path{doi:10.17226/25021}}.

\bibitem{Green2020}
B.~Green, {The false promise of risk assessments: epistemic reform and the
  limits of fairness}, in: Proceedings of the 2020 Conference on Fairness,
  Accountability, and Transparency, 2020, pp. 594--606.

\bibitem{Hutchinson2019}
B.~Hutchinson, M.~Mitchell, \href{https://doi.org/10.1145/3287560.3287600}{{50
  years of test (un) fairness: Lessons for machine learning}}, in: Proceedings
  of the Conference on Fairness, Accountability, and Transparency, 2019, pp.
  49--58.
\newblock \href {http://arxiv.org/abs/1811.10104v2}
  {\path{arXiv:1811.10104v2}}, \href
  {http://dx.doi.org/10.1145/3287560.3287600}
  {\path{doi:10.1145/3287560.3287600}}.
\newline\urlprefix\url{https://doi.org/10.1145/3287560.3287600}

\bibitem{aif360-oct-2018}
R.~K.~E. Bellamy, K.~Dey, M.~Hind, S.~C. Hoffman, S.~Houde, K.~Kannan,
  P.~Lohia, J.~Martino, S.~Mehta, A.~Mojsilovic, S.~Nagar, K.~N. Ramamurthy,
  J.~Richards, D.~Saha, P.~Sattigeri, M.~Singh, K.~R. Varshney, Y.~Zhang,
  {{\{}AI Fairness{\}} 360: An Extensible Toolkit for Detecting, Understanding,
  and Mitigating Unwanted Algorithmic Bias} (oct 2018).

\bibitem{Dudik2020}
M.~Dudik, R.~Edgar, B.~Horn, R.~Lutz,
  \href{https://github.com/fairlearn/fairlearn}{{Fairlearn}} (2020).
\newline\urlprefix\url{https://github.com/fairlearn/fairlearn}

\bibitem{google_fairness}
\href{https://ai.googleblog.com/2019/12/fairness-indicators-scalable.html}{{Google
  AI Blog: Fairness Indicators: Scalable Infrastructure for Fair ML Systems}}.
\newline\urlprefix\url{https://ai.googleblog.com/2019/12/fairness-indicators-scalable.html}

\bibitem{Pfohl2019MLHC}
S.~R. Pfohl, T.~Duan, D.~Y. Ding, N.~H. Shah,
  \href{http://proceedings.mlr.press/v106/pfohl19a.html}{{Counterfactual
  Reasoning for Fair Clinical Risk Prediction}}, in: F.~Doshi-Velez,
  J.~Fackler, K.~Jung, D.~Kale, R.~Ranganath, B.~Wallace, J.~Wiens (Eds.),
  Proceedings of the 4th Machine Learning for Healthcare Conference, Vol. 106
  of Proceedings of Machine Learning Research, PMLR, Ann Arbor, Michigan, 2019,
  pp. 325--358.
\newblock \href {http://arxiv.org/abs/1907.06260} {\path{arXiv:1907.06260}}.
\newline\urlprefix\url{http://proceedings.mlr.press/v106/pfohl19a.html}

\bibitem{Pfohl2019AIES}
S.~Pfohl, B.~Marafino, A.~Coulet, F.~Rodriguez, L.~Palaniappan, N.~H. Shah,
  {Creating Fair Models of Atherosclerotic Cardiovascular Disease Risk}, in:
  AAAI/ACM Conference on Artificial Intelligence, Ethics, and Society, 2019.
\newblock \href {http://arxiv.org/abs/1809.04663} {\path{arXiv:1809.04663}}.

\bibitem{Zink2019}
A.~Zink, S.~Rose, {Fair regression for health care spending}, Biometrics\href
  {http://arxiv.org/abs/1901.10566} {\path{arXiv:1901.10566}}, \href
  {http://dx.doi.org/10.1111/biom.13206} {\path{doi:10.1111/biom.13206}}.

\bibitem{Zhang2020}
H.~Zhang, A.~X. Lu, M.~Abdalla, M.~McDermott, M.~Ghassemi,
  \href{https://dl.acm.org/doi/10.1145/3368555.3384448}{{Hurtful words:
  quantifying biases in clinical contextual word embeddings}}, in: ACM CHIL
  2020 - Proceedings of the 2020 ACM Conference on Health, Inference, and
  Learning, Association for Computing Machinery (ACM), New York, NY, USA, 2020,
  pp. 110--120.
\newblock \href {http://dx.doi.org/10.1145/3368555.3384448}
  {\path{doi:10.1145/3368555.3384448}}.
\newline\urlprefix\url{https://dl.acm.org/doi/10.1145/3368555.3384448}

\bibitem{Singh2019a}
M.~Singh, K.~N. Ramamurthy, {Understanding racial bias in health using the
  Medical Expenditure Panel Survey data}, arXiv preprint arXiv:1911.01509\href
  {http://arxiv.org/abs/1911.01509} {\path{arXiv:1911.01509}}.

\bibitem{Singh2019b}
H.~Singh, R.~Singh, V.~Mhasawade, R.~Chunara, {Fair Predictors under
  Distribution Shift}, arXiv preprint arXiv:1911.00677\href
  {http://arxiv.org/abs/1911.00677} {\path{arXiv:1911.00677}}.

\bibitem{Zemel2013}
R.~S. Zemel, Y.~Wu, K.~Swersky, T.~Pitassi, C.~Dwork, {Learning Fair
  Representations}, Proceedings of the 30th International Conference on Machine
  Learning 28 (2013) 325--333.

\bibitem{Cotter2018}
A.~Cotter, H.~Jiang, M.~R. Gupta, S.~Wang, T.~Narayan, S.~You, K.~Sridharan,
  M.~R. Gupta, S.~You, K.~Sridharan, {Optimization with Non-Differentiable
  Constraints with Applications to Fairness, Recall, Churn, and Other Goals.},
  Journal of Machine Learning Research 20~(172) (2019) 1--59.
\newblock \href {http://arxiv.org/abs/1809.04198} {\path{arXiv:1809.04198}}.

\bibitem{Cotter2019}
A.~Cotter, M.~Gupta, H.~Jiang, N.~Srebro, K.~Sridharan, S.~Wang, B.~Woodworth,
  S.~You, {Training Well-Generalizing Classifiers for Fairness Metrics and
  Other Data-Dependent Constraints}, in: International Conference on Machine
  Learning, 2019, pp. 1397--1405.
\newblock \href {http://arxiv.org/abs/1807.00028} {\path{arXiv:1807.00028}}.

\bibitem{Agarwal2018}
A.~Agarwal, A.~Beygelzimer, M.~Dudik, J.~Langford, H.~Wallach, A reductions
  approach to fair classification, in: International Conference on Machine
  Learning, 2018, pp. 60--69.

\bibitem{Song2019}
J.~Song, P.~Kalluri, A.~Grover, S.~Zhao, S.~Ermon, {Learning Controllable Fair
  Representations}, in: The 22nd International Conference on Artificial
  Intelligence and Statistics, 2019, pp. 2164--2173.
\newblock \href {http://arxiv.org/abs/1812.04218} {\path{arXiv:1812.04218}}.

\bibitem{Mitchell2019}
M.~Mitchell, S.~Wu, A.~Zaldivar, P.~Barnes, L.~Vasserman, B.~Hutchinson,
  E.~Spitzer, I.~D. Raji, T.~Gebru, Model cards for model reporting, in:
  Proceedings of the conference on fairness, accountability, and transparency,
  2019, pp. 220--229.

\bibitem{Sun2019}
C.~Sun, A.~Asudeh, H.~V. Jagadish, B.~Howe, J.~Stoyanovich,
  \href{https://dl.acm.org/doi/10.1145/3357384.3357853}{{Mithralabel: Flexible
  dataset nutritional labels for responsible data science}}, in: International
  Conference on Information and Knowledge Management, Proceedings, Association
  for Computing Machinery, New York, NY, USA, 2019, pp. 2893--2896.
\newblock \href {http://dx.doi.org/10.1145/3357384.3357853}
  {\path{doi:10.1145/3357384.3357853}}.
\newline\urlprefix\url{https://dl.acm.org/doi/10.1145/3357384.3357853}

\bibitem{Madaio2020}
M.~A. Madaio, L.~Stark, J.~{Wortman Vaughan}, H.~Wallach,
  \href{https://dl.acm.org/doi/10.1145/3313831.3376445}{{Co-Designing
  Checklists to Understand Organizational Challenges and Opportunities around
  Fairness in AI}}, in: Proceedings of the 2020 CHI Conference on Human Factors
  in Computing Systems, ACM, New York, NY, USA, 2020, pp. 1--14.
\newblock \href {http://dx.doi.org/10.1145/3313831.3376445}
  {\path{doi:10.1145/3313831.3376445}}.
\newline\urlprefix\url{https://dl.acm.org/doi/10.1145/3313831.3376445}

\bibitem{ChenJohanssonSontag_NIPS18}
I.~Chen, F.~D. Johansson, D.~Sontag, {Why Is My Classifier Discriminatory?},
  Proceedings of the 32nd International Conference on Neural Information
  Processing Systems\href {http://arxiv.org/abs/1805.12002}
  {\path{arXiv:1805.12002}}.

\bibitem{Corbett-Davies2018}
S.~Corbett-Davies, S.~Goel, {The Measure and Mismeasure of Fairness: A Critical
  Review of Fair Machine Learning}, arXiv preprint arXiv:1808.00023\href
  {http://arxiv.org/abs/1808.00023} {\path{arXiv:1808.00023}}, \href
  {http://dx.doi.org/10.1063/1.3627170} {\path{doi:10.1063/1.3627170}}.

\bibitem{Fazelpour2020}
S.~Fazelpour, Z.~C. Lipton, {Algorithmic Fairness from a Non-ideal
  Perspective}, in: Proceedings of the AAAI/ACM Conference on AI, Ethics, and
  Society, 2020, pp. 57--63.
\newblock \href {http://arxiv.org/abs/2001.09773} {\path{arXiv:2001.09773}},
  \href {http://dx.doi.org/10.1145/3375627.3375828}
  {\path{doi:10.1145/3375627.3375828}}.

\bibitem{Herington2020}
J.~Herington, \href{https://doi.org/10.1145/3375627.3375854}{Measuring fairness
  in an unfair world}, in: Proceedings of the AAAI/ACM Conference on AI,
  Ethics, and Society, AIES ’20, Association for Computing Machinery, New
  York, NY, USA, 2020, p. 286–292.
\newblock \href {http://dx.doi.org/10.1145/3375627.3375854}
  {\path{doi:10.1145/3375627.3375854}}.
\newline\urlprefix\url{https://doi.org/10.1145/3375627.3375854}

\bibitem{Liu2018}
L.~T. Liu, S.~Dean, E.~Rolf, M.~Simchowitz, M.~Hardt, {Delayed Impact of Fair
  Machine Learning}, in: Proceedings of the 35th International Conference on
  Machine Learning, 2018.
\newblock \href {http://arxiv.org/abs/1803.04383} {\path{arXiv:1803.04383}}.

\bibitem{Hanna2020}
A.~Hanna, E.~Denton, A.~Smart, J.~Smith-Loud, {Towards a critical race
  methodology in algorithmic fairness}, in: Proceedings of the 2020 Conference
  on Fairness, Accountability, and Transparency, 2020, pp. 501--512.
\newblock \href {http://arxiv.org/abs/1912.03593} {\path{arXiv:1912.03593}},
  \href {http://dx.doi.org/10.1145/3351095.3372826}
  {\path{doi:10.1145/3351095.3372826}}.

\bibitem{Jacobs2019}
A.~Z. Jacobs, H.~Wallach, {Measurement and Fairness}, arXiv preprint
  arXiv:1912.05511\href {http://arxiv.org/abs/1912.05511}
  {\path{arXiv:1912.05511}}.

\bibitem{Hicken2018}
M.~T. Hicken, N.~Kravitz-Wirtz, M.~Durkee, J.~S. Jackson, {Racial inequalities
  in health: Framing future research}, Social Science and Medicine 199 (2018)
  11--18.
\newblock \href {http://dx.doi.org/10.1016/j.socscimed.2017.12.027}
  {\path{doi:10.1016/j.socscimed.2017.12.027}}.

\bibitem{vitale2017under}
C.~Vitale, M.~Fini, I.~Spoletini, M.~Lainscak, P.~Seferovic, G.~M. Rosano,
  Under-representation of elderly and women in clinical trials, International
  journal of cardiology 232 (2017) 216--221.

\bibitem{hussain2004ethnic}
M.~Hussain-Gambles, K.~Atkin, B.~Leese, Why ethnic minority groups are
  under-represented in clinical trials: a review of the literature, Health \&
  social care in the community 12~(5) (2004) 382--388.

\bibitem{dickman2017inequality}
S.~L. Dickman, D.~U. Himmelstein, S.~Woolhandler, Inequality and the
  health-care system in the usa, The Lancet 389~(10077) (2017) 1431--1441.

\bibitem{Shah2019}
N.~H. Shah, A.~Milstein, S.~C. Bagley, {Making Machine Learning Models
  Clinically Useful}, JAMA - Journal of the American Medical Association
  322~(14) (2019) 1351--1352.
\newblock \href {http://dx.doi.org/10.1001/jama.2019.10306}
  {\path{doi:10.1001/jama.2019.10306}}.

\bibitem{Jung2020}
K.~Jung, S.~Kashyap, A.~Avati, S.~Harman, H.~Shaw, R.~Li, M.~Smith, K.~Shum,
  J.~Javitz, Y.~Vetteth, T.~Seto, S.~C. Bagley, N.~H. Shah,
  \href{https://doi.org/10.1101/2020.07.10.20149419}{{A framework for making
  predictive models useful in practice}}, medRxiv (2020)
  2020.07.10.20149419\href {http://dx.doi.org/10.1101/2020.07.10.20149419}
  {\path{doi:10.1101/2020.07.10.20149419}}.
\newline\urlprefix\url{https://doi.org/10.1101/2020.07.10.20149419}

\bibitem{Creager2019}
E.~Creager, D.~Madras, T.~Pitassi, R.~Zemel, {Causal Modeling for Fairness in
  Dynamical Systems}, arXiv preprint arXiv:1909.09141\href
  {http://arxiv.org/abs/1909.09141} {\path{arXiv:1909.09141}}.

\bibitem{Kleinberg2016}
J.~Kleinberg, S.~Mullainathan, M.~Raghavan, {Inherent Trade-Offs in the Fair
  Determination of Risk Scores}, arXiv preprint arXiv:1609.05807\href
  {http://arxiv.org/abs/1609.05807} {\path{arXiv:1609.05807}}, \href
  {http://dx.doi.org/10.1111/j.1740-9713.2017.01012.x}
  {\path{doi:10.1111/j.1740-9713.2017.01012.x}}.

\bibitem{Chouldechova2017}
A.~Chouldechova, {Fair prediction with disparate impact: A study of bias in
  recidivism prediction instruments}, ArXiv e-prints\href
  {http://arxiv.org/abs/1703.00056} {\path{arXiv:1703.00056}}, \href
  {http://dx.doi.org/10.1089/big.2016.0047} {\path{doi:10.1089/big.2016.0047}}.

\bibitem{Binns2020}
R.~Binns, {On the apparent conflict between individual and group fairness},
  Proceedings of the 2020 Conference on Fairness, Accountability, and
  Transparency (2020) 514--524\href {http://arxiv.org/abs/1912.06883}
  {\path{arXiv:1912.06883}}, \href {http://dx.doi.org/10.1145/3351095.3372864}
  {\path{doi:10.1145/3351095.3372864}}.

\bibitem{Friedler2016}
S.~A. Friedler, C.~Scheidegger, S.~Venkatasubramanian, {On the (im)possibility
  of fairness *}, arXiv preprint arXiv:1609.07236\href
  {http://arxiv.org/abs/1609.07236v1} {\path{arXiv:1609.07236v1}}.

\bibitem{Kearns2017}
M.~Kearns, S.~Neel, A.~Roth, Z.~S. Wu, {Preventing Fairness Gerrymandering:
  Auditing and Learning for Subgroup Fairness}, International Conference on
  Machine Learning (2018) 2564--2572\href {http://arxiv.org/abs/1711.05144}
  {\path{arXiv:1711.05144}}.

\bibitem{Khani2019}
F.~Khani, P.~Liang, {Noise Induces Loss Discrepancy Across Groups for Linear
  Regression}, arXiv preprint arXiv:1911.09876\href
  {http://arxiv.org/abs/1911.09876} {\path{arXiv:1911.09876}}.

\bibitem{Friedler2019}
S.~A. Friedler, S.~Choudhary, C.~Scheidegger, E.~P. Hamilton,
  S.~Venkatasubramanian, D.~Roth,
  \href{https://doi.org/10.1145/3287560.3287589}{{A comparative study of
  fairness-enhancing interventions in machine learning}}, FAT* 2019 -
  Proceedings of the 2019 Conference on Fairness, Accountability, and
  Transparency (2019) 329--338\href {http://arxiv.org/abs/1802.04422}
  {\path{arXiv:1802.04422}}, \href {http://dx.doi.org/10.1145/3287560.3287589}
  {\path{doi:10.1145/3287560.3287589}}.
\newline\urlprefix\url{https://doi.org/10.1145/3287560.3287589}

\bibitem{Lipton2018}
Z.~C. Lipton, J.~McAuley, A.~Chouldechova, J.~McAuley,
  \href{http://papers.nips.cc/paper/8035-does-mitigating-mls-impact-disparity-require-treatment-disparity.pdf}{{Does
  mitigating ML's impact disparity require treatment disparity?}}, Advances in
  Neural Information Processing Systems 2018-Decem (2018) 8125--8135.
\newblock \href {http://arxiv.org/abs/1711.07076} {\path{arXiv:1711.07076}}.
\newline\urlprefix\url{http://papers.nips.cc/paper/8035-does-mitigating-mls-impact-disparity-require-treatment-disparity.pdf}

\bibitem{Calders2009}
T.~Calders, F.~Kamiran, M.~Pechenizkiy,
  \href{https://www.win.tue.nl/{~}mpechen/publications/pubs/CaldersICDM09.pdf}{{Building
  classifiers with independency constraints}}, in: 2009 IEEE International
  Conference on Data Mining Workshops, IEEE, 2009, pp. 13--18.
\newblock \href {http://dx.doi.org/10.1109/ICDMW.2009.83}
  {\path{doi:10.1109/ICDMW.2009.83}}.
\newline\urlprefix\url{https://www.win.tue.nl/{~}mpechen/publications/pubs/CaldersICDM09.pdf}

\bibitem{Celis2018}
L.~E. Celis, L.~Huang, V.~Keswani, N.~K. Vishnoi, {Classification with Fairness
  Constraints: A Meta-Algorithm with Provable Guarantees}, Proceedings of the
  Conference on Fairness, Accountability, and Transparency (2018) 319--328\href
  {http://arxiv.org/abs/1806.06055} {\path{arXiv:1806.06055}}.

\bibitem{Sriperumbudur2009}
B.~K. Sriperumbudur, K.~Fukumizu, A.~Gretton, B.~Sch{\"o}lkopf, G.~R.
  Lanckriet, On integral probability metrics,$\backslash$phi-divergences and
  binary classification, arXiv preprint arXiv:0901.2698.

\bibitem{Ramdas2017}
A.~Ramdas, N.~G. Trillos, M.~Cuturi, {On wasserstein two-sample testing and
  related families of nonparametric tests}, Entropy 19~(2).
\newblock \href {http://arxiv.org/abs/1509.02237} {\path{arXiv:1509.02237}},
  \href {http://dx.doi.org/10.3390/e19020047} {\path{doi:10.3390/e19020047}}.

\bibitem{Yadlowsky2019}
S.~Yadlowsky, S.~Basu, L.~Tian, A calibration metric for risk scores with
  survival data, in: Machine Learning for Healthcare Conference, 2019, pp.
  424--450.

\bibitem{Austin2019}
P.~C. Austin, E.~W. Steyerberg,
  \href{https://onlinelibrary.wiley.com/doi/abs/10.1002/sim.8281}{{The
  Integrated Calibration Index (ICI) and related metrics for quantifying the
  calibration of logistic regression models}}, Statistics in Medicine 38~(21)
  (2019) 4051--4065.
\newblock \href {http://dx.doi.org/10.1002/sim.8281}
  {\path{doi:10.1002/sim.8281}}.
\newline\urlprefix\url{https://onlinelibrary.wiley.com/doi/abs/10.1002/sim.8281}

\bibitem{Liu2019}
L.~T. Liu, M.~Simchowitz, M.~Hardt,
  \href{http://proceedings.mlr.press/v97/liu19f.html}{{The Implicit Fairness
  Criterion of Unconstrained Learning}}, in: K.~Chaudhuri, R.~Salakhutdinov
  (Eds.), Proceedings of the 36th International Conference on Machine Learning,
  Vol.~97 of Proceedings of Machine Learning Research, PMLR, Long Beach,
  California, USA, 2019, pp. 4051--4060.
\newline\urlprefix\url{http://proceedings.mlr.press/v97/liu19f.html}

\bibitem{pleiss2017fairness}
G.~Pleiss, M.~Raghavan, F.~Wu, J.~Kleinberg, K.~Q. Weinberger, {On fairness and
  calibration}, in: Advances in Neural Information Processing Systems, 2017,
  pp. 5680--5689.
\newblock \href {http://arxiv.org/abs/1709.02012} {\path{arXiv:1709.02012}}.

\bibitem{Kallus2019}
N.~Kallus, A.~Zhou, {The fairness of risk scores beyond classification:
  Bipartite ranking and the xauc metric}, in: Advances in Neural Information
  Processing Systems, 2019, pp. 3438--3448.
\newblock \href {http://arxiv.org/abs/1902.05826} {\path{arXiv:1902.05826}}.

\bibitem{beutel2019fairness}
A.~Beutel, J.~Chen, T.~Doshi, H.~Qian, L.~Wei, Y.~Wu, L.~Heldt, Z.~Zhao,
  L.~Hong, E.~H. Chi, C.~Goodrow, {Fairness in recommendation ranking through
  pairwise comparisons}, Proceedings of the ACM SIGKDD International Conference
  on Knowledge Discovery and Data Mining (2019) 2212--2220\href
  {http://arxiv.org/abs/1903.00780} {\path{arXiv:1903.00780}}, \href
  {http://dx.doi.org/10.1145/3292500.3330745}
  {\path{doi:10.1145/3292500.3330745}}.

\bibitem{Louizos2015}
C.~Louizos, K.~Swersky, Y.~Li, M.~Welling, R.~Zemel, {The Variational Fair
  Autoencoder}, arXiv preprint arXiv:1511.00830\href
  {http://arxiv.org/abs/1511.00830} {\path{arXiv:1511.00830}}.

\bibitem{Madras2018}
D.~Madras, E.~Creager, T.~Pitassi, R.~Zemel,
  \href{http://proceedings.mlr.press/v80/madras18a.html
  http://arxiv.org/abs/1802.06309}{{Learning Adversarially Fair and
  Transferable Representations}}, Proceedings of the 35th International
  Conference on Machine Learning 80 (2018) 3384--3393.
\newblock \href {http://arxiv.org/abs/1802.06309} {\path{arXiv:1802.06309}}.
\newline\urlprefix\url{http://proceedings.mlr.press/v80/madras18a.html
  http://arxiv.org/abs/1802.06309}

\bibitem{Ilvento2020}
C.~Ilvento, Metric learning for individual fairness, in: 1st Symposium on
  Foundations of Responsible Computing (FORC 2020), Schloss
  Dagstuhl-Leibniz-Zentrum f{\"u}r Informatik, 2020.

\bibitem{pmlr-v54-zafar17a}
M.~B. Zafar, I.~Valera, M.~{Gomez Rodriguez}, K.~P. Gummadi, M.~G. Rogriguez,
  K.~P. Gummadi, {Fairness Constraints: Mechanisms for Fair Classification},
  in: A.~Singh, J.~Zhu (Eds.), Proceedings of the 20th International Conference
  on Artificial Intelligence and Statistics, Vol.~54 of Proceedings of Machine
  Learning Research, PMLR, Fort Lauderdale, FL, USA, 2017, pp. 962--970.

\bibitem{gretton2012kernel}
A.~Gretton, K.~M. Borgwardt, M.~J. Rasch, B.~Sch{\"{o}}lkopf, A.~Smola, {A
  kernel two-sample test}, Journal of Machine Learning Research 13~(Mar) (2012)
  723--773.

\bibitem{Datta2020}
S.~Datta, J.~Posada, G.~Olson, W.~Li, D.~Balraj, J.~Mesterhazy, J.~Pallas,
  P.~Desai, N.~H. Shah, {A new paradigm for accelerating clinical data science
  at Stanford Medicine}, arXiv preprint arXiv:2003.10534\href
  {http://arxiv.org/abs/2003.10534} {\path{arXiv:2003.10534}}.

\bibitem{Hripcsak2015}
G.~Hripcsak, J.~D. Duke, N.~H. Shah, C.~G. Reich, V.~Huser, M.~J. Schuemie,
  M.~A. Suchard, R.~W. Park, I.~C.~K. Wong, P.~R. Rijnbeek, J.~{Van Der Lei},
  N.~Pratt, G.~N. Nor{\'{e}}n, Y.-C.~C. Li, P.~E. Stang, D.~Madigan, P.~B.
  Ryan, {Observational Health Data Sciences and Informatics (OHDSI):
  Opportunities for Observational Researchers}, in: Studies in Health
  Technology and Informatics, Vol. 216, NIH Public Access, 2015, pp. 574--578.
\newblock \href {http://dx.doi.org/10.3233/978-1-61499-564-7-574}
  {\path{doi:10.3233/978-1-61499-564-7-574}}.

\bibitem{Overhage2012}
J.~{Marc Overhage}, P.~B. Ryan, C.~G. Reich, A.~G. Hartzema, P.~E. Stang,
  {Validation of a common data model for active safety surveillance research},
  Journal of the American Medical Informatics Association 19~(1) (2012) 54--60.
\newblock \href {http://dx.doi.org/10.1136/amiajnl-2011-000376}
  {\path{doi:10.1136/amiajnl-2011-000376}}.

\bibitem{Reps2018}
J.~M. Reps, M.~J. Schuemie, M.~A. Suchard, P.~B. Ryan, P.~R. Rijnbeek,
  \href{https://academic.oup.com/jamia/advance-article/doi/10.1093/jamia/ocy032/4989437}{{Design
  and implementation of a standardized framework to generate and evaluate
  patient-level prediction models using observational healthcare data}},
  Journal of the American Medical Informatics Association 25~(8) (2018)
  969--975.
\newblock \href {http://dx.doi.org/10.1093/jamia/ocy032}
  {\path{doi:10.1093/jamia/ocy032}}.
\newline\urlprefix\url{https://academic.oup.com/jamia/advance-article/doi/10.1093/jamia/ocy032/4989437}

\bibitem{Johnson2016}
A.~E. Johnson, T.~J. Pollard, L.~Shen, L.-w.~H. Lehman, M.~Feng, M.~Ghassemi,
  B.~Moody, P.~Szolovits, L.~{Anthony Celi}, R.~G. Mark,
  \href{http://www.nature.com/articles/sdata201635}{{MIMIC-III, a freely
  accessible critical care database}}, Scientific Data 3 (2016) 160035.
\newblock \href {http://dx.doi.org/10.1038/sdata.2016.35}
  {\path{doi:10.1038/sdata.2016.35}}.
\newline\urlprefix\url{http://www.nature.com/articles/sdata201635}

\bibitem{Wang2020}
S.~Wang, M.~B.~A. McDermott, G.~Chauhan, M.~Ghassemi, M.~C. Hughes, T.~Naumann,
  M.~Ghassemi, {Mimic-extract: A data extraction, preprocessing, and
  representation pipeline for mimic-iii}, in: Proceedings of the ACM Conference
  on Health, Inference, and Learning, 2020, pp. 222--235.
\newblock \href {http://arxiv.org/abs/1907.08322} {\path{arXiv:1907.08322}}.

\bibitem{kingma2014adam}
D.~Kingma, J.~Ba, {Adam: A method for stochastic optimization}, arXiv preprint
  arXiv:1412.6980.

\bibitem{pytorch}
A.~Paszke, S.~Gross, F.~Massa, A.~Lerer, J.~Bradbury, G.~Chanan, T.~Killeen,
  Z.~Lin, N.~Gimelshein, L.~Antiga, A.~Desmaison, A.~Kopf, E.~Yang, Z.~DeVito,
  M.~Raison, A.~Tejani, S.~Chilamkurthy, B.~Steiner, L.~Fang, J.~Bai,
  S.~Chintala,
  \href{http://papers.neurips.cc/paper/9015-pytorch-an-imperative-style-high-performance-deep-learning-library.pdf}{Pytorch:
  An imperative style, high-performance deep learning library}, in: H.~Wallach,
  H.~Larochelle, A.~Beygelzimer, F.~d'~Alch'{e}-Buc, E.~Fox, R.~Garnett (Eds.),
  Advances in Neural Information Processing Systems 32, Curran Associates,
  Inc., 2019, pp. 8024--8035.
\newline\urlprefix\url{http://papers.neurips.cc/paper/9015-pytorch-an-imperative-style-high-performance-deep-learning-library.pdf}

\bibitem{fletcher2013practical}
R.~Fletcher, {Practical methods of optimization}, John Wiley {\&} Sons, 2013.

\bibitem{scikit-learn}
F.~Pedregosa, G.~Varoquaux, A.~Gramfort, V.~Michel, B.~Thirion, O.~Grisel,
  M.~Blondel, P.~Prettenhofer, R.~Weiss, V.~Dubourg, J.~Vanderplas, A.~Passos,
  D.~Cournapeau, M.~Brucher, M.~Perrot, E.~Duchesnay, {Scikit-learn: Machine
  Learning in Python}, Journal of Machine Learning Research 12 (2011)
  2825--2830.
\newblock \href {http://arxiv.org/abs/1201.0490} {\path{arXiv:1201.0490}},
  \href {http://dx.doi.org/10.1007/s13398-014-0173-7.2}
  {\path{doi:10.1007/s13398-014-0173-7.2}}.

\bibitem{Corbett-Davies:2017:ADM:3097983.3098095}
S.~Corbett-Davies, E.~Pierson, A.~Feller, S.~Goel, A.~Huq, {Algorithmic
  Decision Making and the Cost of Fairness}, in: Proceedings of the 23rd ACM
  SIGKDD International Conference on Knowledge Discovery and Data Mining, KDD
  '17, ACM, New York, NY, USA, 2017, pp. 797--806.
\newblock \href {http://arxiv.org/abs/1701.08230} {\path{arXiv:1701.08230}},
  \href {http://dx.doi.org/10.1145/3097983.3098095}
  {\path{doi:10.1145/3097983.3098095}}.

\bibitem{Sewell2016}
A.~A. Sewell,
  \href{http://journals.sagepub.com/doi/10.1177/2332649215626936}{{The
  Racism-Race Reification Process}}, Sociology of Race and Ethnicity 2~(4)
  (2016) 402--432.
\newblock \href {http://dx.doi.org/10.1177/2332649215626936}
  {\path{doi:10.1177/2332649215626936}}.
\newline\urlprefix\url{http://journals.sagepub.com/doi/10.1177/2332649215626936}

\bibitem{VanderWeele2014}
T.~J. VanderWeele, W.~R. Robinson, {On the Causal Interpretation of Race in
  Regressions Adjusting for Confounding and Mediating Variables}, Epidemiology
  25~(4) (2014) 473--484.
\newblock \href {http://dx.doi.org/10.1097/EDE.0000000000000105}
  {\path{doi:10.1097/EDE.0000000000000105}}.

\bibitem{Duster2005}
T.~Duster, \href{https://science.sciencemag.org/content/307/5712/1050}{{Race
  and Reification in Science}}, Science 307~(5712) (2005) 1050--1051.
\newblock \href {http://dx.doi.org/10.1126/science.1110303}
  {\path{doi:10.1126/science.1110303}}.
\newline\urlprefix\url{https://science.sciencemag.org/content/307/5712/1050}

\bibitem{Braun2007}
L.~Braun, A.~Fausto-Sterling, D.~Fullwiley, E.~M. Hammonds, A.~Nelson,
  W.~Quivers, S.~M. Reverby, A.~E. Shields,
  \href{https://dx.plos.org/10.1371/journal.pmed.0040271}{{Racial Categories in
  Medical Practice: How Useful Are They?}}, PLoS Medicine 4~(9) (2007) e271.
\newblock \href {http://dx.doi.org/10.1371/journal.pmed.0040271}
  {\path{doi:10.1371/journal.pmed.0040271}}.
\newline\urlprefix\url{https://dx.plos.org/10.1371/journal.pmed.0040271}

\bibitem{crenshaw1989demarginalizing}
K.~Crenshaw, {Demarginalizing the intersection of race and sex: A black
  feminist critique of antidiscrimination doctrine, feminist theory and
  antiracist politics}, u. Chi. Legal f. (1989) 139.

\bibitem{Hoffmann2019}
A.~L. Hoffmann, {Where fairness fails: data, algorithms, and the limits of
  antidiscrimination discourse}, Information Communication and Society 22~(7)
  (2019) 900--915.
\newblock \href {http://dx.doi.org/10.1080/1369118X.2019.1573912}
  {\path{doi:10.1080/1369118X.2019.1573912}}.

\bibitem{pmlr-v80-hebert-johnson18a}
{\'{U}}.~H{\'{e}}bert-Johnson, M.~P. Kim, O.~Reingold, G.~N. Rothblum,
  {Calibration for the (Computationally-Identifiable) Masses}, in: J.~Dy,
  A.~Krause (Eds.), Proceedings of the 35th International Conference on Machine
  Learning, Vol.~80 of Proceedings of Machine Learning Research, PMLR,
  Stockholmsm{\"{a}}ssan, Stockholm Sweden, 2017, pp. 1939--1948.
\newblock \href {http://arxiv.org/abs/1711.08513} {\path{arXiv:1711.08513}}.

\bibitem{Cirillo2020}
D.~Cirillo, S.~Catuara-Solarz, C.~Morey, E.~Guney, L.~Subirats, S.~Mellino,
  A.~Gigante, A.~Valencia, M.~J. Rementeria, A.~S. Chadha, N.~Mavridis,
  \href{https://www.nature.com/articles/s41746-020-0288-5}{{Sex and gender
  differences and biases in artificial intelligence for biomedicine and
  healthcare}}, npj Digital Medicine 3~(1) (2020) 1--11.
\newblock \href {http://dx.doi.org/10.1038/s41746-020-0288-5}
  {\path{doi:10.1038/s41746-020-0288-5}}.
\newline\urlprefix\url{https://www.nature.com/articles/s41746-020-0288-5}

\bibitem{Kusner2017}
M.~J. Kusner, J.~Loftus, C.~Russell, R.~Silva, {Counterfactual Fairness}, in:
  I.~Guyon, U.~V. Luxburg, S.~Bengio, H.~Wallach, R.~Fergus, S.~Vishwanathan,
  R.~Garnett (Eds.), Advances in Neural Information Processing Systems 30,
  Curran Associates, Inc., 2017, pp. 4066--4076.

\bibitem{Ford2010}
C.~L. Ford, C.~O. Airhihenbuwa, {The public health critical race methodology:
  Praxis for antiracism research}, Social Science and Medicine 71~(8) (2010)
  1390--1398.
\newblock \href {http://dx.doi.org/10.1016/j.socscimed.2010.07.030}
  {\path{doi:10.1016/j.socscimed.2010.07.030}}.

\bibitem{kilbertus2017avoiding}
N.~Kilbertus, M.~R. Carulla, G.~Parascandolo, M.~Hardt, D.~Janzing,
  B.~Sch{\"{o}}lkopf, M.~Rojas-Carulla, G.~Parascandolo, M.~Hardt, D.~Janzing,
  B.~Sch{\"{o}}lkopf, {Avoiding discrimination through causal reasoning}, in:
  Advances in Neural Information Processing Systems, 2017, pp. 656--666.
\newblock \href {http://arxiv.org/abs/1706.02744} {\path{arXiv:1706.02744}}.

\bibitem{Benjamin421}
R.~Benjamin,
  \href{https://science.sciencemag.org/content/366/6464/421}{{Assessing risk,
  automating racism}}, Science 366~(6464) (2019) 421--422.
\newblock \href {http://dx.doi.org/10.1126/science.aaz3873}
  {\path{doi:10.1126/science.aaz3873}}.
\newline\urlprefix\url{https://science.sciencemag.org/content/366/6464/421}

\bibitem{Kalluri2020}
P.~Kalluri, \href{http://www.nature.com/articles/d41586-020-02003-2}{{Don't ask
  if artificial intelligence is good or fair, ask how it shifts power}}, Nature
  583~(7815) (2020) 169--169.
\newblock \href {http://dx.doi.org/10.1038/d41586-020-02003-2}
  {\path{doi:10.1038/d41586-020-02003-2}}.
\newline\urlprefix\url{http://www.nature.com/articles/d41586-020-02003-2}

\bibitem{Sendak2020}
M.~Sendak, M.~C. Elish, M.~Gao, J.~Futoma, W.~Ratli, M.~Nichols, A.~Bedoya,
  C.~O. Brien, W.~Ratliff, M.~Nichols, A.~Bedoya, S.~Balu, C.~O'Brien,
  W.~Ratli, M.~Nichols, A.~Bedoya, C.~O. Brien, {"The Human Body is a Black
  Box": Supporting Clinical Decision-Making with Deep Learning}, in:
  Proceedings of the 2020 Conference on Fairness, Accountability, and
  Transparency, 2020, pp. 99--109.
\newblock \href {http://arxiv.org/abs/1911.08089} {\path{arXiv:1911.08089}}.

\bibitem{Martin2020}
D.~Martin, V.~Prabhakaran, J.~Kuhlberg, A.~Smart, W.~S. Isaac, {Participatory
  Problem Formulation for Fairer Machine Learning Through Community Based
  System Dynamics}, arXiv preprint arXiv:2005.07572\href
  {http://arxiv.org/abs/2005.07572} {\path{arXiv:2005.07572}}.

\bibitem{Martin2020a}
D.~Martin, V.~Prabhakaran, J.~Kuhlberg, A.~Smart, W.~S. Isaac, {Extending the
  Machine Learning Abstraction Boundary: A Complex Systems Approach to
  Incorporate Societal Context}, arXiv preprint arXiv:2006.09663\href
  {http://arxiv.org/abs/2006.09663} {\path{arXiv:2006.09663}}.

\bibitem{selbst2019fairness}
A.~D. Selbst, D.~Boyd, S.~A. Friedler, S.~Venkatasubramanian, J.~Vertesi,
  {Fairness and abstraction in sociotechnical systems}, in: Proceedings of the
  Conference on Fairness, Accountability, and Transparency, 2019, pp. 59--68.

\bibitem{Baumer2011}
E.~P. Baumer, M.~S. Silberman,
  \href{http://dl.acm.org/citation.cfm?doid=1978942.1979275}{{When the
  implication is not to design (technology)}}, in: Proceedings of the 2011
  annual conference on Human factors in computing systems - CHI '11, ACM Press,
  New York, New York, USA, 2011, p. 2271.
\newblock \href {http://dx.doi.org/10.1145/1978942.1979275}
  {\path{doi:10.1145/1978942.1979275}}.
\newline\urlprefix\url{http://dl.acm.org/citation.cfm?doid=1978942.1979275}

\end{thebibliography}

\clearpage
\appendix

\counterwithin{figure}{section}
\counterwithin{table}{section}
\renewcommand*\thetable{\Alph{section}.\arabic{table}} 
\renewcommand*\thefigure{\Alph{section}.\arabic{figure}}

\section{Supplementary Cohort Tables}
\begin{table}[!htb]
\centering
\caption{Cohort characteristics for patients drawn from Optum CDM. Data are grouped on the basis of the age group and sex. Shown are the number of patients extracted and the incidence of 30-day readmission and prolonged length of stay (hospital length of stay greater than or equal to 7 days)}
\label{tab:cohort_optum}
\begin{tabular}{lrrr}
\toprule
{} & {} & \multicolumn{2}{c}{Outcome Incidence} \\
\cmidrule{3-4}
Group &   Count & 30-Day Readmission & Prolonged Length of Stay \\
\midrule
\lbrack18-30)             &  1,067,423 &          0.0346 & 0.0608 \\
\lbrack30-45)             &  1,854,239 &          0.0347 & 0.0611 \\
\lbrack45-55)             &  1,006,924 &          0.0611 &  0.138 \\
\lbrack55-65)             &  1,173,140 &          0.0808 &  0.195 \\
\lbrack65-75)             &  1,294,273 &        0.100 &  0.258 \\
\lbrack75-90)             &  1,678,572 &           0.168 &  0.386 \\
\midrule
Female              &  5,040,564 &          0.0765 &  0.168 \\
Male                &  3,032,831 &          0.0938 &  0.224 \\
\bottomrule

\end{tabular}
\end{table}

\begin{table}[!htb]
\centering
\caption{Cohort characteristics for patients drawn from MIMIC-III. Data are grouped on the basis of the age group, sex, and the race and ethnicity category. Shown are the number of patients extracted and the incidence of an ICU length of stay greater than three and seven days and of hospital and ICU mortality.}
\label{tab:cohort_mimic}
\begin{tabular}{lrrrrr}
\toprule
{} & {} & \multicolumn{4}{c}{Outcome Incidence} \\
\cmidrule{3-6}
Group &   Count & ICU LOS > 3  &  ICU LOS > 7 &  Hospital Mortality &  ICU Mortality \\
\midrule
\lbrack15-30) &   1,345 &          0.274 &         0.0491 &              0.0387 &         0.0238 \\
\lbrack30-45) &   2,621 &          0.274 &           0.0500 &              0.0542 &         0.0332 \\
\lbrack45-55) &   3,865 &          0.297 &         0.0505 &              0.0743 &         0.0422 \\
\lbrack55-65) &   5,358 &          0.308 &         0.0524 &              0.0769 &         0.0455 \\
\lbrack65-75) &   5,620 &          0.328 &         0.0571 &              0.0961 &         0.0557 \\
\lbrack75-90) &   7,361 &          0.356 &         0.0583 &                0.140 &         0.0793 \\
\midrule
Female  &  11,108 &          0.326 &         0.0568 &               0.102 &         0.0593 \\
Male    &  15,062 &          0.314 &         0.0526 &              0.0889 &         0.0507 \\
\midrule
Other   &   7,639 &          0.325 &         0.0579 &               0.106 &         0.0624 \\
White   &  18,531 &          0.316 &         0.0529 &              0.0895 &          0.0510 \\
\bottomrule
\end{tabular}
\end{table}

\clearpage
\section{Feature Extraction}
\begin{figure}[!ht]
	\centering
	\includegraphics[width=0.85\linewidth]{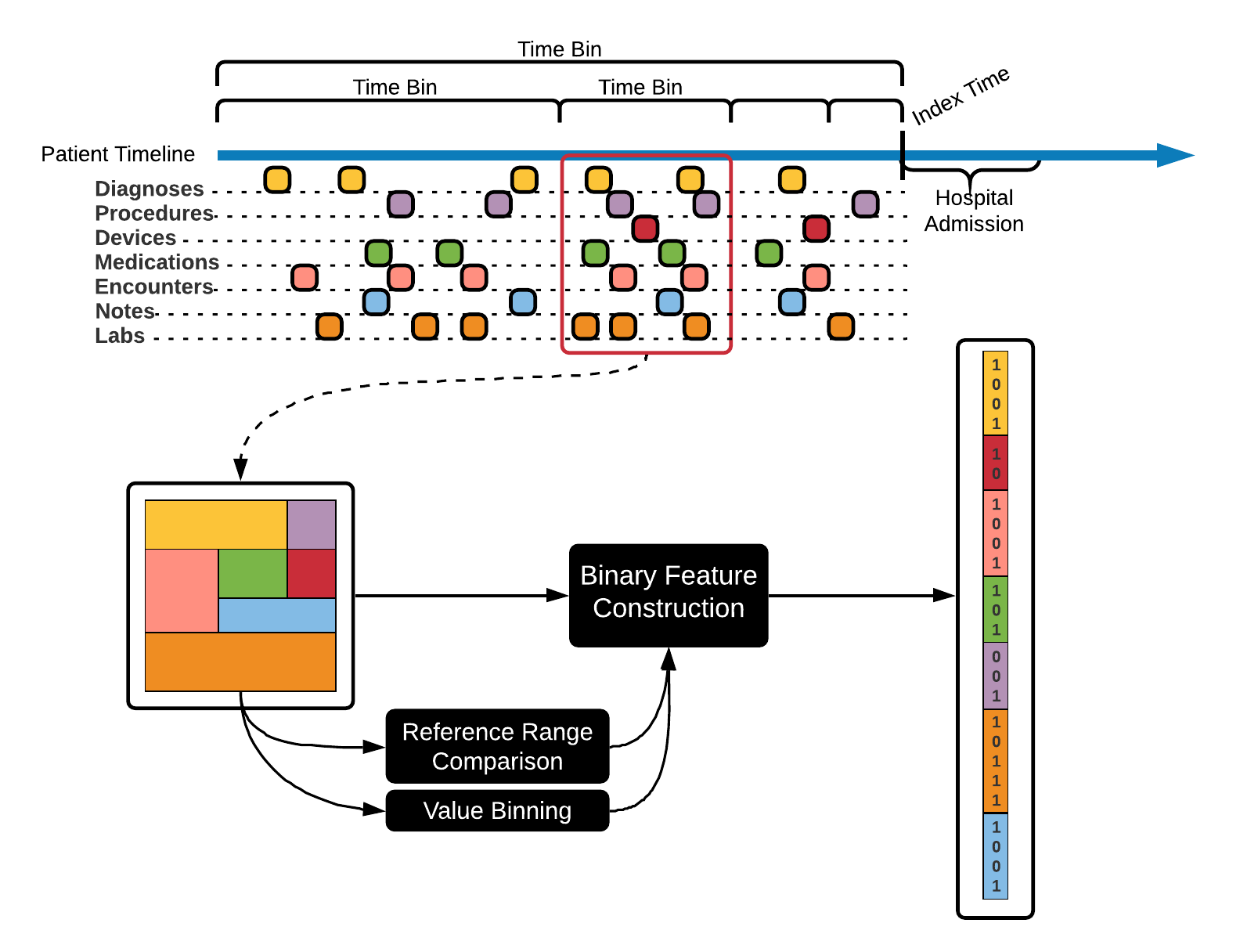}
	\caption{
	    An overview of the procedure for extracting features from a patient timeline in the STARR cohort. The procedure occurs nested within intervals defined relative to the start of an index hospital admission. Within each interval, a set of binary features is constructed on the basis of observation of unique OMOP CDM concepts. A binary representation of numeric lab results is generated by both comparing the results to the associated reference ranges, and via mapping the results to bins (defined on the basis of the empirical quintiles of the associated lab within the time bin) across patients in the cohort. The final feature space is constructed via the concatenation of the features derived in each interval.
	}
	\label{fig:supplement/feature_extraction}
\end{figure}

\clearpage
\section{Hyperparameters} \label{app:hyperparameters}

\begin{table}[!htb]
\centering
\caption{The hyperparameter grid used for tuning feedforward neural networks with a fixed hidden layer size. The full grid is constructed via the cartesian product of the listed grid values for each hyperparameter. The random search procedure evaluates fifty elements from the full grid.
}
\label{tab:hyperparameter_grid}
\begin{tabular}{ll}
\toprule
          Hyperparameter &                    Grid Values \\
\midrule
              Batch Size &         [128, 256, 512] \\
     Dropout Probability &  [0.0, 0.25, 0.5, 0.75] \\
        Hidden Dimension &              [128, 256] \\
           Learning Rate &  [$10^{-3}$, $10^{-4}$, $10^{-5}$] \\
 Number of Hidden Layers &               [1, 2, 3] \\
\bottomrule
\end{tabular}
\end{table}

\begin{table}[!htb]
\centering
\caption{
Selected model hyperparameters for each outcome defined for the cohort derived from the STARR database.
}
\label{tab:hyperparameter_starr}
\begin{tabular}{llll}
\toprule
          Hyperparameter & Hospital Mortality &   Prolonged Length of Stay & 30-Day Readmission \\
\midrule
              Batch Size &                512 &     256 &            512 \\
     Dropout Probability &               0.75 &    0.75 &           0.75 \\
        Hidden Dimension &                256 &     128 &            128 \\
           Learning Rate &             $10^{-4}$ &  $10^{-4}$ &          $10^{-5}$ \\
 Number of Hidden Layers &                  3 &       1 &              3 \\
\bottomrule
\end{tabular}
\end{table}

\begin{table}[!htb]
\centering
\caption{
Selected model hyperparameters for each outcome defined for the cohort derived from the Optum CDM database.
}
\label{tab:hyperparameter_optum}
\begin{tabular}{lll}
\toprule
          Hyperparameter & 30-Day Readmission &  Prolonged Length of Stay \\
\midrule
              Batch Size &            512 &    512 \\
     Dropout Probability &           0.25 &   0.25 \\
        Hidden Dimension &            128 &    128 \\
           Learning Rate &          $10^{-5}$ &  $10^{-5}$ \\
 Number of Hidden Layers &              3 &      3 \\
\bottomrule
\end{tabular}
\end{table}

\begin{table}[!htb]
\centering
\caption{
Selected model hyperparameters for each outcome defined for the cohort derived from the MIMIC-III database.
}
\label{tab:hyperparameter_mimic}
\begin{tabular}{lllll}
\toprule
          Hyperparameter & ICU LOS > 3 & ICU LOS > 7 & Hospital Mortality & ICU Mortality \\
\midrule
              Batch Size &           128 &           512 &                128 &           128 \\
     Dropout Probability &          0.75 &          0.75 &               0.75 &          0.75 \\
        Hidden Dimension &           256 &           128 &                256 &           256 \\
           Learning Rate &         1e-05 &         1e-05 &              1e-05 &         1e-05 \\
 Number of Hidden Layers &             1 &             3 &                  1 &             1 \\
\bottomrule
\end{tabular}
\end{table}

\clearpage
\section{Supplementary Figures}
\subsection{STARR}
\begin{figure}[!htb]
	\centering
	\includegraphics[width=0.9\linewidth]{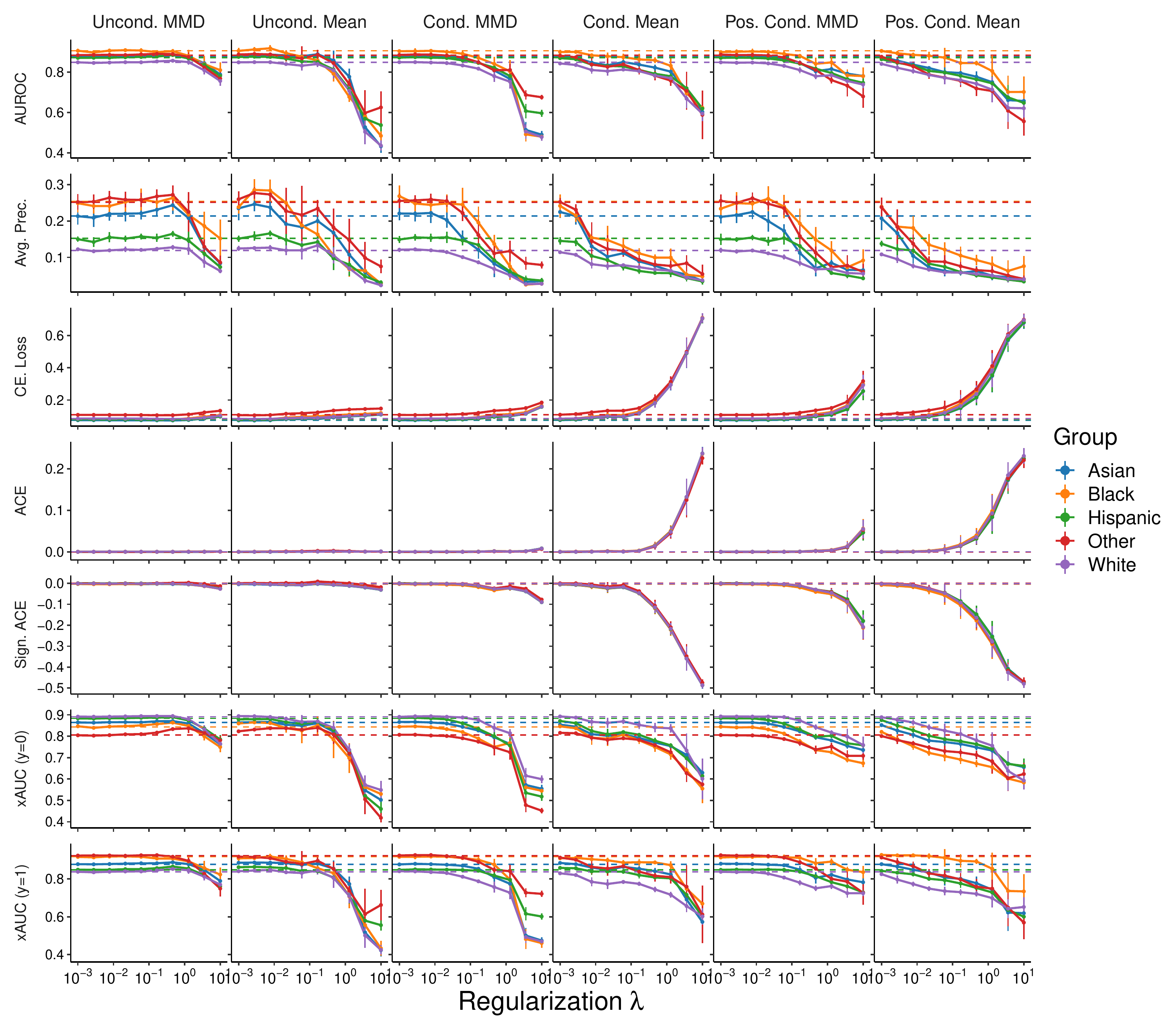}
	\caption{
	    Group-level model performance measures as a function of the extent $\lambda$ that violation of the fairness criterion is penalized when the \textbf{race and ethnicity} category is considered as the sensitive attribute for prediction of \textbf{hospital mortality} in the \textbf{STARR} database. 
	    Results shown are the mean $\pm$ SD for the area under the ROC curve (AUROC), average precision (Avg. Prec), the cross entropy loss (CE Loss), the absolute calibration error (ACE), the signed absolute calibration error (Sign. ACE), and cross group ranking performance (xAUC; $\textrm{xAUC}_k^1$ is indicated by (y=1) and $\textrm{xAUC}_k^0$ by (y=0)) for each group for objectives that penalize violation of threshold-free Demographic Parity (Uncond. MMD and Mean), Equalized Odds (Cond. MMD and Mean), and Equal Opportunity (Pos. Cond. MMD and Mean) with MMD- and mean-based penalties. Dashed lines correspond to the mean result for the unpenalized training procedure.
	}
	\label{fig:supplement/starr/all_performance/hospital_mortality/race_eth}
\end{figure}

\begin{figure}[!htb]
	\centering
	\includegraphics[width=0.9\linewidth]{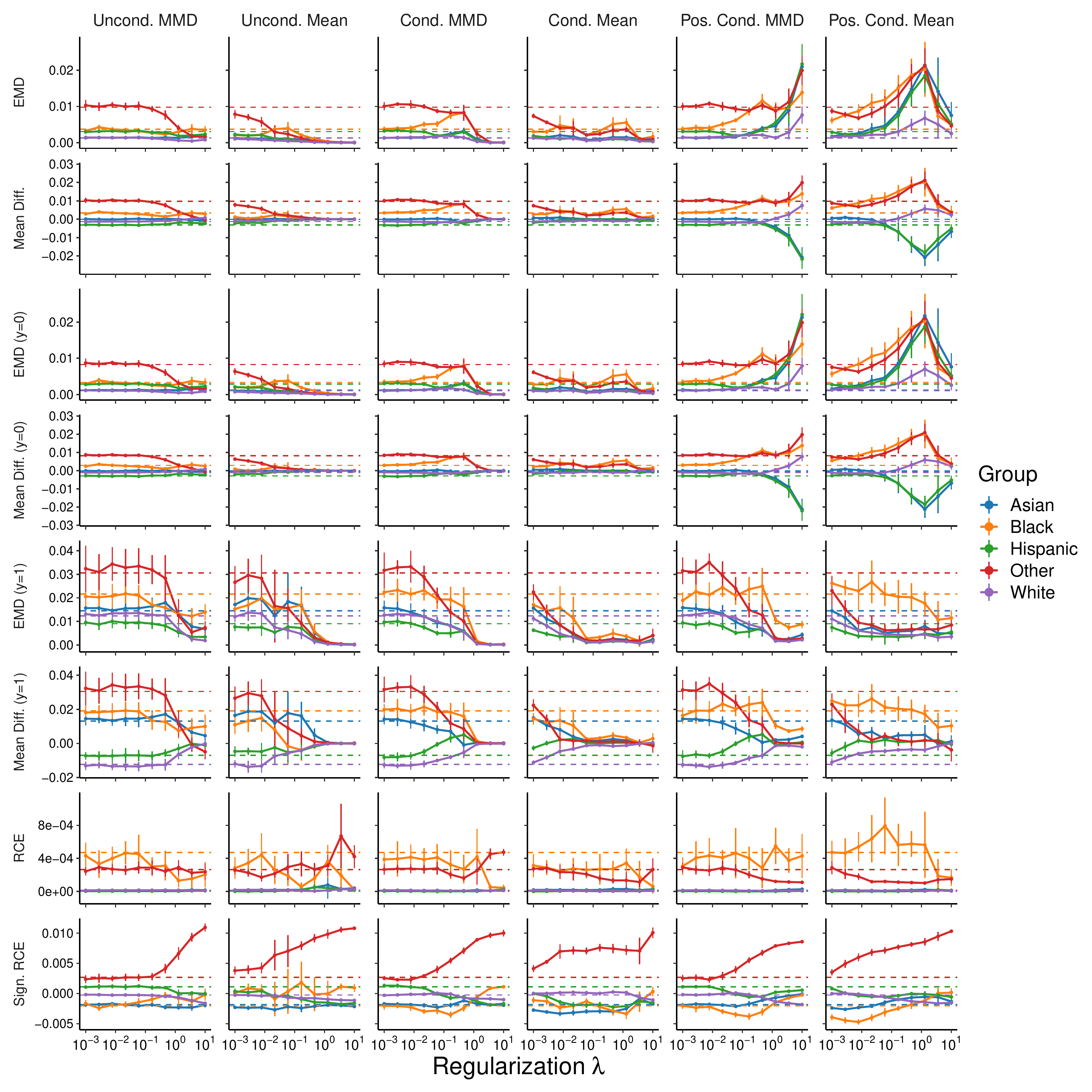}
	\caption{
	    Fairness metrics as a function of the extent $\lambda$ that violation of the fairness criterion is penalized when the \textbf{race and ethnicity} category is considered as the sensitive attribute for prediction of \textbf{hospital mortality} in the \textbf{STARR} database. Results shown are the mean $\pm$ SD for decomposed group-level metrics that assess conditional prediction parity (EMD and Mean Diff.) and relative calibration error (RCE and Sign. RCE) for objectives that penalize violation of threshold-free Demographic Parity (Uncond. MMD and Mean), Equalized Odds (Cond. MMD and Mean), and Equal Opportunity (Pos. Cond. MMD and Mean) on the basis of MMD- and mean-based penalties. Measures of conditional prediction parity are separately assessed in the whole population and in the strata for which the outcome is and is not observed (suffixed with (y=1) and (y=0), respectively). Dashed lines correspond to the mean result for the unpenalized training procedure.
	}
	\label{fig:supplement/starr/all_fairness/hospital_mortality/race_eth}
\end{figure}

\begin{figure}[!htb]
	\centering
	\includegraphics[width=0.9\linewidth]{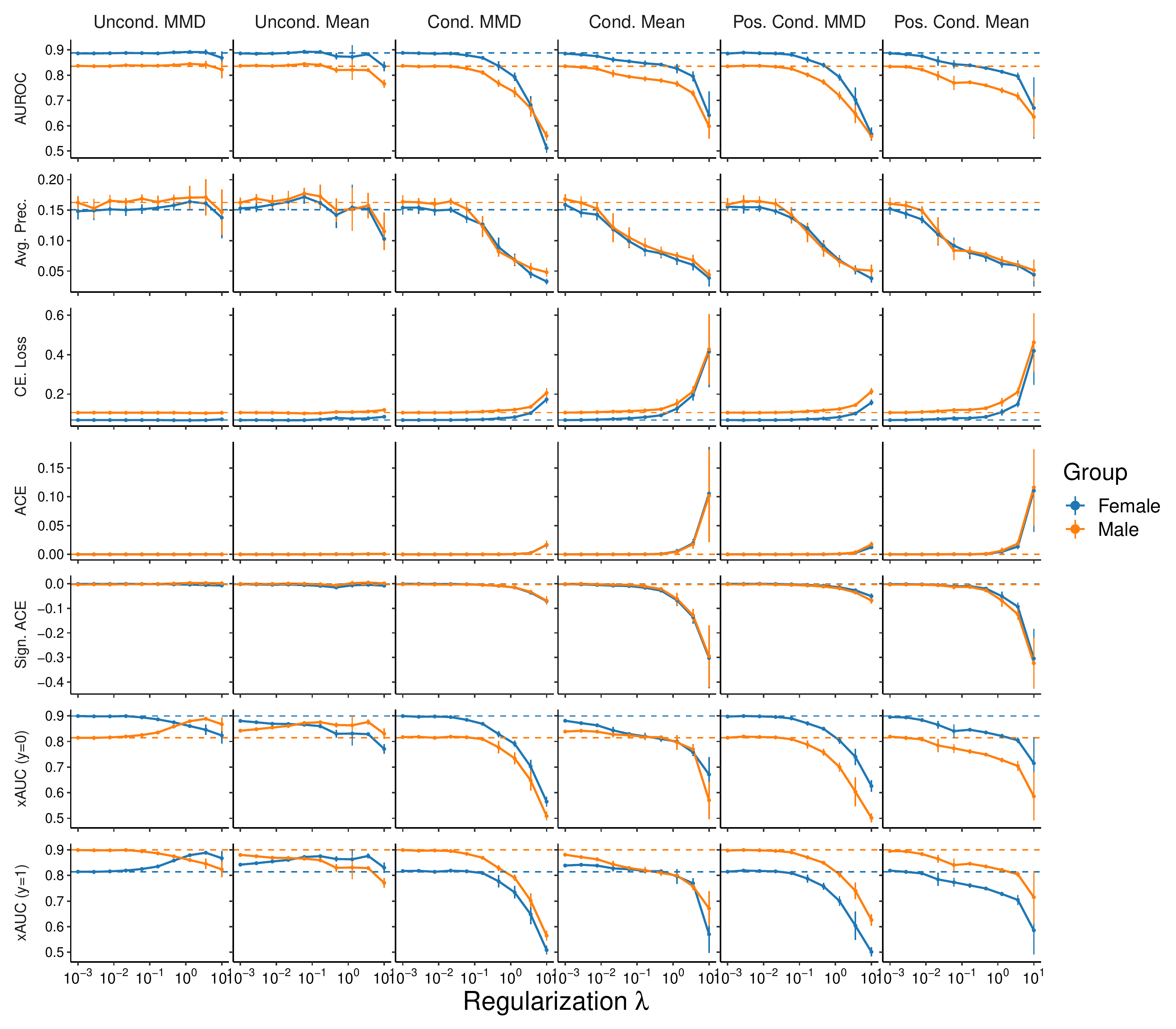}
	\caption{
	    Group-level model performance measures as a function of the extent $\lambda$ that violation of the fairness criterion is penalized when \textbf{sex} is considered as the sensitive attribute for prediction of \textbf{hospital mortality} in the \textbf{STARR} database. Results shown are the mean $\pm$ SD for the area under the ROC curve (AUROC), average precision (Avg. Prec), the cross entropy loss (CE Loss), the absolute calibration error (ACE), the signed absolute calibration error (Sign. ACE), and cross group ranking performance (xAUC; $\textrm{xAUC}_k^1$ is indicated by (y=1) and $\textrm{xAUC}_k^0$ by (y=0)) for each group for objectives that penalize violation of threshold-free Demographic Parity (Uncond. MMD and Mean), Equalized Odds (Cond. MMD and Mean), and Equal Opportunity (Pos. Cond. MMD and Mean) with MMD- and mean-based penalties. Dashed lines correspond to the mean result for the unpenalized training procedure.
	}
	\label{fig:supplement/starr/all_performance/hospital_mortality/gender_concept_name}
\end{figure}

\begin{figure}[!htb]
	\centering
	\includegraphics[width=0.9\linewidth]{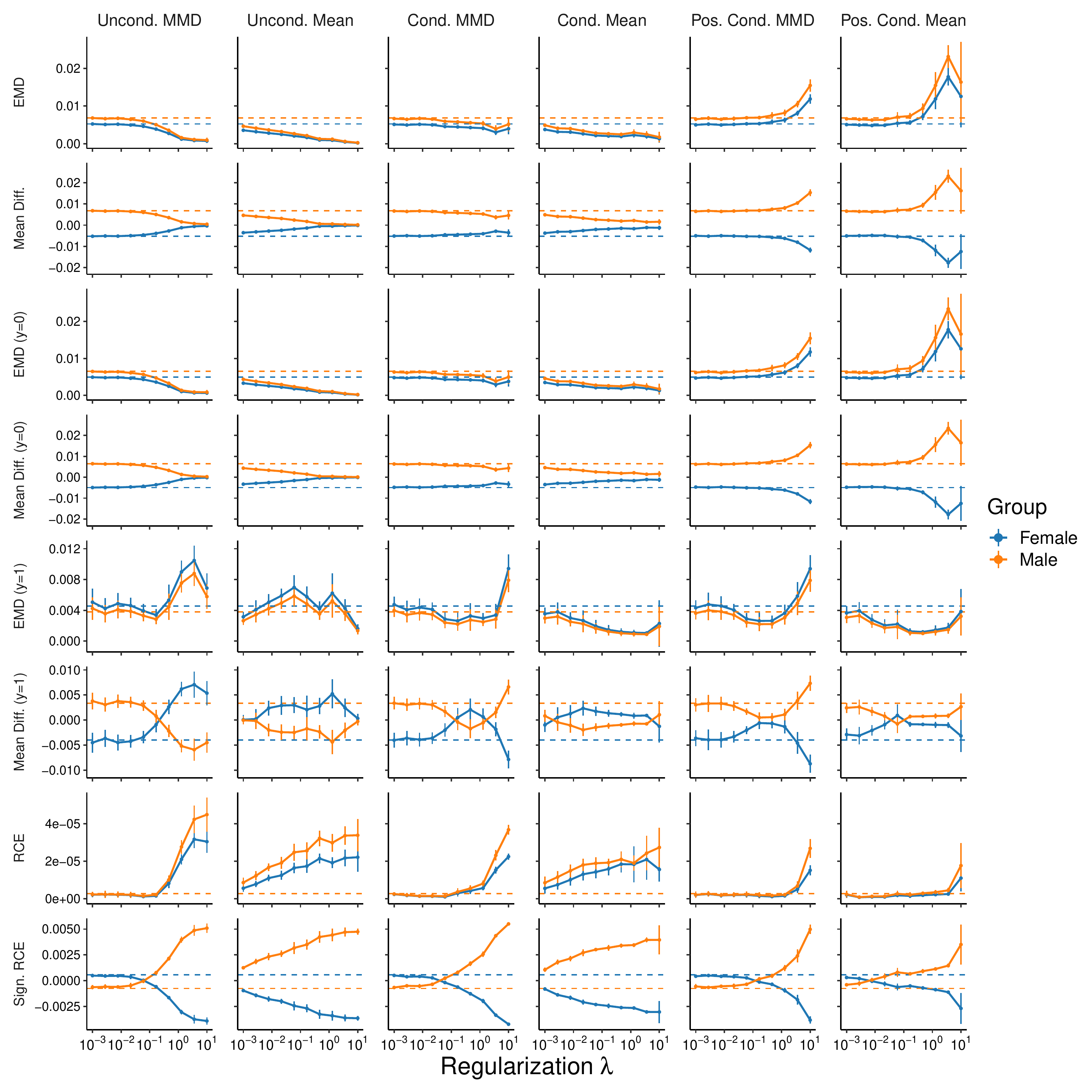}
	\caption{
	    Fairness metrics as a function of the extent $\lambda$ that violation of the fairness criterion is penalized when \textbf{sex} is considered as the sensitive attribute for prediction of \textbf{hospital mortality} in the \textbf{STARR} database. Results shown are the mean $\pm$ SD for decomposed group-level metrics that assess conditional prediction parity (EMD and Mean Diff.) and relative calibration error (RCE and Sign. RCE) for objectives that penalize violation of threshold-free Demographic Parity (Uncond. MMD and Mean), Equalized Odds (Cond. MMD and Mean), and Equal Opportunity (Pos. Cond. MMD and Mean) on the basis of MMD- and mean-based penalties.  Measures of conditional prediction parity are separately assessed in the whole population and in the strata for which the outcome is and is not observed (suffixed with (y=1) and (y=0), respectively). Dashed lines correspond to the mean result for the unpenalized training procedure.
	}
	\label{fig:supplement/starr/all_fairness/hospital_mortality/gender_concept_name}
\end{figure}

\begin{figure}[!htb]
	\centering
	\includegraphics[width=0.9\linewidth]{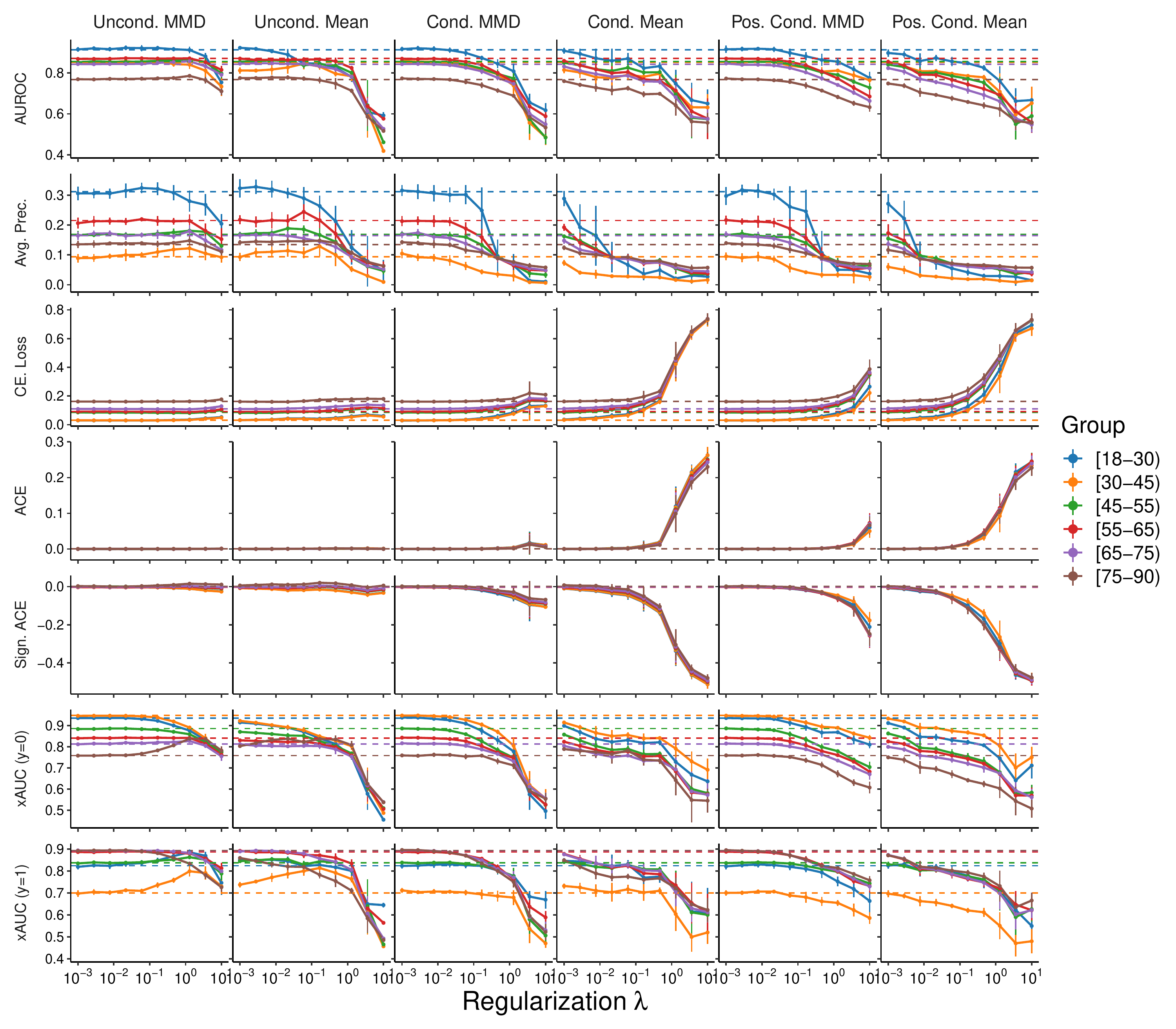}
	\caption{
	    Group-level model performance measures as a function of the extent $\lambda$ that violation of the fairness criterion is penalized when the \textbf{age} group is considered as the sensitive attribute for prediction of \textbf{hospital mortality} in the \textbf{STARR} database. Results shown are the mean $\pm$ SD for the area under the ROC curve (AUROC), average precision (Avg. Prec), the cross entropy loss (CE Loss), the absolute calibration error (ACE), the signed absolute calibration error (Sign. ACE), and cross group ranking performance (xAUC; $\textrm{xAUC}_k^1$ is indicated by (y=1) and $\textrm{xAUC}_k^0$ by (y=0)) for each group for objectives that penalize violation of threshold-free Demographic Parity (Uncond. MMD and Mean), Equalized Odds (Cond. MMD and Mean), and Equal Opportunity (Pos. Cond. MMD and Mean) with MMD- and mean-based penalties. Dashed lines correspond to the mean result for the unpenalized training procedure.
	}
	\label{fig:supplement/starr/all_performance/hospital_mortality/age_group}
\end{figure}

\begin{figure}[!htb]
	\centering
	\includegraphics[width=0.9\linewidth]{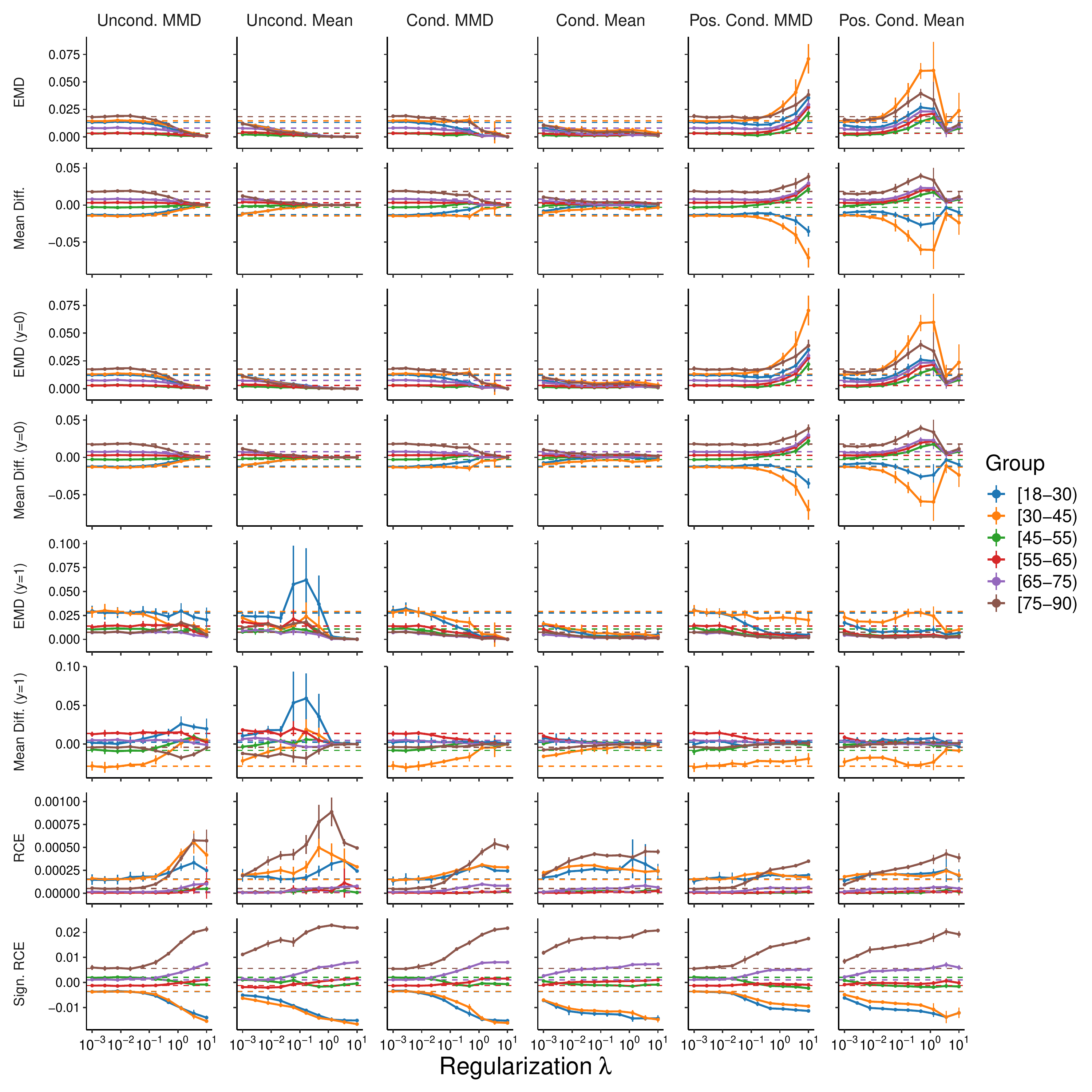}
	\caption{
	    Fairness metrics as a function of the extent $\lambda$ that violation of the fairness criterion is penalized when the \textbf{age} group is considered as the sensitive attribute for prediction of \textbf{hospital mortality} in the \textbf{STARR} database. Results shown are the mean $\pm$ SD for decomposed group-level metrics that assess conditional prediction parity (EMD and Mean Diff.) and relative calibration error (RCE and Sign. RCE) for objectives that penalize violation of threshold-free Demographic Parity (Uncond. MMD and Mean), Equalized Odds (Cond. MMD and Mean), and Equal Opportunity (Pos. Cond. MMD and Mean) on the basis of MMD- and mean-based penalties.  Measures of conditional prediction parity are separately assessed in the whole population and in the strata for which the outcome is and is not observed (suffixed with (y=1) and (y=0), respectively). Dashed lines correspond to the mean result for the unpenalized training procedure.
	}
	\label{fig:supplement/starr/all_fairness/hospital_mortality/age_group}
\end{figure}

\begin{figure}[!htb]
	\centering
	\includegraphics[width=0.9\linewidth]{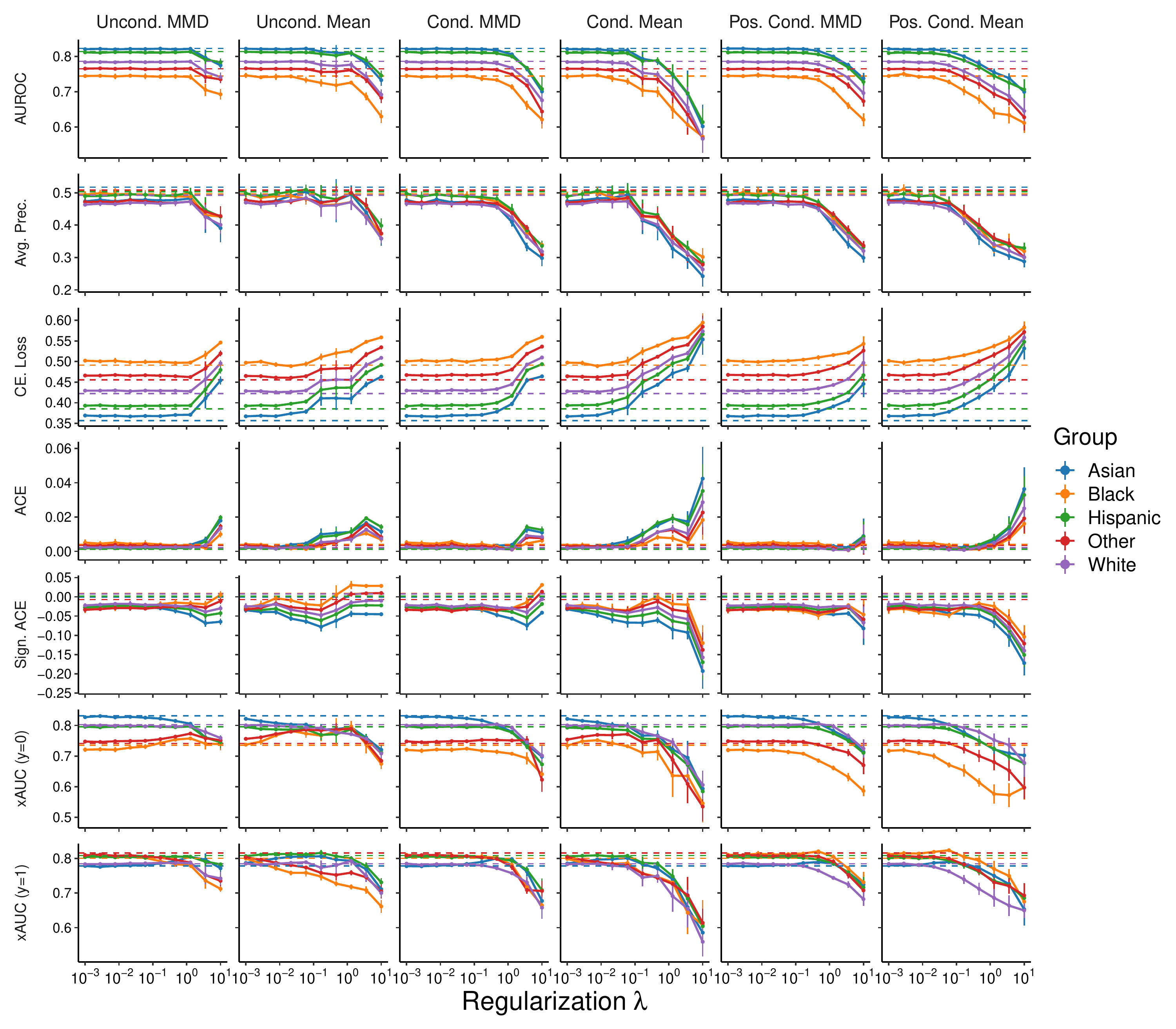}
	\caption{
	    Group-level model performance measures as a function of the extent $\lambda$ that violation of the fairness criterion is penalized when the \textbf{race and ethnicity} category is considered as the sensitive attribute for prediction of \textbf{prolonged length of stay} in the \textbf{STARR} database. Results shown are the mean $\pm$ SD for the area under the ROC curve (AUROC), average precision (Avg. Prec), the cross entropy loss (CE Loss), the absolute calibration error (ACE), the signed absolute calibration error (Sign. ACE), and cross group ranking performance (xAUC; $\textrm{xAUC}_k^1$ is indicated by (y=1) and $\textrm{xAUC}_k^0$ by (y=0)) for each group for objectives that penalize violation of threshold-free Demographic Parity (Uncond. MMD and Mean), Equalized Odds (Cond. MMD and Mean), and Equal Opportunity (Pos. Cond. MMD and Mean) with MMD- and mean-based penalties. Dashed lines correspond to the mean result for the unpenalized training procedure.
	}
	\label{fig:supplement/starr/all_performance/LOS_7/race_eth}
\end{figure}

\begin{figure}[!htb]
	\centering
	\includegraphics[width=0.9\linewidth]{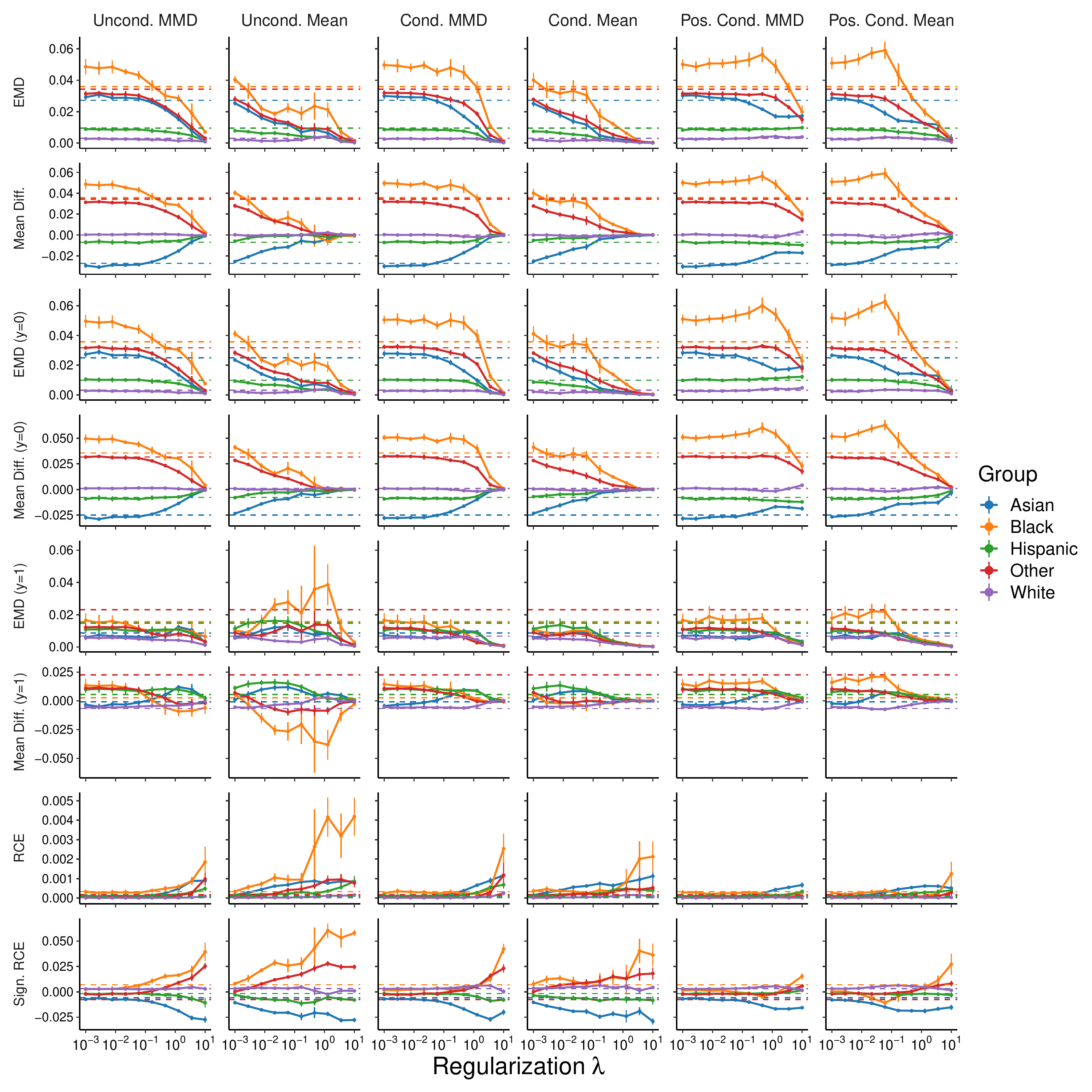}
	\caption{
	    Fairness metrics as a function of the extent $\lambda$ that violation of the fairness criterion is penalized when the \textbf{race and ethnicity} category is considered as the sensitive attribute for prediction of \textbf{prolonged length of stay} in the \textbf{STARR} database. Results shown are the mean $\pm$ SD for decomposed group-level metrics that assess conditional prediction parity (EMD and Mean Diff.) and relative calibration error (RCE and Sign. RCE) for objectives that penalize violation of threshold-free Demographic Parity (Uncond. MMD and Mean), Equalized Odds (Cond. MMD and Mean), and Equal Opportunity (Pos. Cond. MMD and Mean) on the basis of MMD- and mean-based penalties.  Measures of conditional prediction parity are separately assessed in the whole population and in the strata for which the outcome is and is not observed (suffixed with (y=1) and (y=0), respectively). Dashed lines correspond to the mean result for the unpenalized training procedure.
	}
	\label{fig:supplement/starr/all_fairness/LOS_7/race_eth}
\end{figure}

\begin{figure}[!htb]
	\centering
	\includegraphics[width=0.9\linewidth]{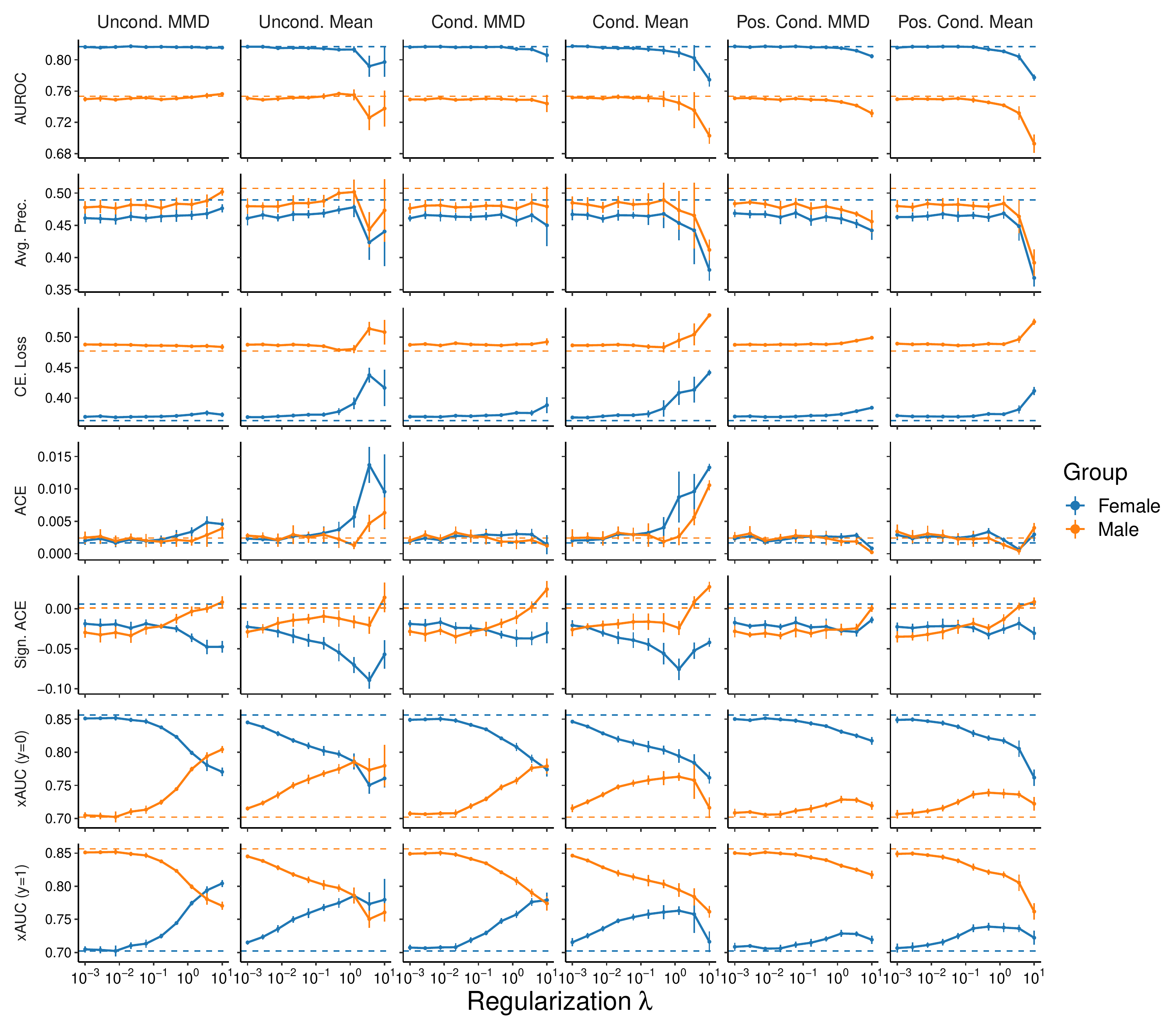}
	\caption{
	    Group-level model performance measures as a function of the extent $\lambda$ that violation of the fairness criterion is penalized when \textbf{sex} is considered as the sensitive attribute for prediction of \textbf{prolonged length of stay} in the \textbf{STARR} database. Results shown are the mean $\pm$ SD for the area under the ROC curve (AUROC), average precision (Avg. Prec), the cross entropy loss (CE Loss), the absolute calibration error (ACE), the signed absolute calibration error (Sign. ACE), and cross group ranking performance (xAUC; $\textrm{xAUC}_k^1$ is indicated by (y=1) and $\textrm{xAUC}_k^0$ by (y=0)) for each group for objectives that penalize violation of threshold-free Demographic Parity (Uncond. MMD and Mean), Equalized Odds (Cond. MMD and Mean), and Equal Opportunity (Pos. Cond. MMD and Mean) with MMD- and mean-based penalties. Dashed lines correspond to the mean result for the unpenalized training procedure.
	}
	\label{fig:supplement/starr/all_performance/LOS_7/gender_concept_name}
\end{figure}

\begin{figure}[!htb]
	\centering
	\includegraphics[width=0.9\linewidth]{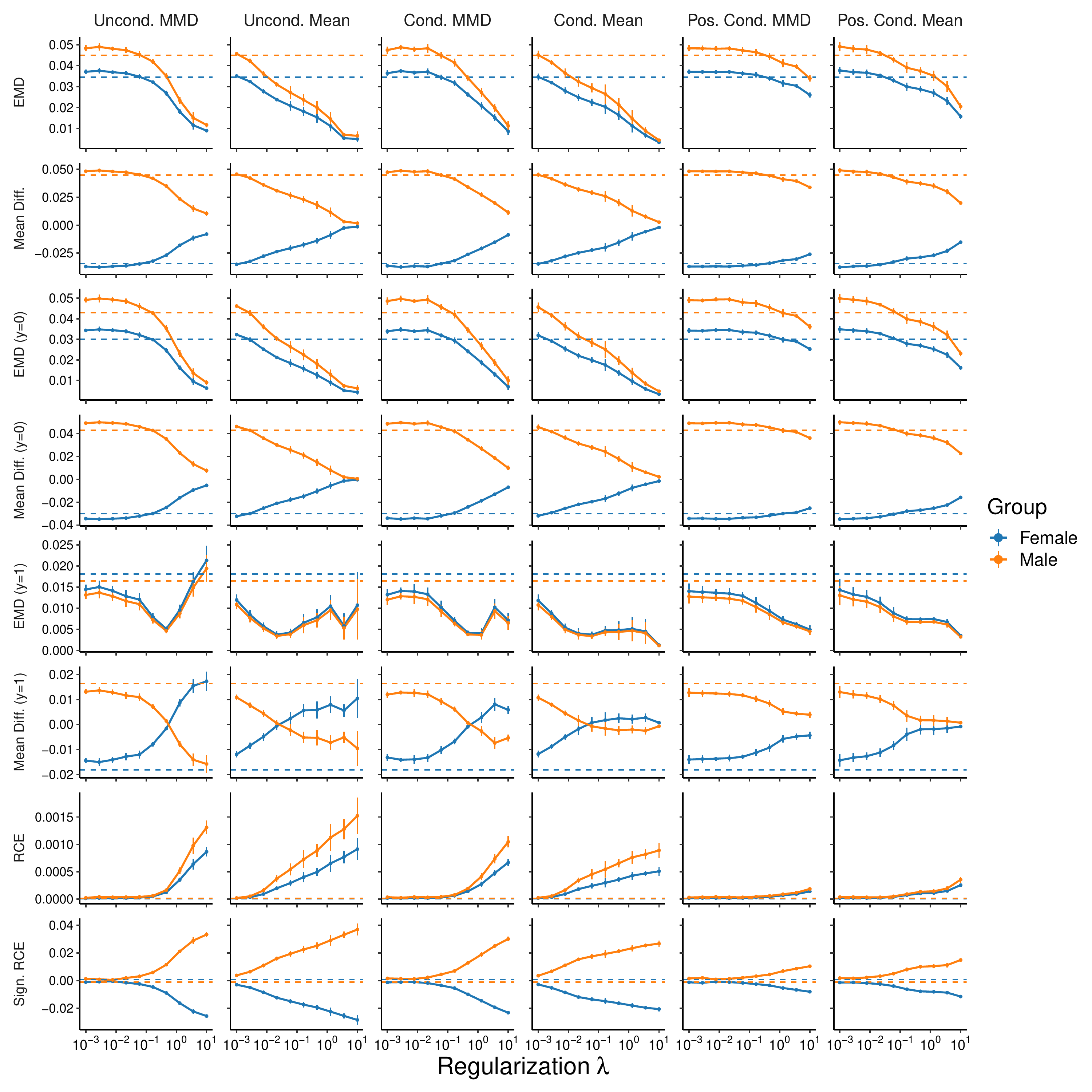}
	\caption{
	    Fairness metrics as a function of the extent $\lambda$ that violation of the fairness criterion is penalized when \textbf{sex} is considered as the sensitive attribute for prediction of \textbf{prolonged length of stay} in the \textbf{STARR} database. Results shown are the mean $\pm$ SD for decomposed group-level metrics that assess conditional prediction parity (EMD and Mean Diff.) and relative calibration error (RCE and Sign. RCE) for objectives that penalize violation of threshold-free Demographic Parity (Uncond. MMD and Mean), Equalized Odds (Cond. MMD and Mean), and Equal Opportunity (Pos. Cond. MMD and Mean) on the basis of MMD- and mean-based penalties.  Measures of conditional prediction parity are separately assessed in the whole population and in the strata for which the outcome is and is not observed (suffixed with (y=1) and (y=0), respectively). Dashed lines correspond to the mean result for the unpenalized training procedure.
	}
	\label{fig:supplement/starr/all_fairness/LOS_7/gender_concept_name}
\end{figure}

\begin{figure}[!htb]
	\centering
	\includegraphics[width=0.9\linewidth]{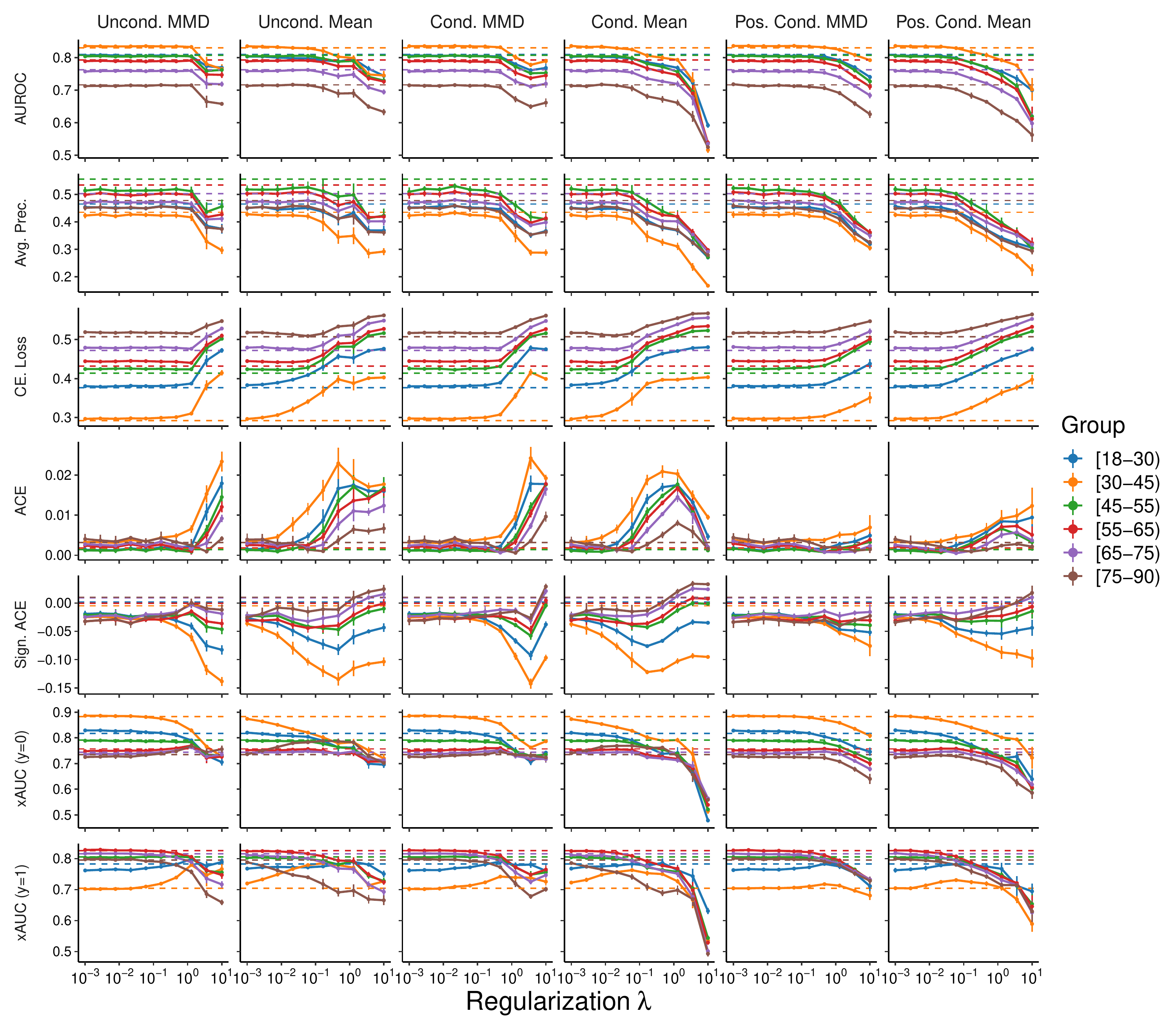}
	\caption{
	    Group-level model performance measures as a function of the extent $\lambda$ that violation of the fairness criterion is penalized when the \textbf{age} group is considered as the sensitive attribute for prediction of \textbf{prolonged length of stay} in the \textbf{STARR} database. Results shown are the mean $\pm$ SD for the area under the ROC curve (AUROC), average precision (Avg. Prec), the cross entropy loss (CE Loss), the absolute calibration error (ACE), the signed absolute calibration error (Sign. ACE), and cross group ranking performance (xAUC; $\textrm{xAUC}_k^1$ is indicated by (y=1) and $\textrm{xAUC}_k^0$ by (y=0)) for each group for objectives that penalize violation of threshold-free Demographic Parity (Uncond. MMD and Mean), Equalized Odds (Cond. MMD and Mean), and Equal Opportunity (Pos. Cond. MMD and Mean) with MMD- and mean-based penalties. Dashed lines correspond to the mean result for the unpenalized training procedure.
	}
	\label{fig:supplement/starr/all_performance/LOS_7/age_group}
\end{figure}

\begin{figure}[!htb]
	\centering
	\includegraphics[width=0.9\linewidth]{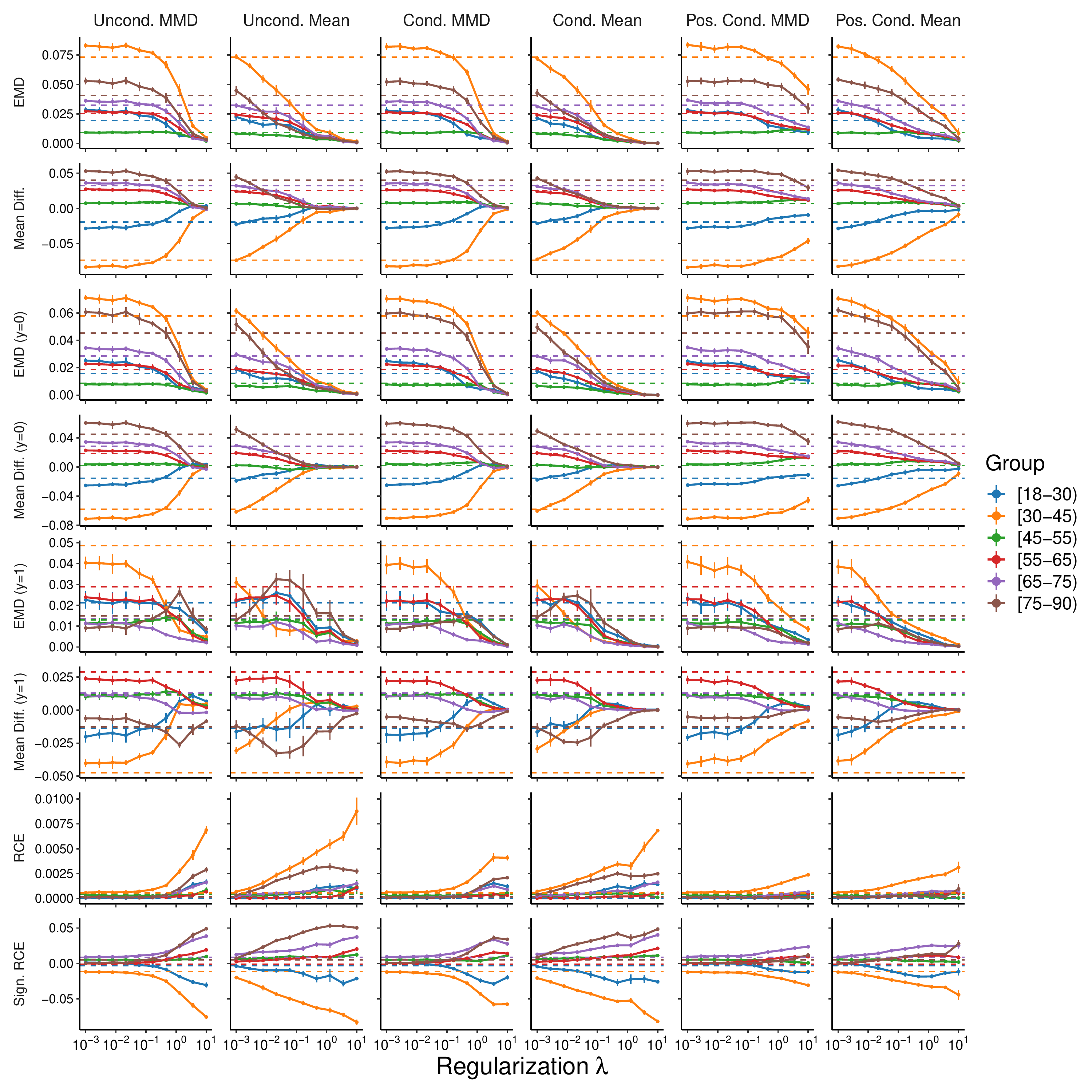}
	\caption{
	    Fairness metrics as a function of the extent $\lambda$ that violation of the fairness criterion is penalized when the \textbf{age} group is considered as the sensitive attribute for prediction of \textbf{prolonged length of stay} in the \textbf{STARR} database. Results shown are the mean $\pm$ SD for decomposed group-level metrics that assess conditional prediction parity (EMD and Mean Diff.) and relative calibration error (RCE and Sign. RCE) for objectives that penalize violation of threshold-free Demographic Parity (Uncond. MMD and Mean), Equalized Odds (Cond. MMD and Mean), and Equal Opportunity (Pos. Cond. MMD and Mean) on the basis of MMD- and mean-based penalties.  Measures of conditional prediction parity are separately assessed in the whole population and in the strata for which the outcome is and is not observed (suffixed with (y=1) and (y=0), respectively). Dashed lines correspond to the mean result for the unpenalized training procedure.
	}
	\label{fig:supplement/starr/all_fairness/LOS_7/age_group}
\end{figure}

\begin{figure}[!htb]
	\centering
	\includegraphics[width=0.9\linewidth]{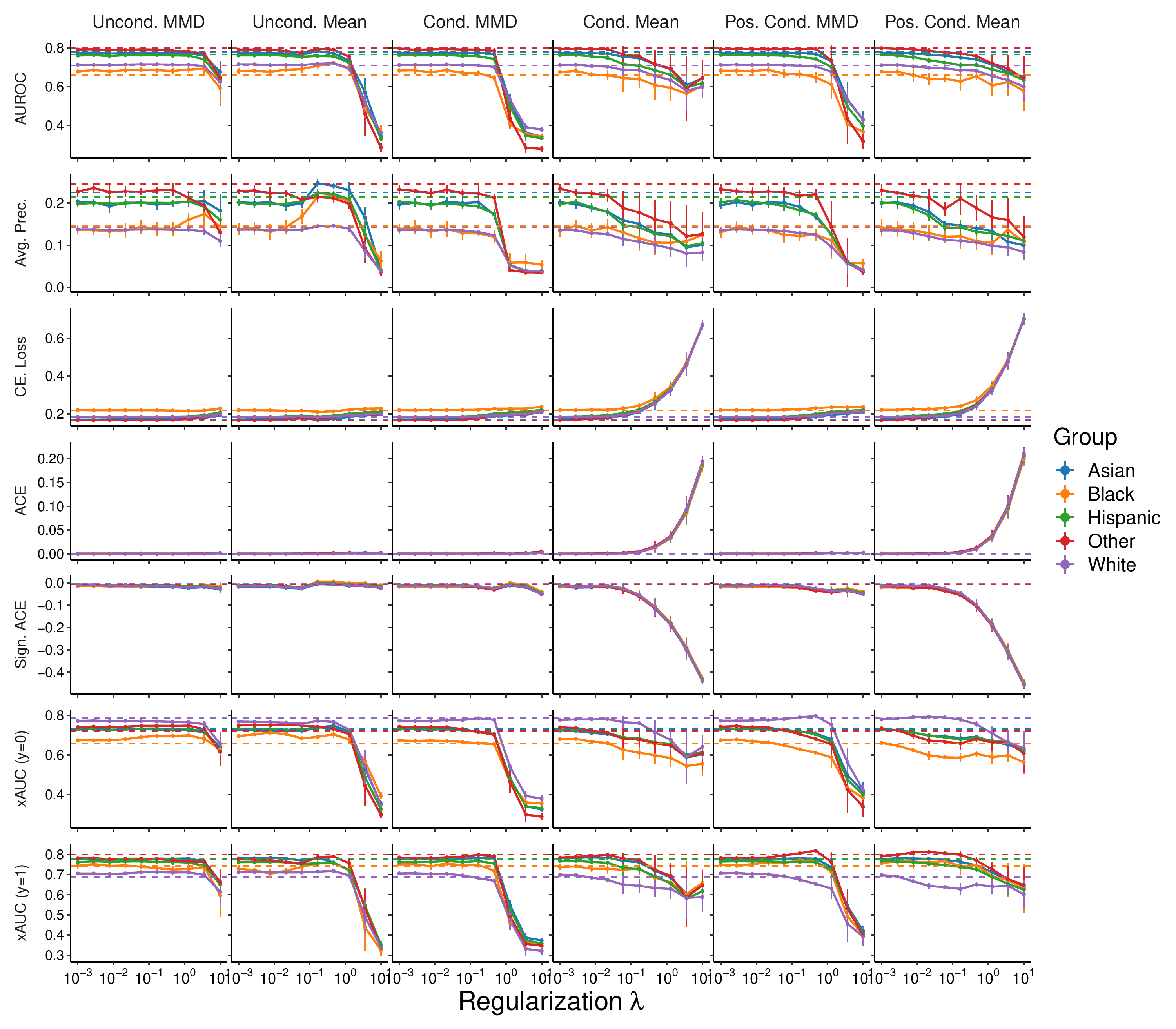}
	\caption{
	    Group-level model performance measures as a function of the extent $\lambda$ that violation of the fairness criterion is penalized when the \textbf{race and ethnicity} category is considered as the sensitive attribute for prediction of \textbf{30-day readmission} in the \textbf{STARR} database. Results shown are the mean $\pm$ SD for the area under the ROC curve (AUROC), average precision (Avg. Prec), the cross entropy loss (CE Loss), the absolute calibration error (ACE), the signed absolute calibration error (Sign. ACE), and cross group ranking performance (xAUC; $\textrm{xAUC}_k^1$ is indicated by (y=1) and $\textrm{xAUC}_k^0$ by (y=0)) for each group for objectives that penalize violation of threshold-free Demographic Parity (Uncond. MMD and Mean), Equalized Odds (Cond. MMD and Mean), and Equal Opportunity (Pos. Cond. MMD and Mean) with MMD- and mean-based penalties. Dashed lines correspond to the mean result for the unpenalized training procedure.
	}
	\label{fig:supplement/starr/all_performance/readmission_30/race_eth}
\end{figure}

\begin{figure}[!htb]
	\centering
	\includegraphics[width=0.9\linewidth]{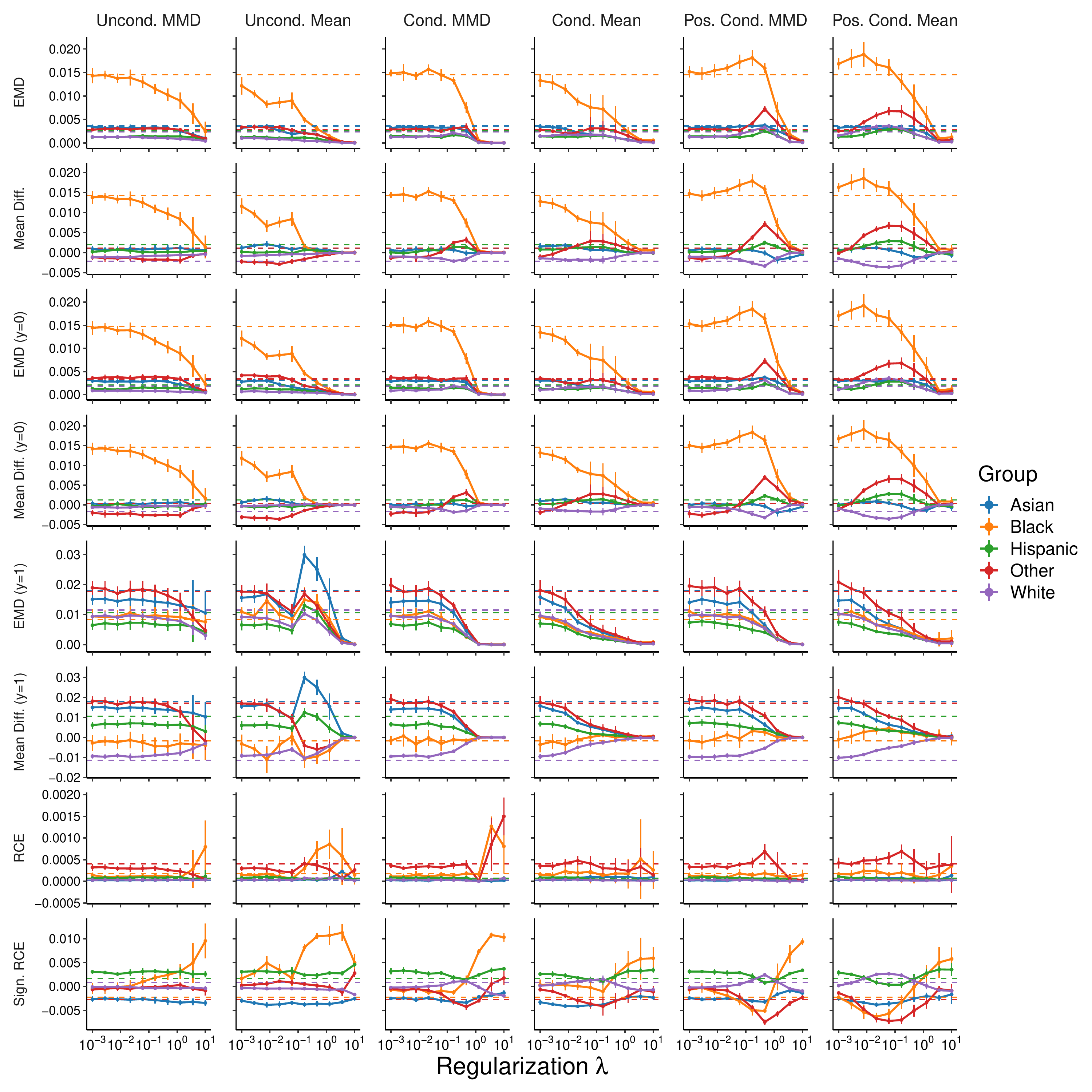}
	\caption{
	    Fairness metrics as a function of the extent $\lambda$ that violation of the fairness criterion is penalized when the \textbf{race and ethnicity} category is considered as the sensitive attribute for prediction of \textbf{30-day readmission} in the \textbf{STARR} database. Results shown are the mean $\pm$ SD for decomposed group-level metrics that assess conditional prediction parity (EMD and Mean Diff.) and relative calibration error (RCE and Sign. RCE) for objectives that penalize violation of threshold-free Demographic Parity (Uncond. MMD and Mean), Equalized Odds (Cond. MMD and Mean), and Equal Opportunity (Pos. Cond. MMD and Mean) on the basis of MMD- and mean-based penalties.  Measures of conditional prediction parity are separately assessed in the whole population and in the strata for which the outcome is and is not observed (suffixed with (y=1) and (y=0), respectively). Dashed lines correspond to the mean result for the unpenalized training procedure.
	}
	\label{fig:supplement/starr/all_fairness/readmission_30/race_eth}
\end{figure}

\begin{figure}[!htb]
	\centering
	\includegraphics[width=0.9\linewidth]{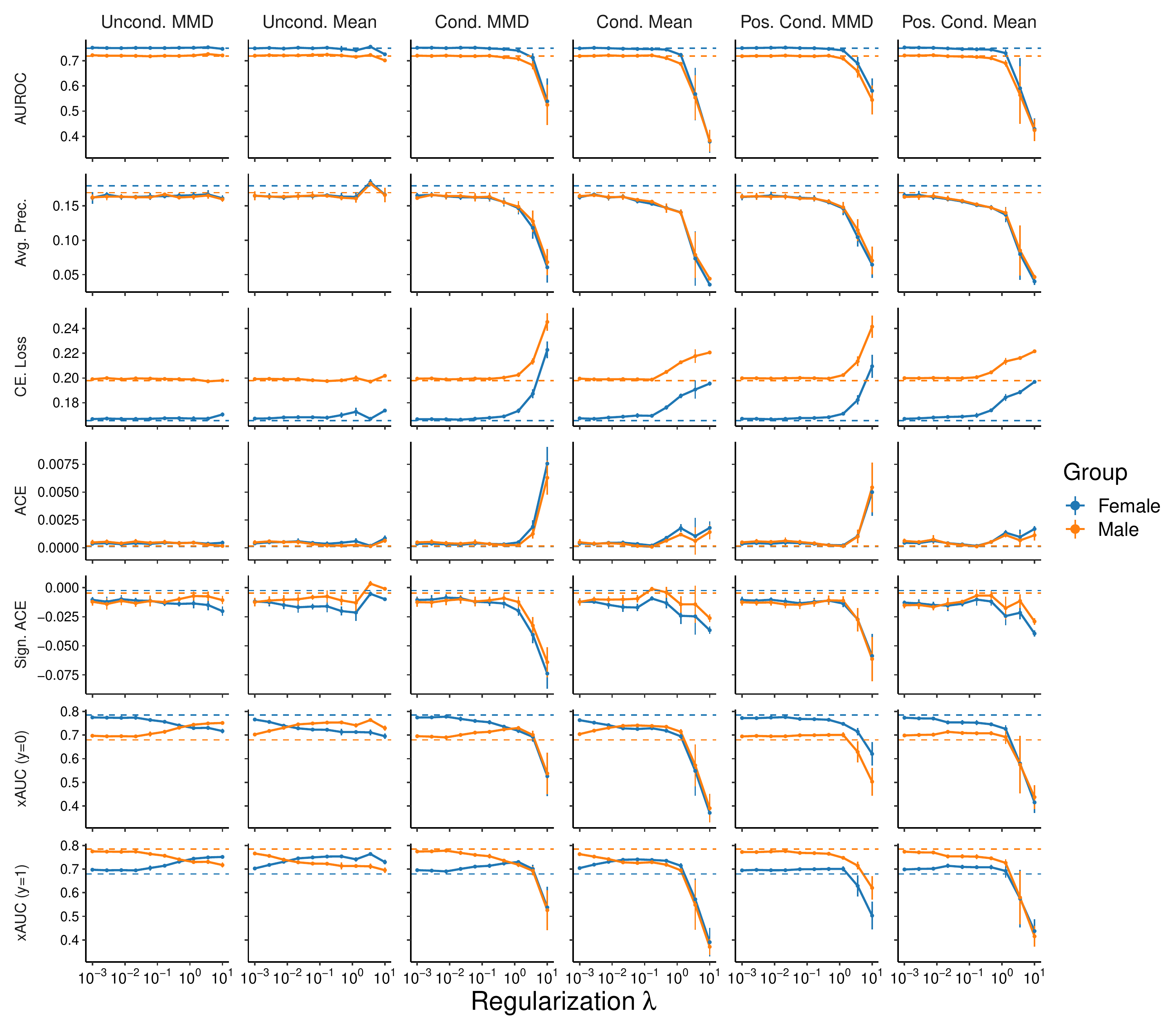}
	\caption{
	    Group-level model performance measures as a function of the extent $\lambda$ that violation of the fairness criterion is penalized when \textbf{sex} is considered as the sensitive attribute for prediction of \textbf{30-day readmission} in the \textbf{STARR} database. Results shown are the mean $\pm$ SD for the area under the ROC curve (AUROC), average precision (Avg. Prec), the cross entropy loss (CE Loss), the absolute calibration error (ACE), the signed absolute calibration error (Sign. ACE), and cross group ranking performance (xAUC; $\textrm{xAUC}_k^1$ is indicated by (y=1) and $\textrm{xAUC}_k^0$ by (y=0)) for each group for objectives that penalize violation of threshold-free Demographic Parity (Uncond. MMD and Mean), Equalized Odds (Cond. MMD and Mean), and Equal Opportunity (Pos. Cond. MMD and Mean) with MMD- and mean-based penalties. Dashed lines correspond to the mean result for the unpenalized training procedure.
	}
	\label{fig:supplement/starr/all_performance/readmission_30/gender_concept_name}
\end{figure}

\begin{figure}[!htb]
	\centering
	\includegraphics[width=0.9\linewidth]{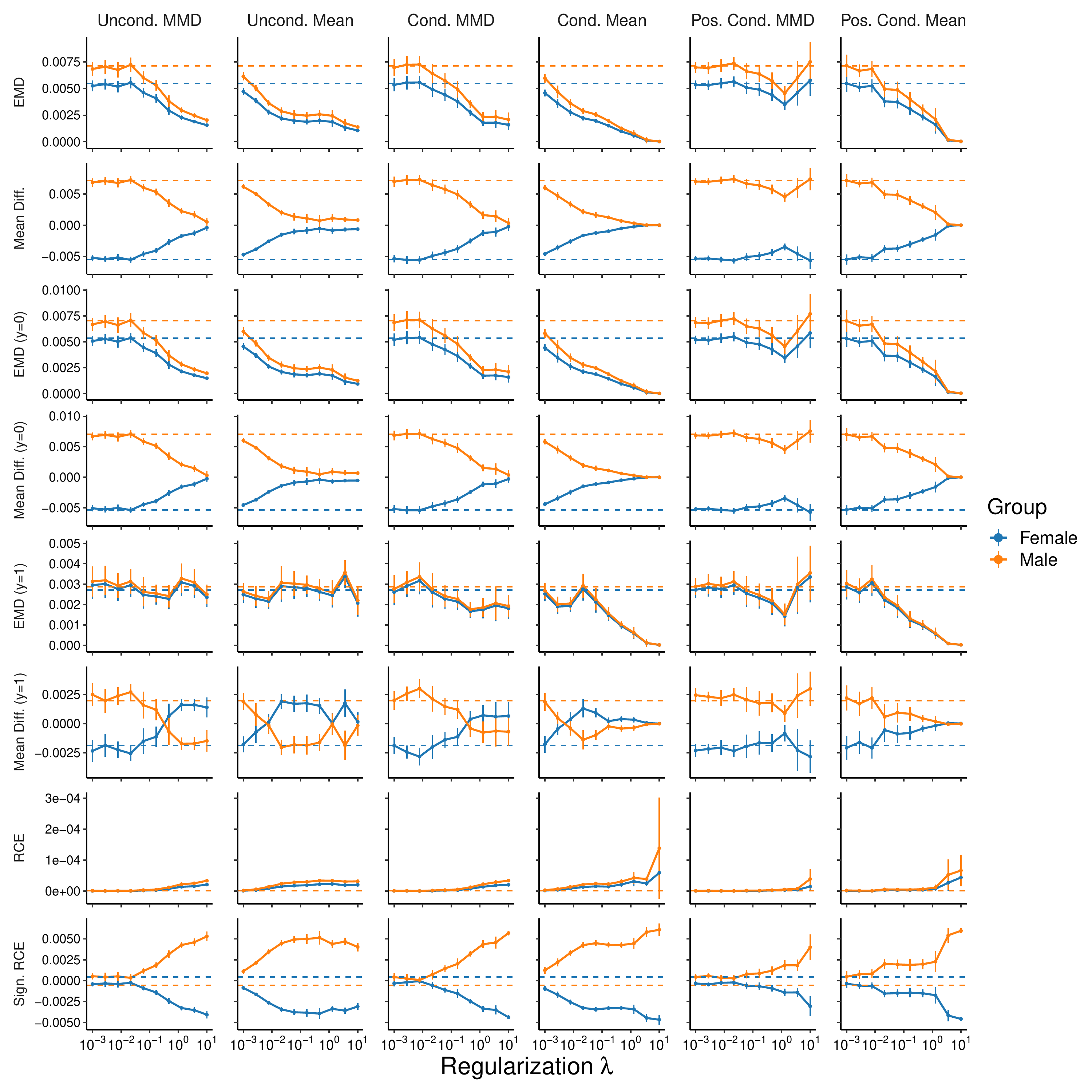}
	\caption{
	    Fairness metrics as a function of the extent $\lambda$ that violation of the fairness criterion is penalized when \textbf{sex} is considered as the sensitive attribute for prediction of \textbf{30-day readmission} in the \textbf{STARR} database. Results shown are the mean $\pm$ SD for decomposed group-level metrics that assess conditional prediction parity (EMD and Mean Diff.) and relative calibration error (RCE and Sign. RCE) for objectives that penalize violation of threshold-free Demographic Parity (Uncond. MMD and Mean), Equalized Odds (Cond. MMD and Mean), and Equal Opportunity (Pos. Cond. MMD and Mean) on the basis of MMD- and mean-based penalties.  Measures of conditional prediction parity are separately assessed in the whole population and in the strata for which the outcome is and is not observed (suffixed with (y=1) and (y=0), respectively). Dashed lines correspond to the mean result for the unpenalized training procedure.
	}
	\label{fig:supplement/starr/all_fairness/readmission_30/gender_concept_name}
\end{figure}

\begin{figure}[!htb]
	\centering
	\includegraphics[width=0.9\linewidth]{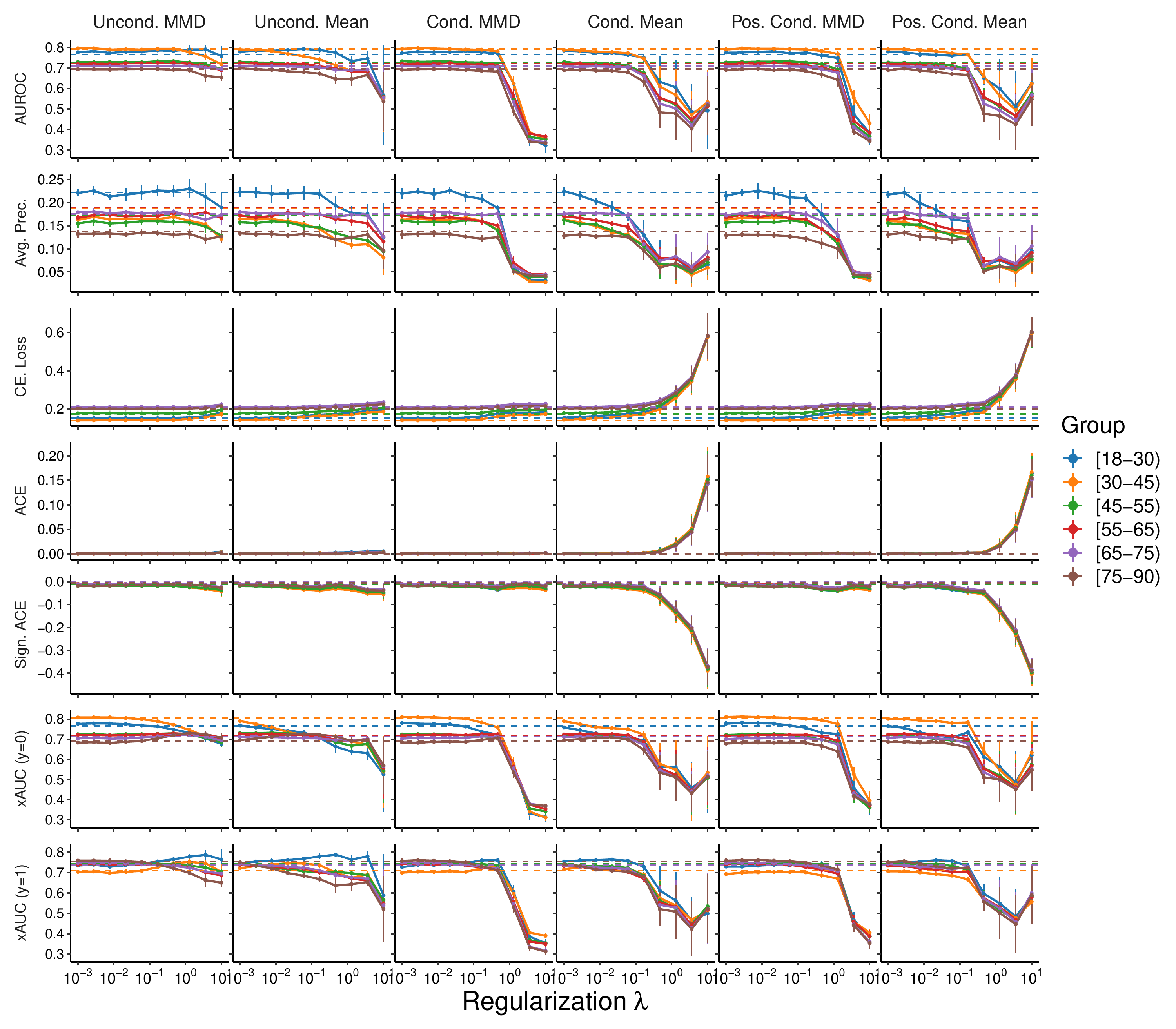}
	\caption{
	    Group-level model performance measures as a function of the extent $\lambda$ that violation of the fairness criterion is penalized when the \textbf{age} group is considered as the sensitive attribute for prediction of \textbf{30-day readmission} in the \textbf{STARR} database. Results shown are the mean $\pm$ SD for the area under the ROC curve (AUROC), average precision (Avg. Prec), the cross entropy loss (CE Loss), the absolute calibration error (ACE), the signed absolute calibration error (Sign. ACE), and cross group ranking performance (xAUC; $\textrm{xAUC}_k^1$ is indicated by (y=1) and $\textrm{xAUC}_k^0$ by (y=0)) for each group for objectives that penalize violation of threshold-free Demographic Parity (Uncond. MMD and Mean), Equalized Odds (Cond. MMD and Mean), and Equal Opportunity (Pos. Cond. MMD and Mean) with MMD- and mean-based penalties. Dashed lines correspond to the mean result for the unpenalized training procedure.
	}
	\label{fig:supplement/starr/all_performance/readmission_30/age_group}
\end{figure}

\begin{figure}[!htb]
	\centering
	\includegraphics[width=0.9\linewidth]{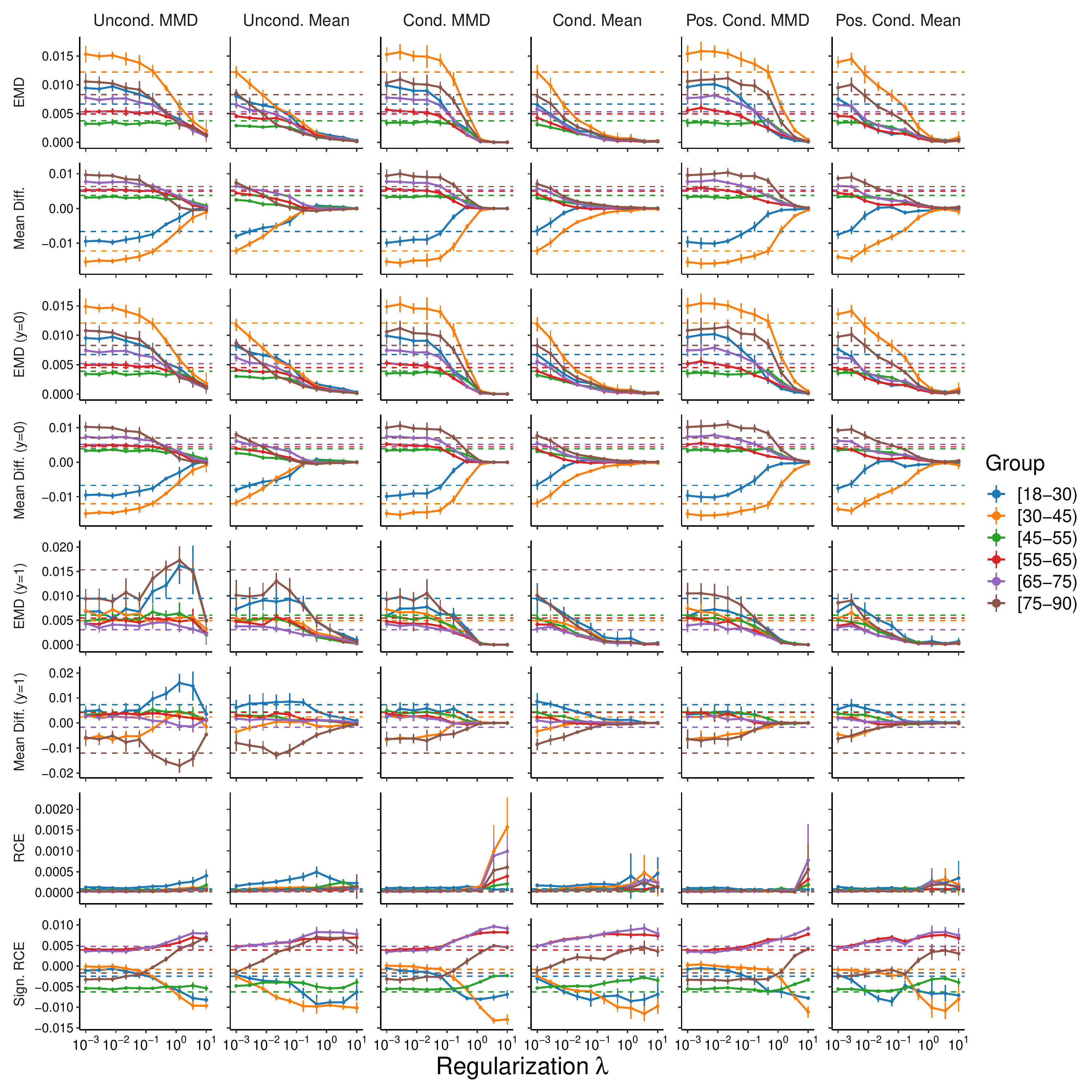}
	\caption{
	    Fairness metrics as a function of the extent $\lambda$ that violation of the fairness criterion is penalized when the \textbf{age} group is considered as the sensitive attribute for prediction of \textbf{30-day readmission} in the \textbf{STARR} database. Results shown are the mean $\pm$ SD for decomposed group-level metrics that assess conditional prediction parity (EMD and Mean Diff.) and relative calibration error (RCE and Sign. RCE) for objectives that penalize violation of threshold-free Demographic Parity (Uncond. MMD and Mean), Equalized Odds (Cond. MMD and Mean), and Equal Opportunity (Pos. Cond. MMD and Mean) on the basis of MMD- and mean-based penalties.  Measures of conditional prediction parity are separately assessed in the whole population and in the strata for which the outcome is and is not observed (suffixed with (y=1) and (y=0), respectively). 
	    Dashed lines correspond to the mean result for the unpenalized training procedure.
	}
	\label{fig:supplement/starr/all_fairness/readmission_30/age_group}
\end{figure}

\clearpage
\subsection{Optum CDM}
\begin{figure}[!htb]
	\centering
	\includegraphics[width=0.9\linewidth]{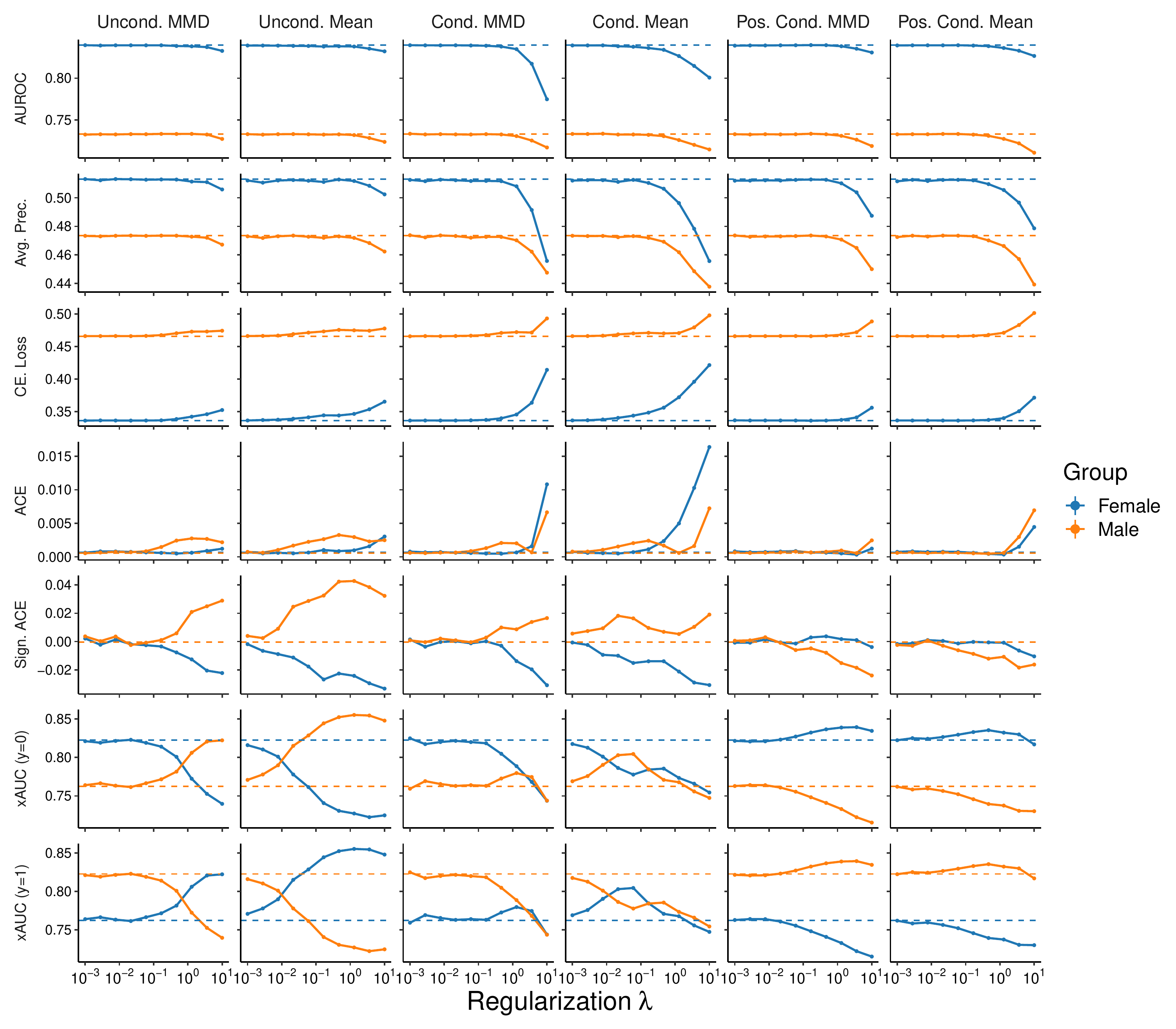}
	\caption{
	    Group-level model performance measures as a function of the extent $\lambda$ that violation of the fairness criterion is penalized when \textbf{sex} is considered as the sensitive attribute for prediction of \textbf{prolonged length of stay} in the \textbf{Optum CDM} database. Results shown are the area under the ROC curve (AUROC), average precision (Avg. Prec), the cross entropy loss (CE Loss), the absolute calibration error (ACE), the signed absolute calibration error (Sign. ACE), and cross group ranking performance (xAUC; $\textrm{xAUC}_k^1$ is indicated by (y=1) and $\textrm{xAUC}_k^0$ by (y=0)) for each group for objectives that penalize violation of threshold-free Demographic Parity (Uncond. MMD and Mean), Equalized Odds (Cond. MMD and Mean), and Equal Opportunity (Pos. Cond. MMD and Mean) with MMD- and mean-based penalties. Dashed lines correspond to the result for the unpenalized training procedure.
	}
	\label{fig:supplement/optum/all_performance/LOS_7/gender_concept_name}
\end{figure}

\begin{figure}[!htb]
	\centering
	\includegraphics[width=0.9\linewidth]{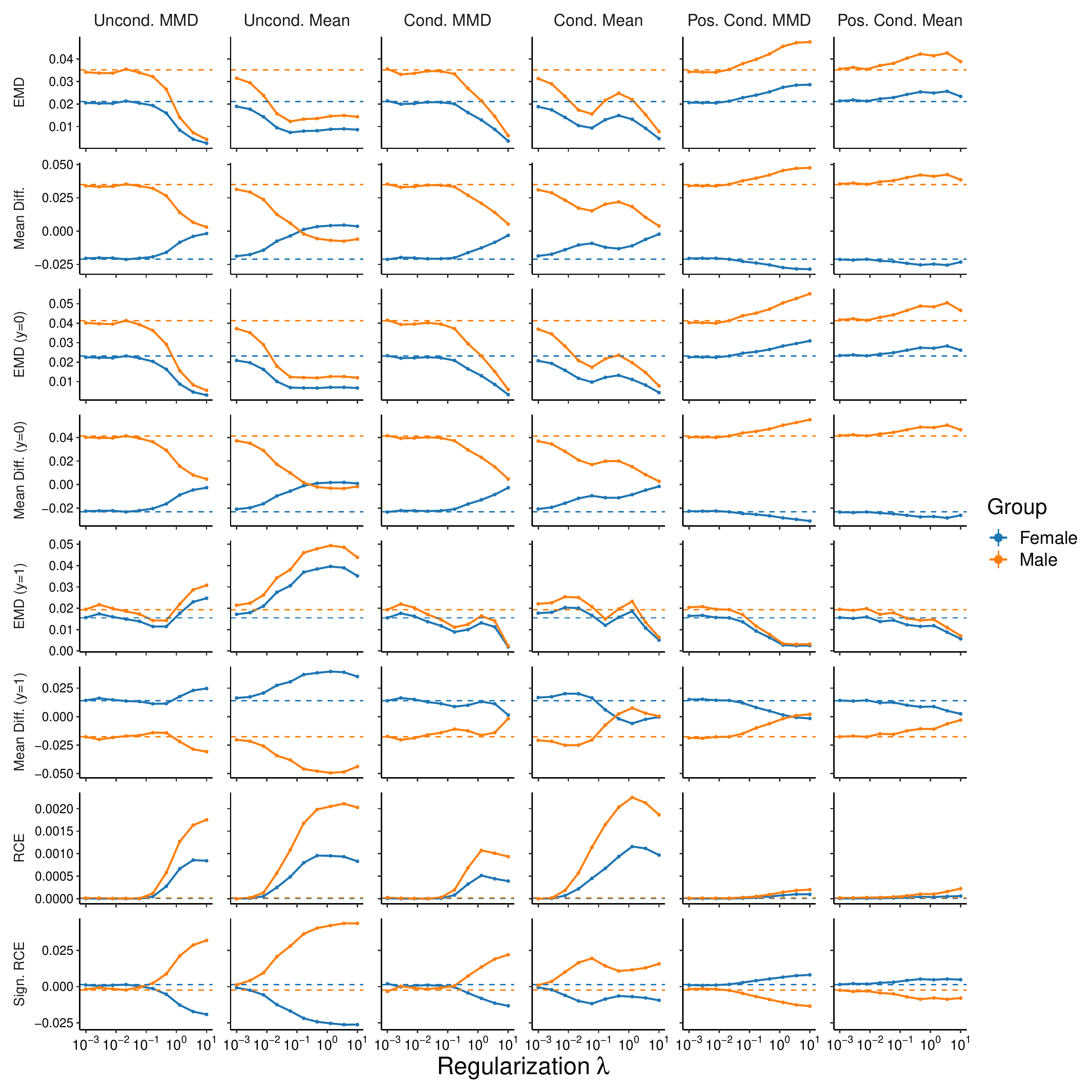}
	\caption{
	    Fairness metrics as a function of the extent $\lambda$ that violation of the fairness criterion is penalized when \textbf{sex} is considered as the sensitive attribute for prediction of \textbf{prolonged length of stay} in the \textbf{Optum CDM} database. Results shown are decomposed group-level metrics that assess conditional prediction parity (EMD and Mean Diff.) and relative calibration error (RCE and Sign. RCE) for objectives that penalize violation of threshold-free Demographic Parity (Uncond. MMD and Mean), Equalized Odds (Cond. MMD and Mean), and Equal Opportunity (Pos. Cond. MMD and Mean) on the basis of MMD- and mean-based penalties.  Measures of conditional prediction parity are separately assessed in the whole population and in the strata for which the outcome is and is not observed (suffixed with (y=1) and (y=0), respectively). 
	    Dashed lines correspond to the result for the unpenalized training procedure.
	}
	\label{fig:supplement/optum/all_fairness/LOS_7/gender_concept_name}
\end{figure}

\begin{figure}[!htb]
	\centering
	\includegraphics[width=0.9\linewidth]{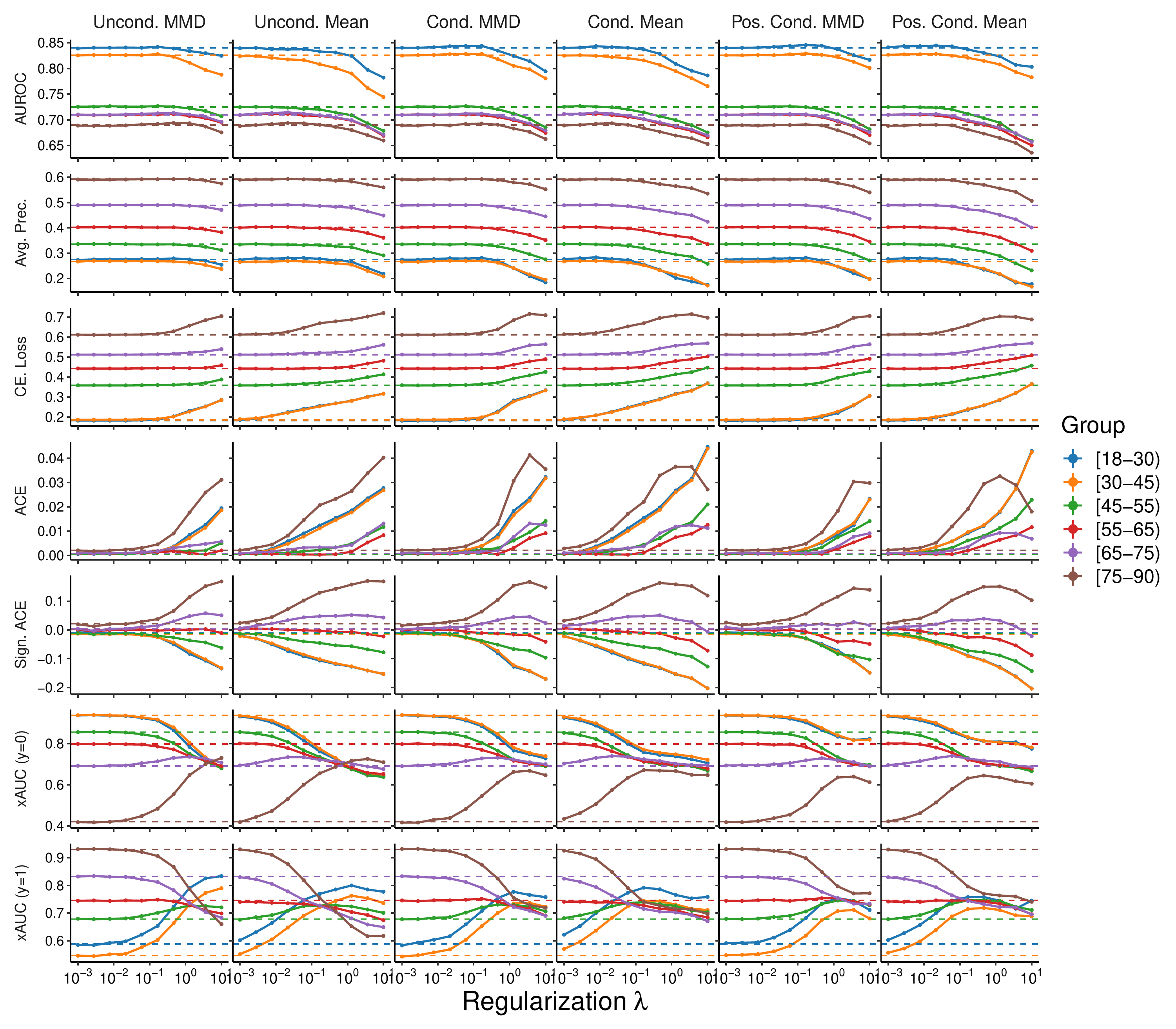}
	\caption{
	    Group-level model performance measures as a function of the extent $\lambda$ that violation of the fairness criterion is penalized when the \textbf{age} group is considered as the sensitive attribute for prediction of \textbf{prolonged length of stay} in the \textbf{Optum CDM} database. Results shown are the area under the ROC curve (AUROC), average precision (Avg. Prec), the cross entropy loss (CE Loss), the absolute calibration error (ACE), the signed absolute calibration error (Sign. ACE), and cross group ranking performance (xAUC; $\textrm{xAUC}_k^1$ is indicated by (y=1) and $\textrm{xAUC}_k^0$ by (y=0)) for each group for objectives that penalize violation of threshold-free Demographic Parity (Uncond. MMD and Mean), Equalized Odds (Cond. MMD and Mean), and Equal Opportunity (Pos. Cond. MMD and Mean) with MMD- and mean-based penalties. Dashed lines correspond to the result for the unpenalized training procedure.
	}
	\label{fig:supplement/optum/all_performance/LOS_7/age_group}
\end{figure}

\begin{figure}[!htb]
	\centering
	\includegraphics[width=0.9\linewidth]{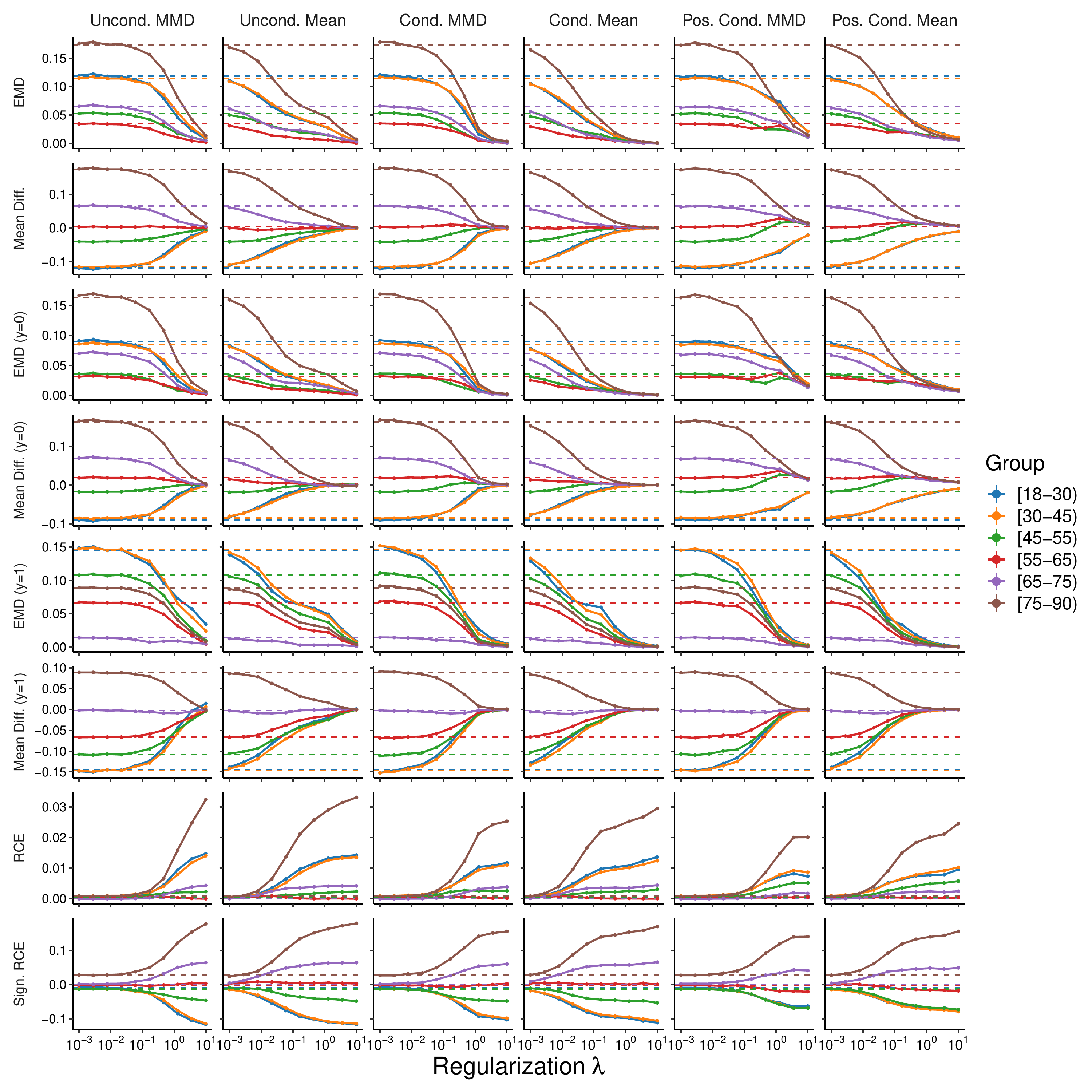}
	\caption{
	    Fairness metrics as a function of the extent $\lambda$ that violation of the fairness criterion is penalized when the \textbf{age} group is considered as the sensitive attribute for prediction of \textbf{prolonged length of stay} in the \textbf{Optum CDM} database. Results shown are decomposed group-level metrics that assess conditional prediction parity (EMD and Mean Diff.) and relative calibration error (RCE and Sign. RCE) for objectives that penalize violation of threshold-free Demographic Parity (Uncond. MMD and Mean), Equalized Odds (Cond. MMD and Mean), and Equal Opportunity (Pos. Cond. MMD and Mean) on the basis of MMD- and mean-based penalties.  Measures of conditional prediction parity are separately assessed in the whole population and in the strata for which the outcome is and is not observed (suffixed with (y=1) and (y=0), respectively). 
	    Dashed lines correspond to the result for the unpenalized training procedure.
	}
	\label{fig:supplement/optum/all_fairness/LOS_7/age_group}
\end{figure}

\begin{figure}[!htb]
	\centering
	\includegraphics[width=0.9\linewidth]{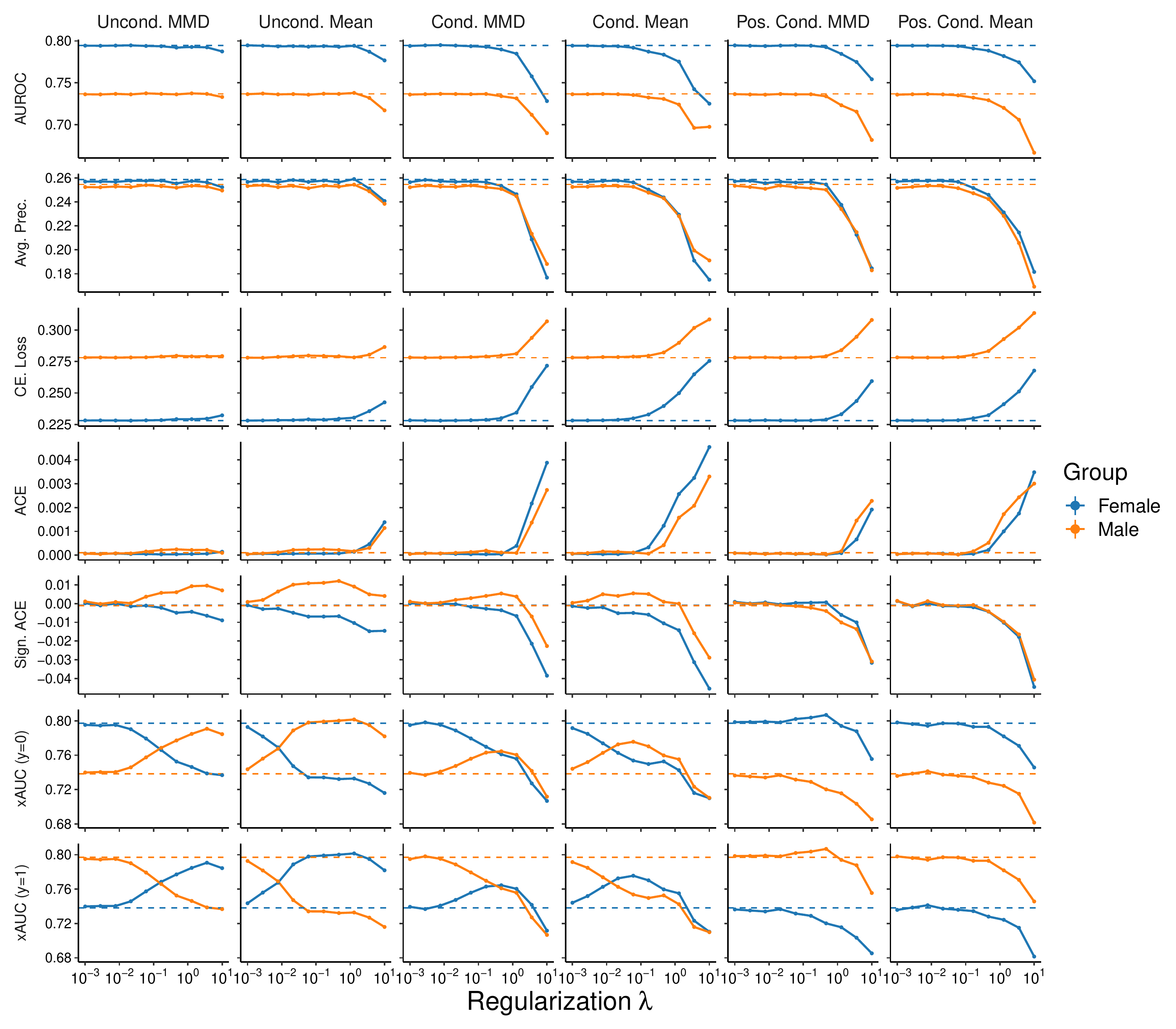}
	\caption{
	    Group-level model performance measures as a function of the extent $\lambda$ that violation of the fairness criterion is penalized when \textbf{sex} is considered as the sensitive attribute for prediction of \textbf{30-day readmission} in the \textbf{Optum CDM} database. Results shown are the area under the ROC curve (AUROC), average precision (Avg. Prec), the cross entropy loss (CE Loss), the absolute calibration error (ACE), the signed absolute calibration error (Sign. ACE), and cross group ranking performance (xAUC; $\textrm{xAUC}_k^1$ is indicated by (y=1) and $\textrm{xAUC}_k^0$ by (y=0)) for each group for objectives that penalize violation of threshold-free Demographic Parity (Uncond. MMD and Mean), Equalized Odds (Cond. MMD and Mean), and Equal Opportunity (Pos. Cond. MMD and Mean) with MMD- and mean-based penalties. Dashed lines correspond to the result for the unpenalized training procedure.
	}
	\label{fig:supplement/optum/all_performance/readmission_30/gender_concept_name}
\end{figure}

\begin{figure}[!htb]
	\centering
	\includegraphics[width=0.9\linewidth]{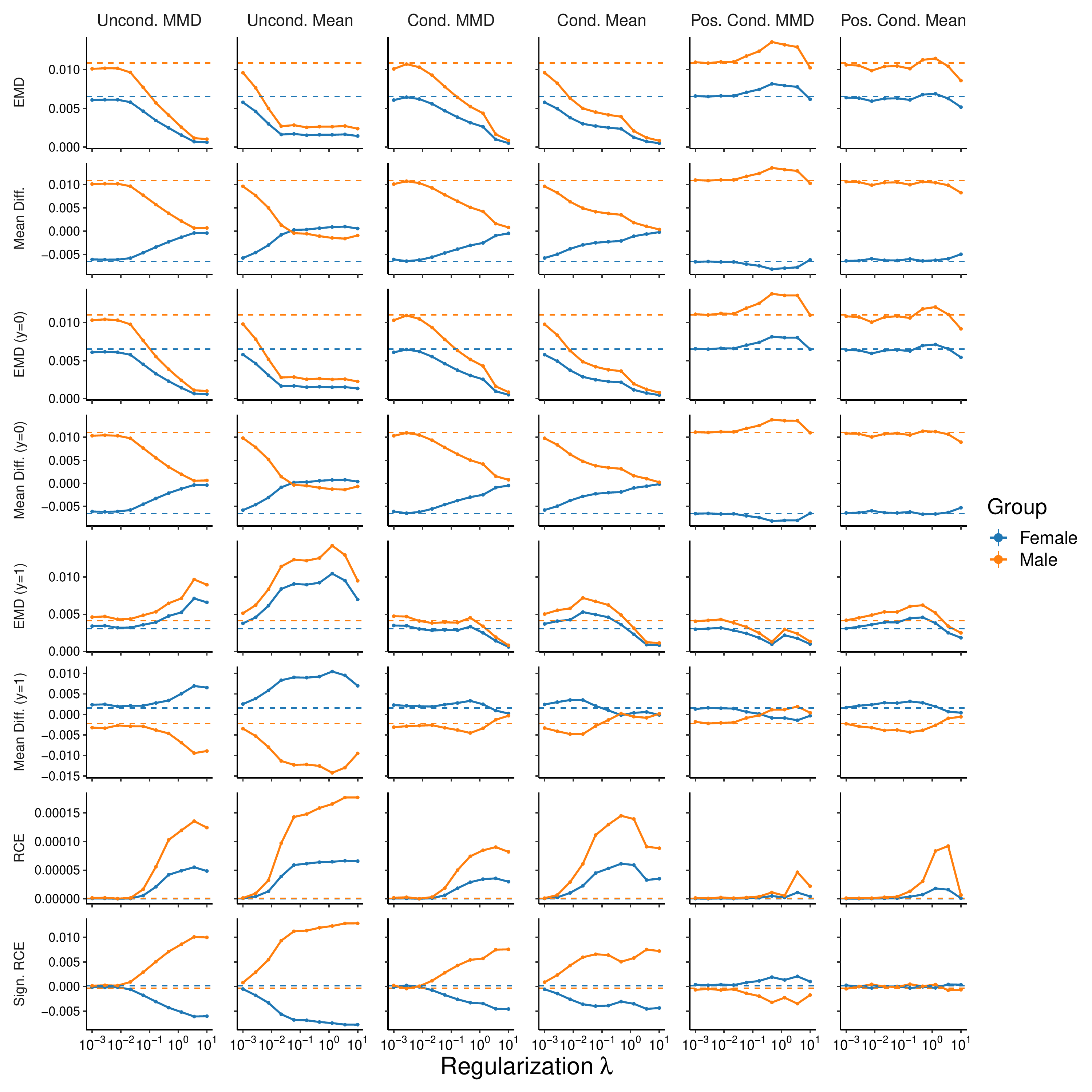}
	\caption{
	    Fairness metrics as a function of the extent $\lambda$ that violation of the fairness criterion is penalized when \textbf{sex} is considered as the sensitive attribute for prediction of \textbf{30-day readmission} in the \textbf{Optum CDM} database. Results shown are decomposed group-level metrics that assess conditional prediction parity (EMD and Mean Diff.) and relative calibration error (RCE and Sign. RCE) for objectives that penalize violation of threshold-free Demographic Parity (Uncond. MMD and Mean), Equalized Odds (Cond. MMD and Mean), and Equal Opportunity (Pos. Cond. MMD and Mean) on the basis of MMD- and mean-based penalties.  Measures of conditional prediction parity are separately assessed in the whole population and in the strata for which the outcome is and is not observed (suffixed with (y=1) and (y=0), respectively). 
	    Dashed lines correspond to the result for the unpenalized training procedure.
	}
	\label{fig:supplement/optum/all_fairness/readmission_30/gender_concept_name}
\end{figure}

\begin{figure}[!htb]
	\centering
	\includegraphics[width=0.9\linewidth]{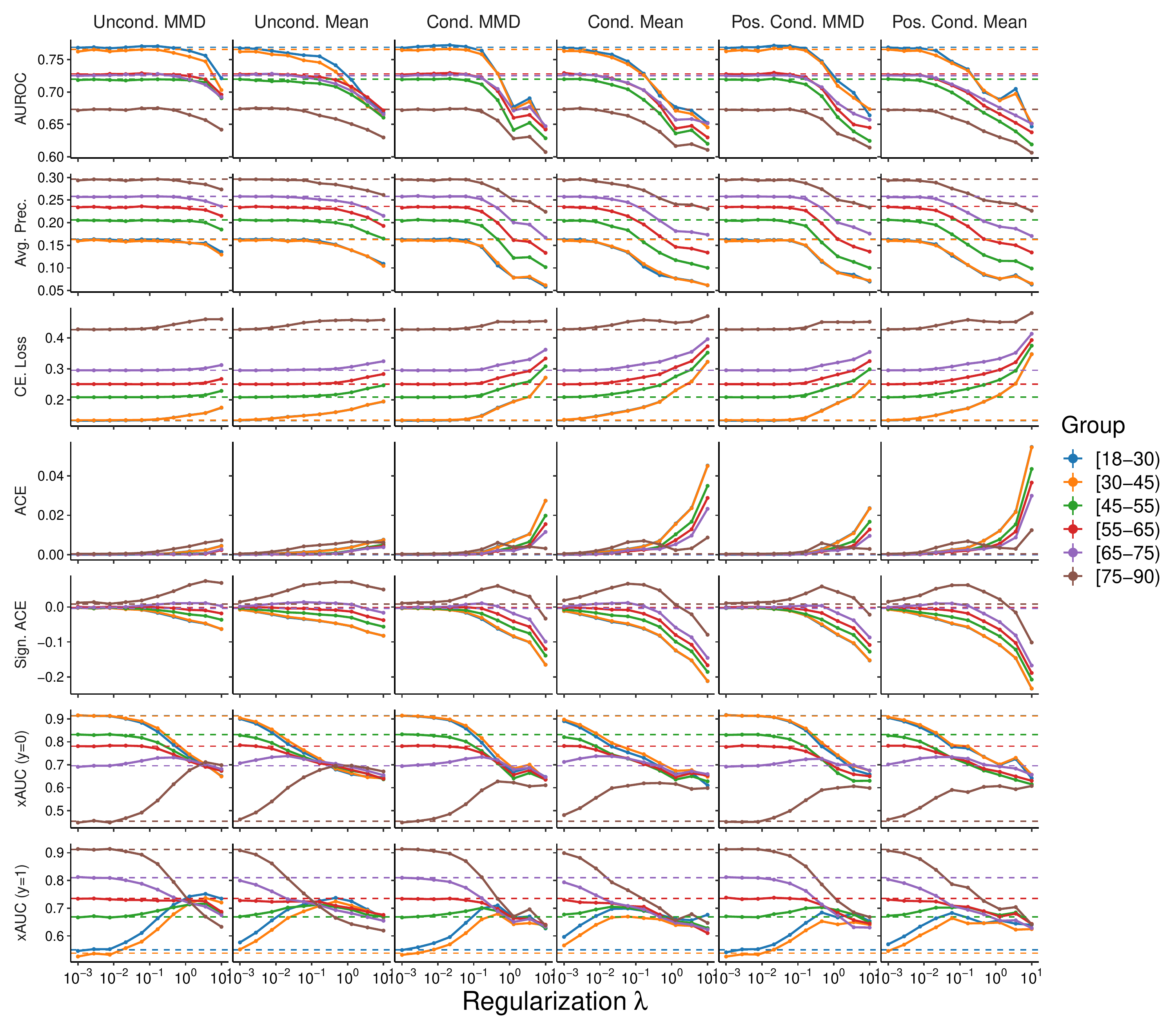}
	\caption{
	    Group-level model performance measures as a function of the extent $\lambda$ that violation of the fairness criterion is penalized when the \textbf{age} group is considered as the sensitive attribute for prediction of \textbf{30-day readmission} in the \textbf{Optum CDM} database. Results shown are the area under the ROC curve (AUROC), average precision (Avg. Prec), the cross entropy loss (CE Loss), the absolute calibration error (ACE), the signed absolute calibration error (Sign. ACE), and cross group ranking performance (xAUC; $\textrm{xAUC}_k^1$ is indicated by (y=1) and $\textrm{xAUC}_k^0$ by (y=0)) for each group for objectives that penalize violation of threshold-free Demographic Parity (Uncond. MMD and Mean), Equalized Odds (Cond. MMD and Mean), and Equal Opportunity (Pos. Cond. MMD and Mean) with MMD- and mean-based penalties. Dashed lines correspond to the result for the unpenalized training procedure.
	}
	\label{fig:supplement/optum/all_performance/readmission_30/age_group}
\end{figure}

\begin{figure}[!htb]
	\centering
	\includegraphics[width=0.9\linewidth]{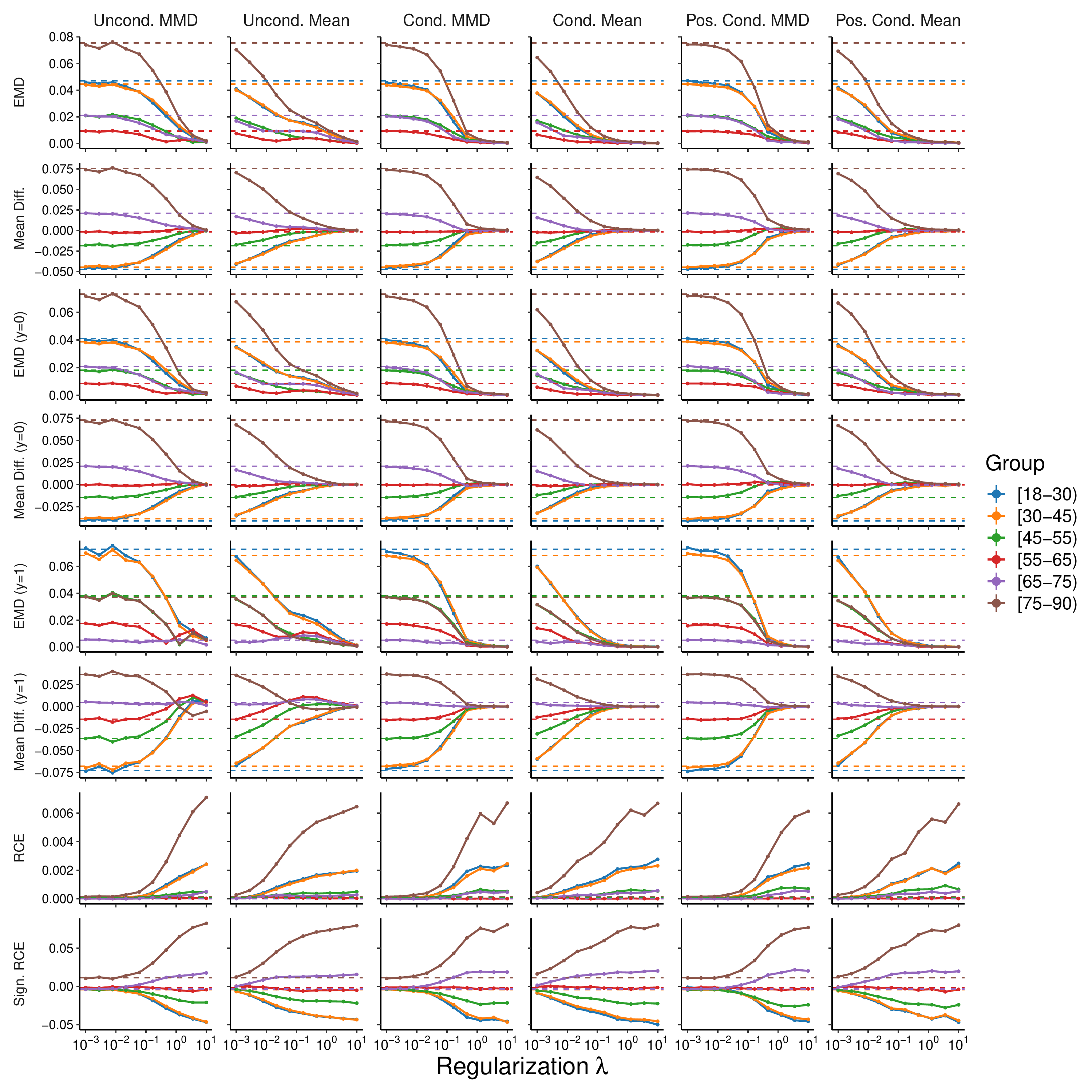}
	\caption{
	    Fairness metrics as a function of the extent $\lambda$ that violation of the fairness criterion is penalized when the \textbf{age} group is considered as the sensitive attribute for prediction of \textbf{30-day readmission} in the \textbf{Optum CDM} database. Results shown are decomposed group-level metrics that assess conditional prediction parity (EMD and Mean Diff.) and relative calibration error (RCE and Sign. RCE) for objectives that penalize violation of threshold-free Demographic Parity (Uncond. MMD and Mean), Equalized Odds (Cond. MMD and Mean), and Equal Opportunity (Pos. Cond. MMD and Mean) on the basis of MMD- and mean-based penalties.  Measures of conditional prediction parity are separately assessed in the whole population and in the strata for which the outcome is and is not observed (suffixed with (y=1) and (y=0), respectively). 
	    Dashed lines correspond to the result for the unpenalized training procedure.
	}
	\label{fig:supplement/optum/all_fairness/readmission_30/age_group}
\end{figure}

\clearpage
\subsection{MIMIC-III}
\begin{figure}[!htb]
	\centering
	\includegraphics[width=0.9\linewidth]{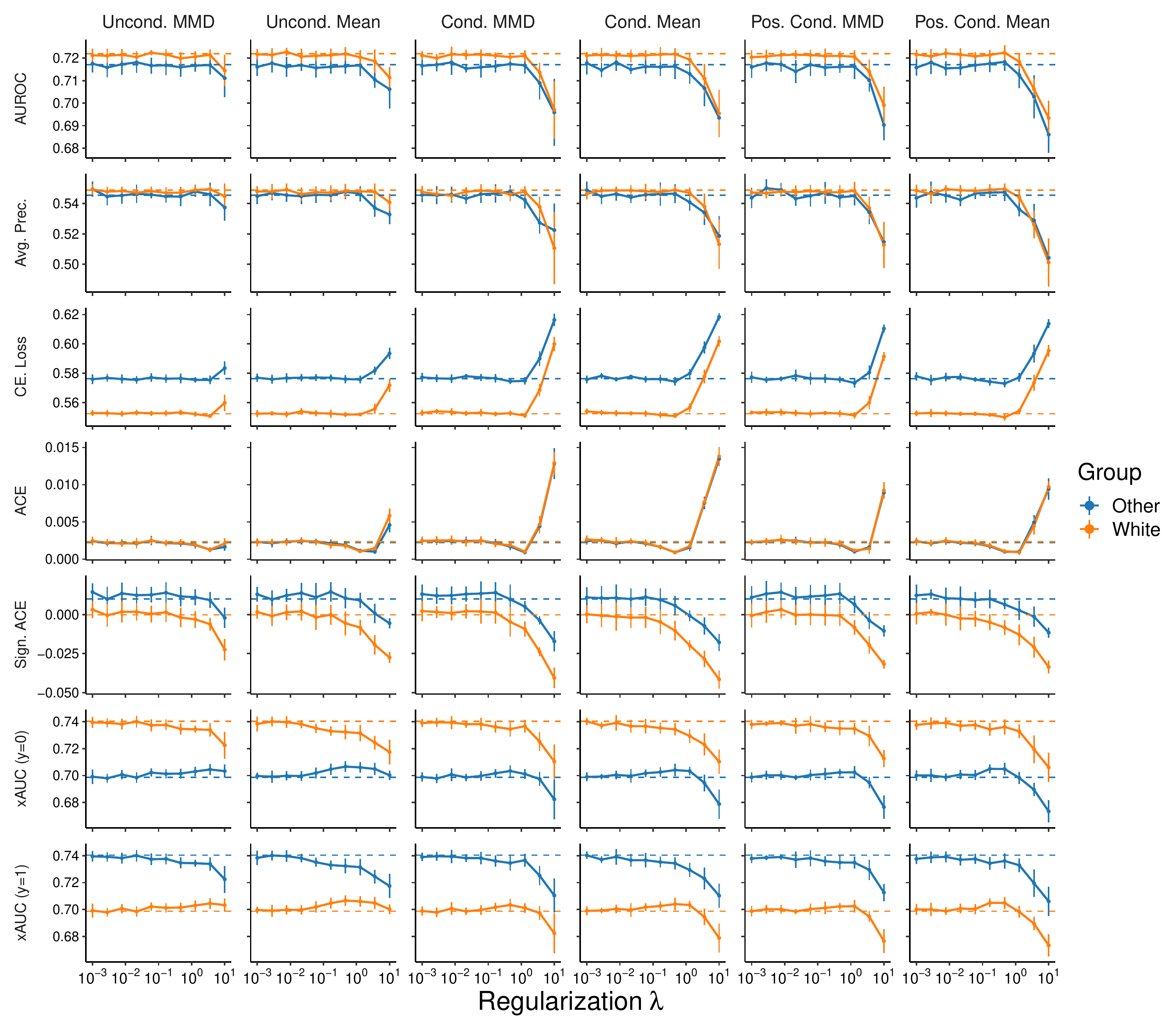}
	\caption{
	    Group-level model performance measures as a function of the extent $\lambda$ that violation of the fairness criterion is penalized when the \textbf{race and ethnicity} category is considered as the sensitive attribute for prediction of \textbf{ICU length of stay greater than 3 days} in the \textbf{MIMIC-III} database. Results shown are the mean $\pm$ SD for the area under the ROC curve (AUROC), average precision (Avg. Prec), the cross entropy loss (CE Loss), the absolute calibration error (ACE), the signed absolute calibration error (Sign. ACE), and cross group ranking performance (xAUC; $\textrm{xAUC}_k^1$ is indicated by (y=1) and $\textrm{xAUC}_k^0$ by (y=0)) for each group for objectives that penalize violation of threshold-free Demographic Parity (Uncond. MMD and Mean), Equalized Odds (Cond. MMD and Mean), and Equal Opportunity (Pos. Cond. MMD and Mean) with MMD- and mean-based penalties. Dashed lines correspond to the mean result for the unpenalized training procedure.
	}
	\label{fig:supplement/mimic/all_performance/los_icu_3days/race_eth}
\end{figure}

\begin{figure}[!htb]
	\centering
	\includegraphics[width=0.9\linewidth]{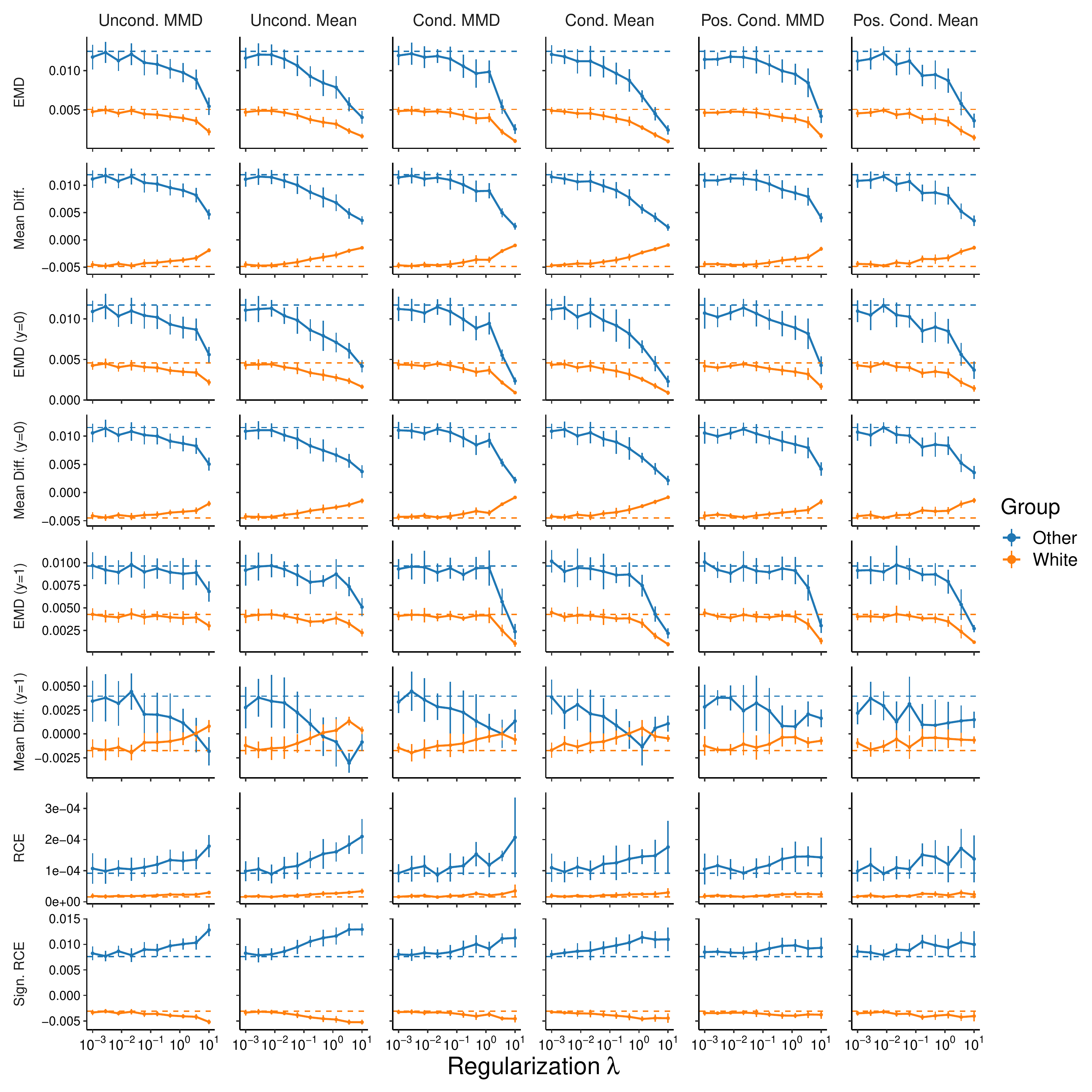}
	\caption{
	    Fairness metrics as a function of the extent $\lambda$ that violation of the fairness criterion is penalized when the \textbf{race and ethnicity} category is considered as the sensitive attribute for prediction of \textbf{ICU length of stay greater than 3 days} in the \textbf{MIMIC-III} database. Results shown are the mean $\pm$ SD for decomposed group-level metrics that assess conditional prediction parity (EMD and Mean Diff.) and relative calibration error (RCE and Sign. RCE) for objectives that penalize violation of threshold-free Demographic Parity (Uncond. MMD and Mean), Equalized Odds (Cond. MMD and Mean), and Equal Opportunity (Pos. Cond. MMD and Mean) on the basis of MMD- and mean-based penalties.  Measures of conditional prediction parity are separately assessed in the whole population and in the strata for which the outcome is and is not observed (suffixed with (y=1) and (y=0), respectively). 
	    Dashed lines correspond to the mean result for the unpenalized training procedure.
	}
	\label{fig:supplement/mimic/all_fairness/los_icu_3days/race_eth}
\end{figure}

\begin{figure}[!htb]
	\centering
	\includegraphics[width=0.9\linewidth]{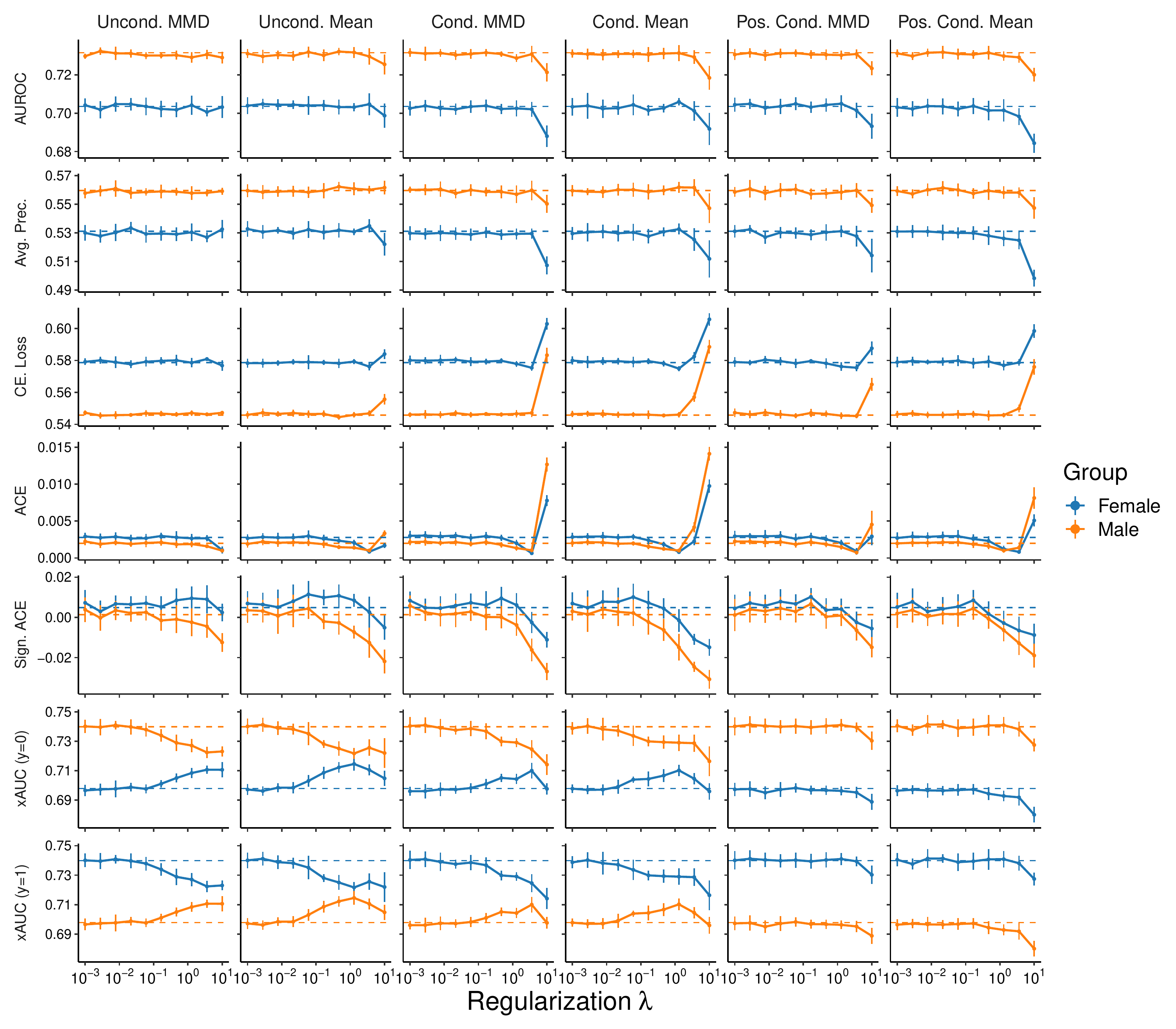}
	\caption{
	    Group-level model performance measures as a function of the extent $\lambda$ that violation of the fairness criterion is penalized when \textbf{sex} is considered as the sensitive attribute for prediction of \textbf{ICU length of stay greater than 3 days} in the \textbf{MIMIC-III} database. Results shown are the mean $\pm$ SD for the area under the ROC curve (AUROC), average precision (Avg. Prec), the cross entropy loss (CE Loss), the absolute calibration error (ACE), the signed absolute calibration error (Sign. ACE), and cross group ranking performance (xAUC; $\textrm{xAUC}_k^1$ is indicated by (y=1) and $\textrm{xAUC}_k^0$ by (y=0)) for each group for objectives that penalize violation of threshold-free Demographic Parity (Uncond. MMD and Mean), Equalized Odds (Cond. MMD and Mean), and Equal Opportunity (Pos. Cond. MMD and Mean) with MMD- and mean-based penalties. Dashed lines correspond to the mean result for the unpenalized training procedure.
	}
	\label{fig:supplement/mimic/all_performance/los_icu_3days/gender_concept_name}
\end{figure}

\begin{figure}[!htb]
	\centering
	\includegraphics[width=0.9\linewidth]{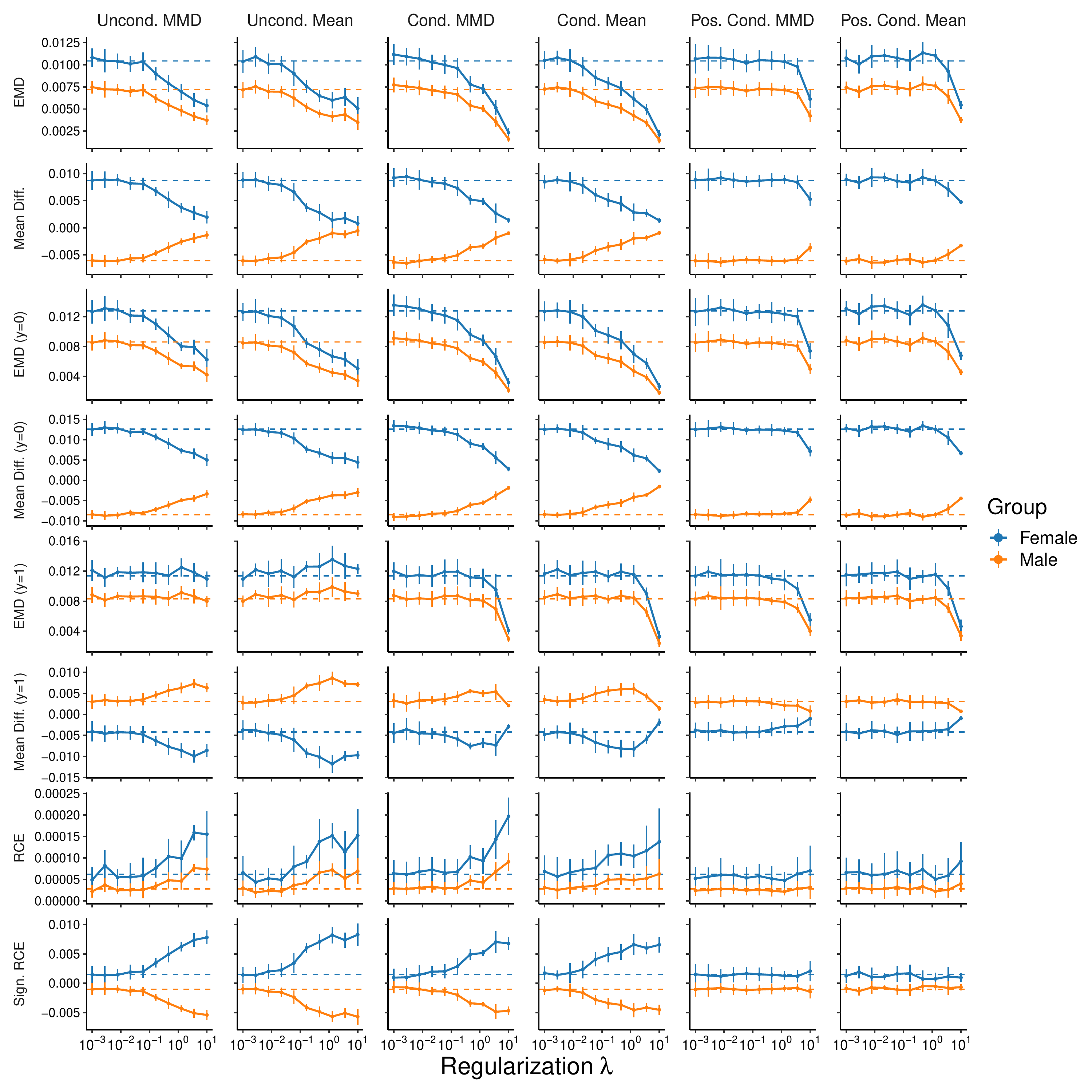}
	\caption{
	    Fairness metrics as a function of the extent $\lambda$ that violation of the fairness criterion is penalized when \textbf{sex} is considered as the sensitive attribute for prediction of \textbf{ICU length of stay greater than 3 days} in the \textbf{MIMIC-III} database. Results shown are the mean $\pm$ SD for decomposed group-level metrics that assess conditional prediction parity (EMD and Mean Diff.) and relative calibration error (RCE and Sign. RCE) for objectives that penalize violation of threshold-free Demographic Parity (Uncond. MMD and Mean), Equalized Odds (Cond. MMD and Mean), and Equal Opportunity (Pos. Cond. MMD and Mean) on the basis of MMD- and mean-based penalties.  Measures of conditional prediction parity are separately assessed in the whole population and in the strata for which the outcome is and is not observed (suffixed with (y=1) and (y=0), respectively). 
	    Dashed lines correspond to the mean result for the unpenalized training procedure.
	}
	\label{fig:supplement/mimic/all_fairness/los_icu_3days/gender_concept_name}
\end{figure}

\begin{figure}[!htb]
	\centering
	\includegraphics[width=0.9\linewidth]{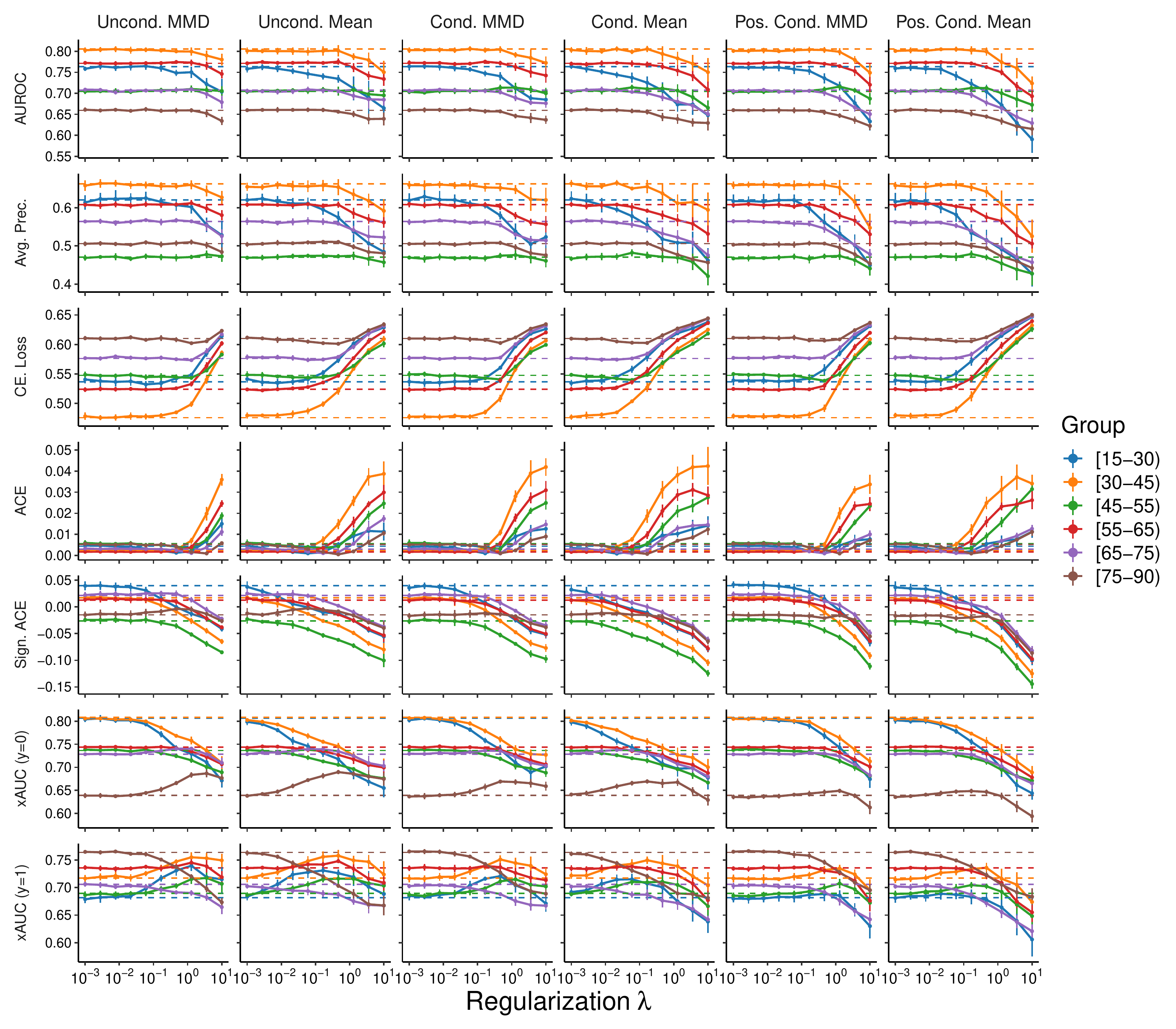}
	\caption{
	    Group-level model performance measures as a function of the extent $\lambda$ that violation of the fairness criterion is penalized when the \textbf{age} group is considered as the sensitive attribute for prediction of \textbf{ICU length of stay greater than 3 days} in the \textbf{MIMIC-III} database. Results shown are the mean $\pm$ SD for the area under the ROC curve (AUROC), average precision (Avg. Prec), the cross entropy loss (CE Loss), the absolute calibration error (ACE), the signed absolute calibration error (Sign. ACE), and cross group ranking performance (xAUC; $\textrm{xAUC}_k^1$ is indicated by (y=1) and $\textrm{xAUC}_k^0$ by (y=0)) for each group for objectives that penalize violation of threshold-free Demographic Parity (Uncond. MMD and Mean), Equalized Odds (Cond. MMD and Mean), and Equal Opportunity (Pos. Cond. MMD and Mean) with MMD- and mean-based penalties. Dashed lines correspond to the mean result for the unpenalized training procedure.
	}
	\label{fig:supplement/mimic/all_performance/los_icu_3days/age_group}
\end{figure}

\begin{figure}[!htb]
	\centering
	\includegraphics[width=0.9\linewidth]{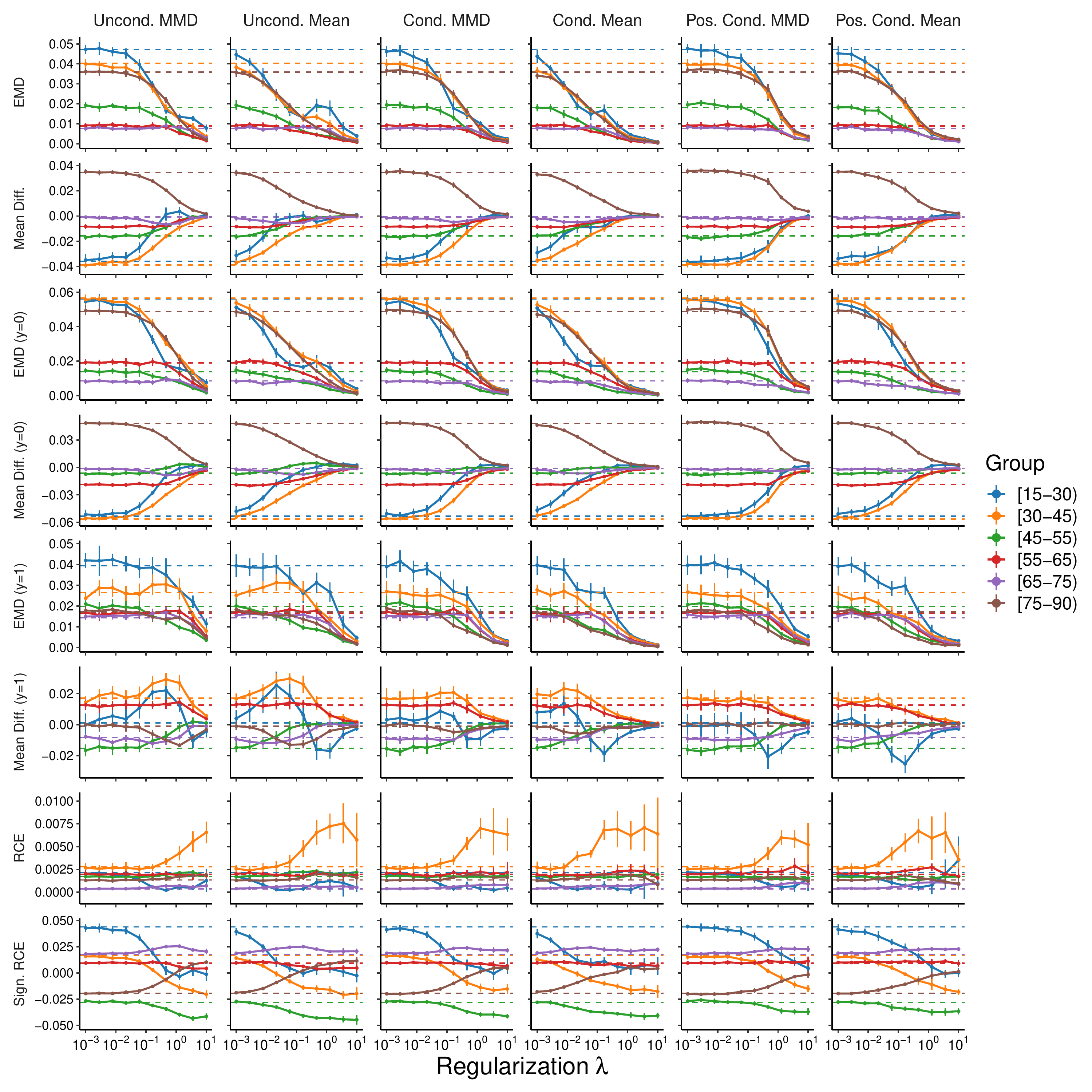}
	\caption{
	    Fairness metrics as a function of the extent $\lambda$ that violation of the fairness criterion is penalized when the \textbf{age} group is considered as the sensitive attribute for prediction of \textbf{ICU length of stay greater than 3 days} in the \textbf{MIMIC-III} database. Results shown are the mean $\pm$ SD for decomposed group-level metrics that assess conditional prediction parity (EMD and Mean Diff.) and relative calibration error (RCE and Sign. RCE) for objectives that penalize violation of threshold-free Demographic Parity (Uncond. MMD and Mean), Equalized Odds (Cond. MMD and Mean), and Equal Opportunity (Pos. Cond. MMD and Mean) on the basis of MMD- and mean-based penalties.  Measures of conditional prediction parity are separately assessed in the whole population and in the strata for which the outcome is and is not observed (suffixed with (y=1) and (y=0), respectively). 
	    Dashed lines correspond to the mean result for the unpenalized training procedure.
	}
	\label{fig:supplement/mimic/all_fairness/los_icu_3days/age_group}
\end{figure}

\begin{figure}[!htb]
	\centering
	\includegraphics[width=0.9\linewidth]{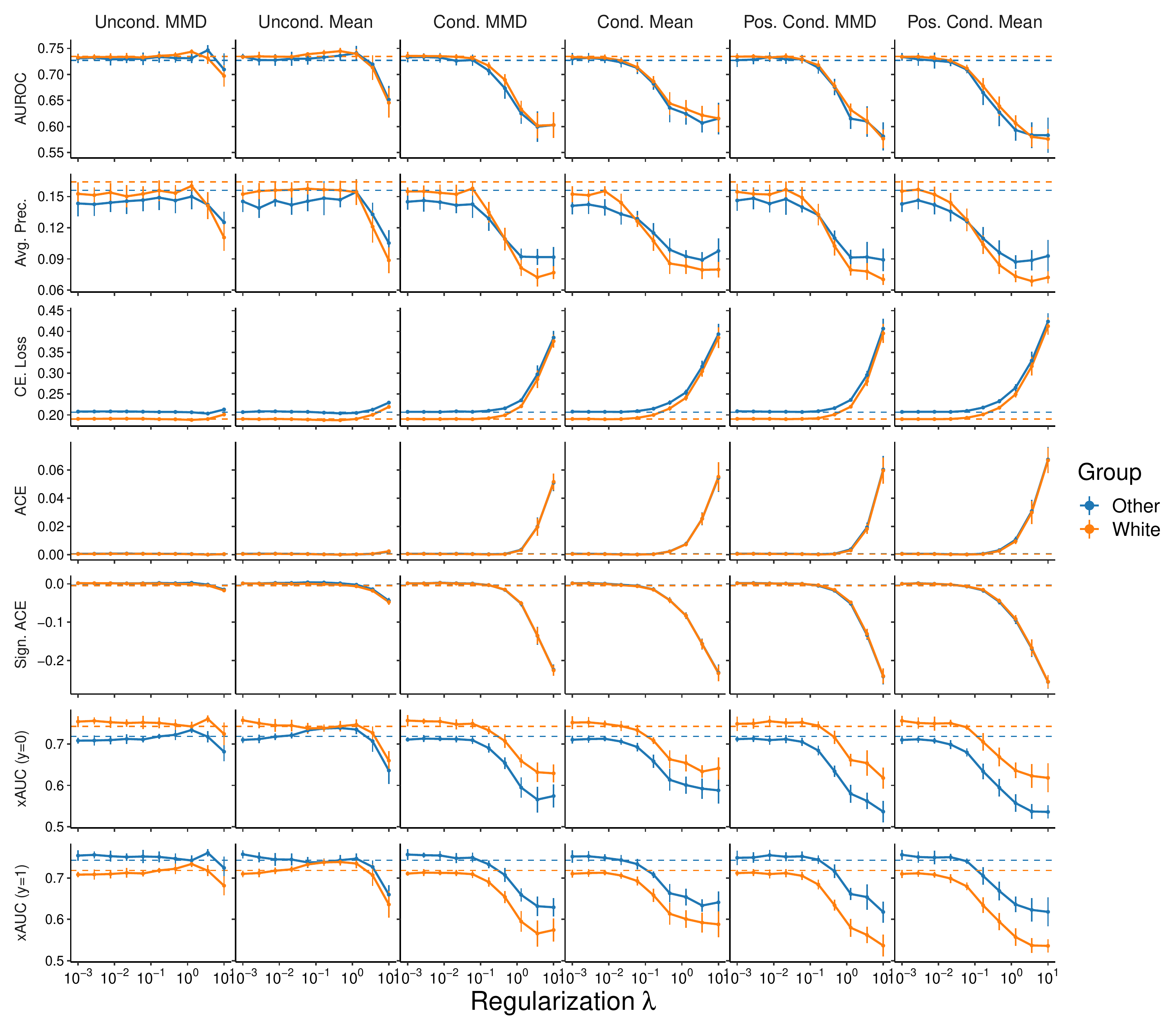}
	\caption{
	    Group-level model performance measures as a function of the extent $\lambda$ that violation of the fairness criterion is penalized when the \textbf{race and ethnicity} category is considered as the sensitive attribute for prediction of \textbf{ICU length of stay greater than 7 days} in the \textbf{MIMIC-III} database. Results shown are the mean $\pm$ SD for the area under the ROC curve (AUROC), average precision (Avg. Prec), the cross entropy loss (CE Loss), the absolute calibration error (ACE), the signed absolute calibration error (Sign. ACE), and cross group ranking performance (xAUC; $\textrm{xAUC}_k^1$ is indicated by (y=1) and $\textrm{xAUC}_k^0$ by (y=0)) for each group for objectives that penalize violation of threshold-free Demographic Parity (Uncond. MMD and Mean), Equalized Odds (Cond. MMD and Mean), and Equal Opportunity (Pos. Cond. MMD and Mean) with MMD- and mean-based penalties. Dashed lines correspond to the mean result for the unpenalized training procedure.
	}
	\label{fig:supplement/mimic/all_performance/los_icu_7days/race_eth}
\end{figure}

\begin{figure}[!htb]
	\centering
	\includegraphics[width=0.9\linewidth]{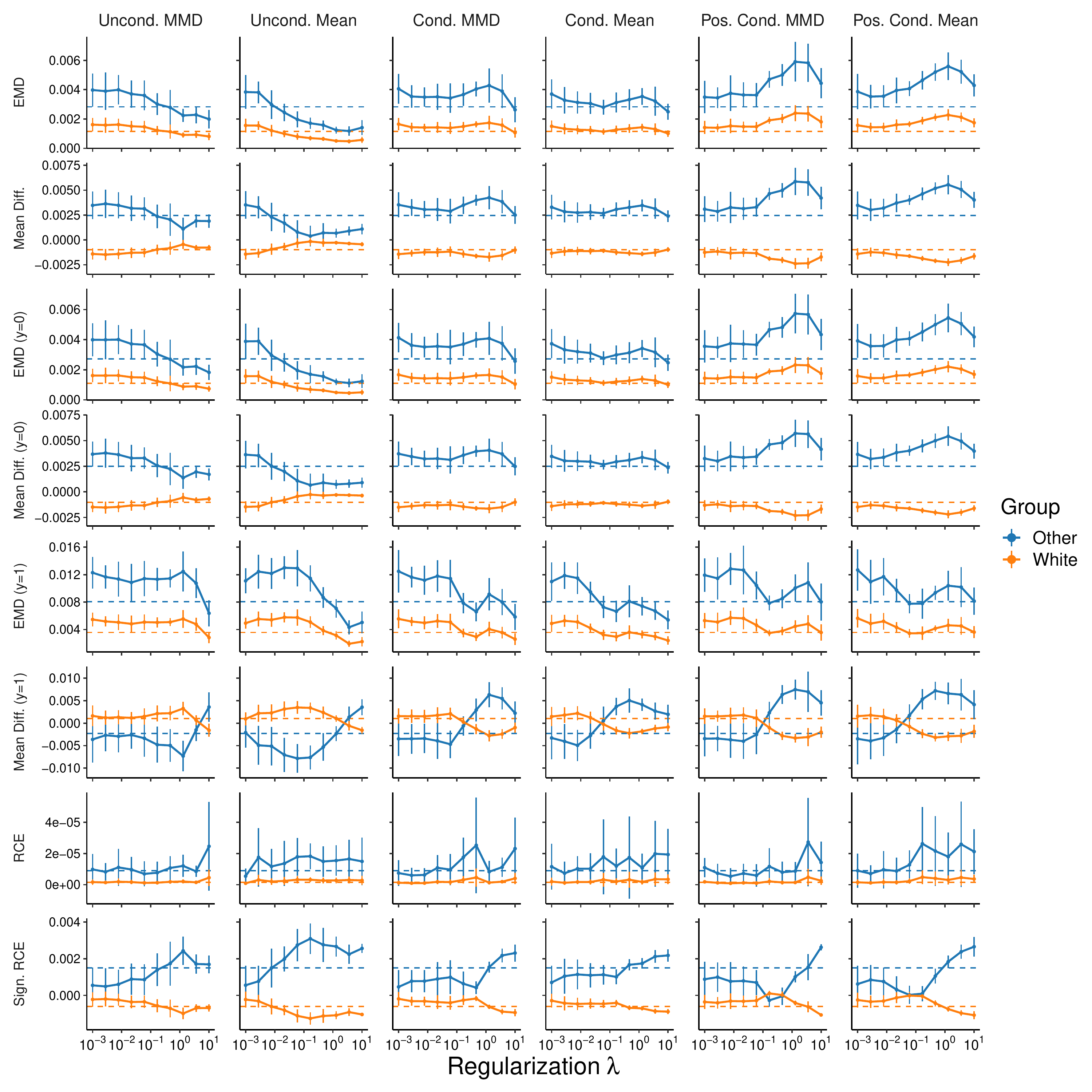}
	\caption{
	    Fairness metrics as a function of the extent $\lambda$ that violation of the fairness criterion is penalized when the \textbf{race and ethnicity} category is considered as the sensitive attribute for prediction of \textbf{ICU length of stay greater than 7 days} in the \textbf{MIMIC-III} database. Results shown are the mean $\pm$ SD for decomposed group-level metrics that assess conditional prediction parity (EMD and Mean Diff.) and relative calibration error (RCE and Sign. RCE) for objectives that penalize violation of threshold-free Demographic Parity (Uncond. MMD and Mean), Equalized Odds (Cond. MMD and Mean), and Equal Opportunity (Pos. Cond. MMD and Mean) on the basis of MMD- and mean-based penalties.  Measures of conditional prediction parity are separately assessed in the whole population and in the strata for which the outcome is and is not observed (suffixed with (y=1) and (y=0), respectively). 
	    Dashed lines correspond to the mean result for the unpenalized training procedure.
	}
	\label{fig:supplement/mimic/all_fairness/los_icu_7days/race_eth}
\end{figure}

\begin{figure}[!htb]
	\centering
	\includegraphics[width=0.9\linewidth]{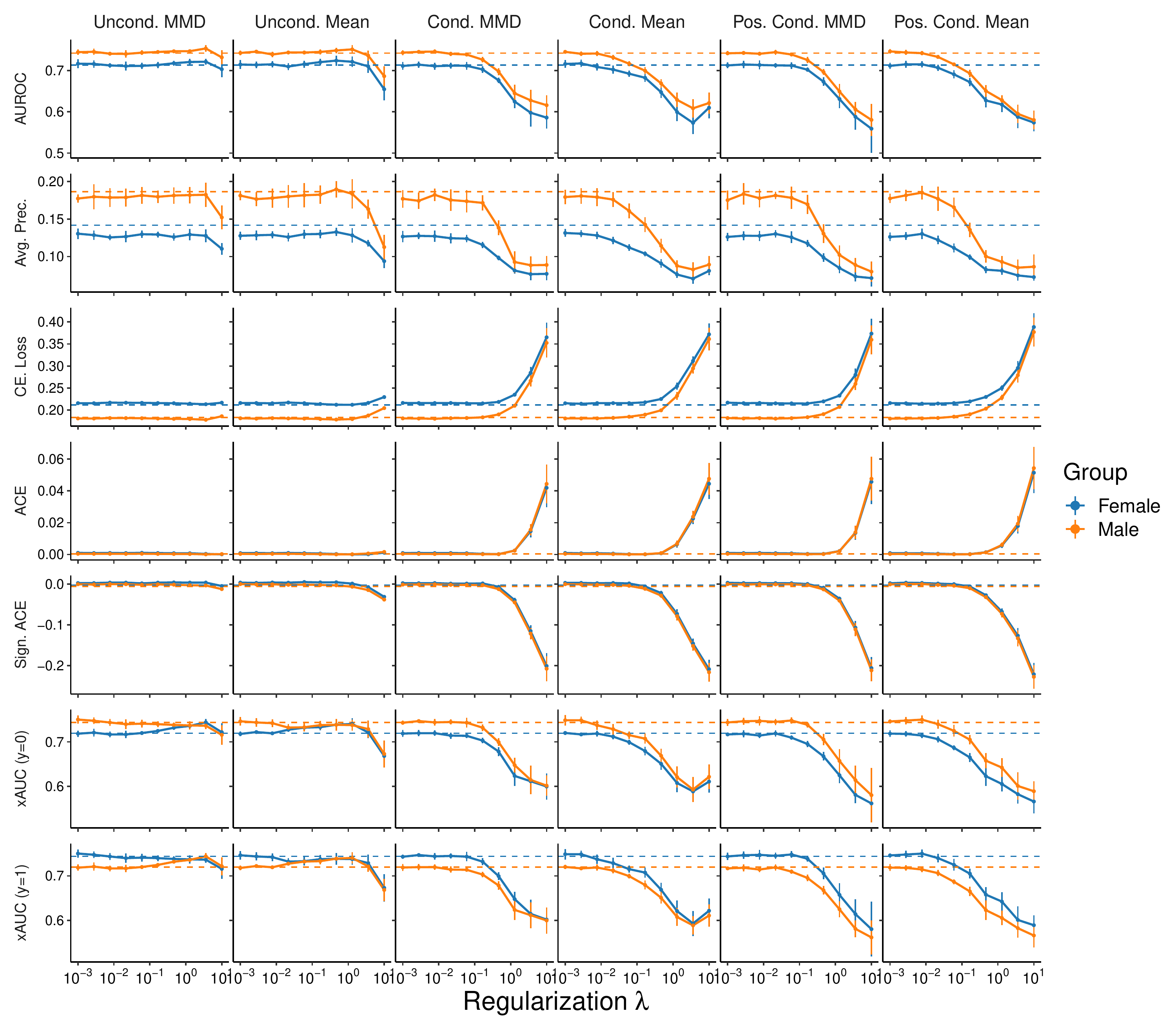}
	\caption{
	    Group-level model performance measures as a function of the extent $\lambda$ that violation of the fairness criterion is penalized when \textbf{sex} is considered as the sensitive attribute for prediction of \textbf{ICU length of stay greater than 7 days} in the \textbf{MIMIC-III} database. Results shown are the mean $\pm$ SD for the area under the ROC curve (AUROC), average precision (Avg. Prec), the cross entropy loss (CE Loss), the absolute calibration error (ACE), the signed absolute calibration error (Sign. ACE), and cross group ranking performance (xAUC; $\textrm{xAUC}_k^1$ is indicated by (y=1) and $\textrm{xAUC}_k^0$ by (y=0)) for each group for objectives that penalize violation of threshold-free Demographic Parity (Uncond. MMD and Mean), Equalized Odds (Cond. MMD and Mean), and Equal Opportunity (Pos. Cond. MMD and Mean) with MMD- and mean-based penalties. Dashed lines correspond to the mean result for the unpenalized training procedure.
	}
	\label{fig:supplement/mimic/all_performance/los_icu_7days/gender_concept_name}
\end{figure}

\begin{figure}[!htb]
	\centering
	\includegraphics[width=0.9\linewidth]{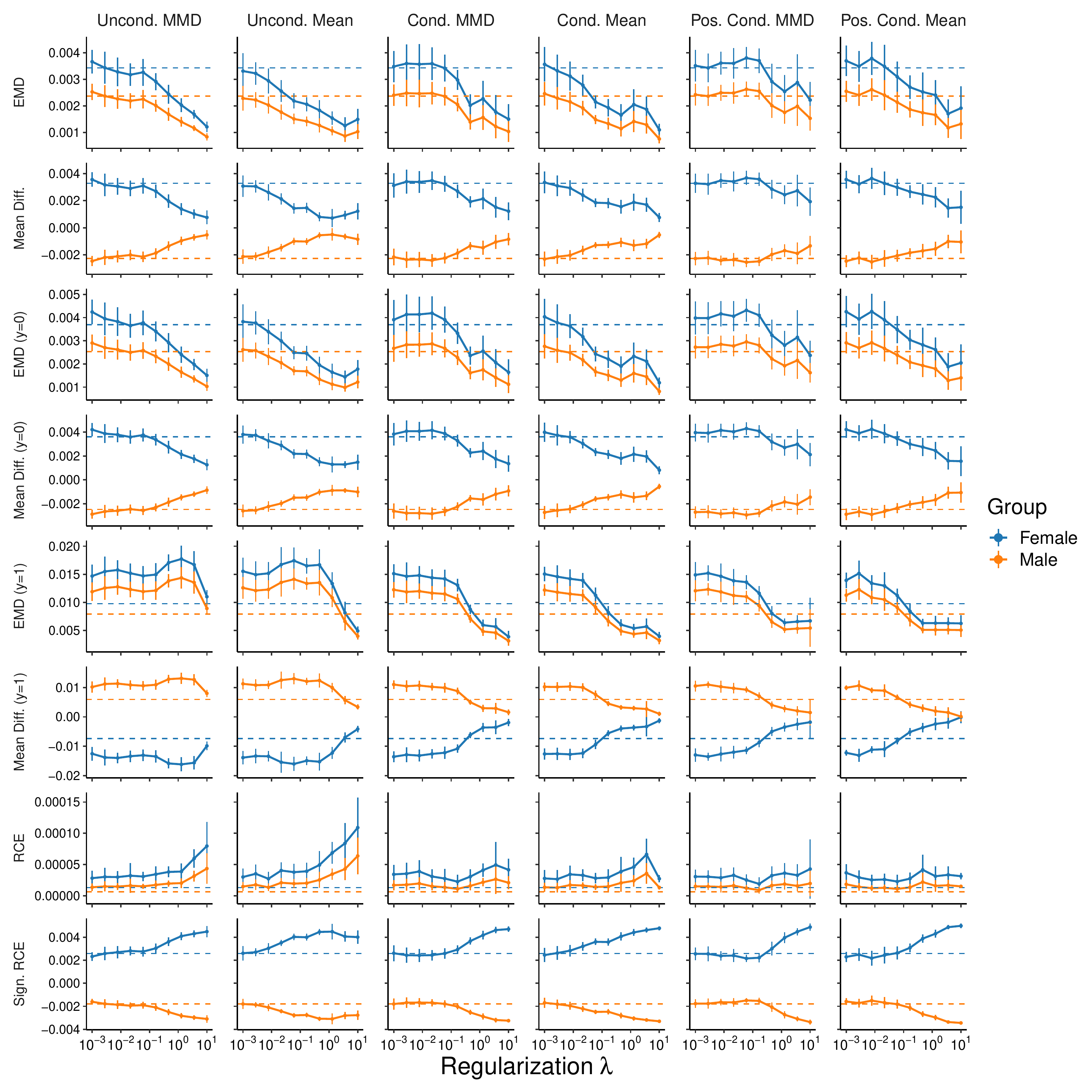}
	\caption{
	    Fairness metrics as a function of the extent $\lambda$ that violation of the fairness criterion is penalized when \textbf{sex} is considered as the sensitive attribute for prediction of \textbf{ICU length of stay greater than 7 days} in the \textbf{MIMIC-III} database. Results shown are the mean $\pm$ SD for decomposed group-level metrics that assess conditional prediction parity (EMD and Mean Diff.) and relative calibration error (RCE and Sign. RCE) for objectives that penalize violation of threshold-free Demographic Parity (Uncond. MMD and Mean), Equalized Odds (Cond. MMD and Mean), and Equal Opportunity (Pos. Cond. MMD and Mean) on the basis of MMD- and mean-based penalties.  Measures of conditional prediction parity are separately assessed in the whole population and in the strata for which the outcome is and is not observed (suffixed with (y=1) and (y=0), respectively). 
	    Dashed lines correspond to the mean result for the unpenalized training procedure.
	}
	\label{fig:supplement/mimic/all_fairness/los_icu_7days/gender_concept_name}
\end{figure}

\begin{figure}[!htb]
	\centering
	\includegraphics[width=0.9\linewidth]{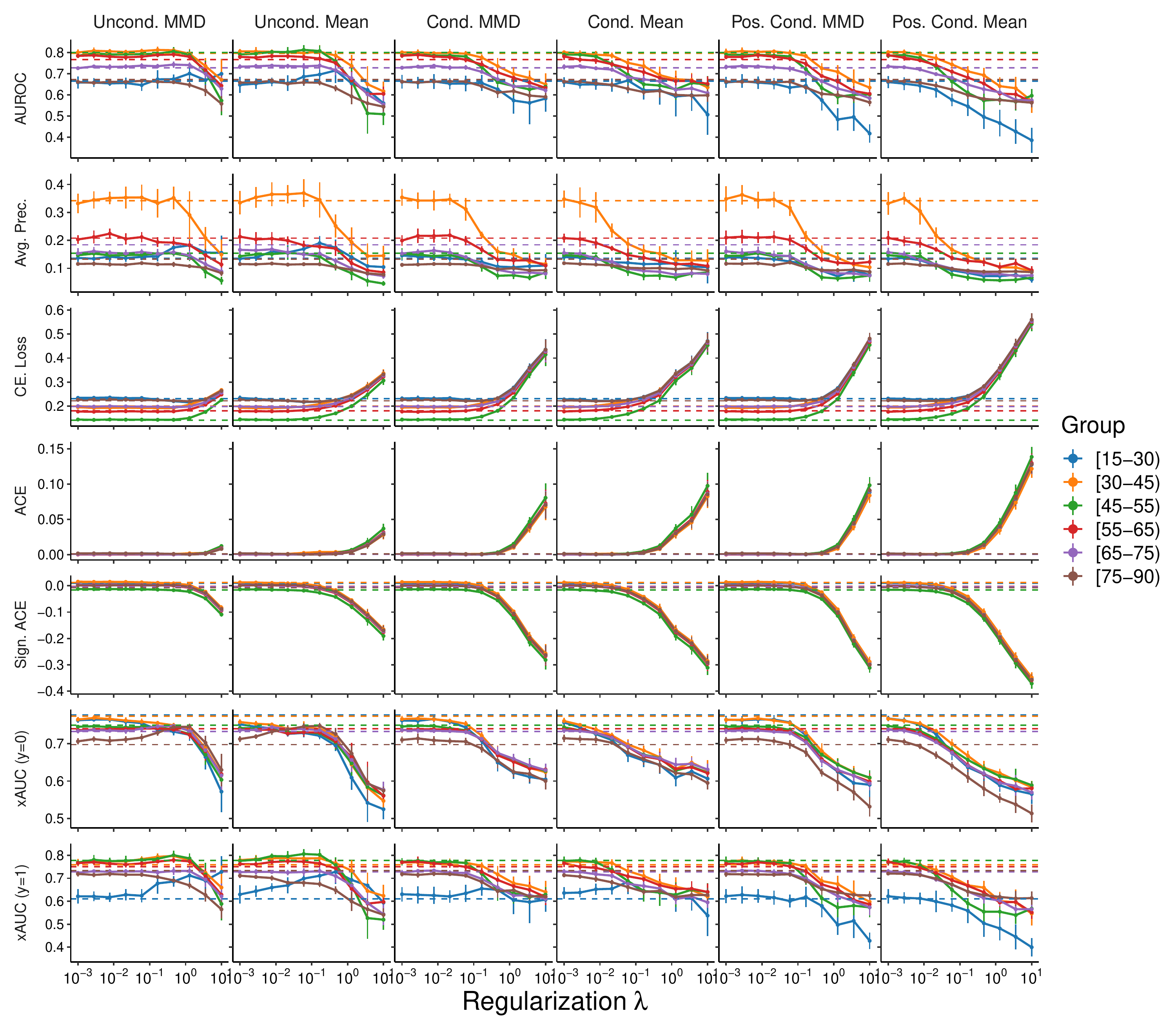}
	\caption{
	    Group-level model performance measures as a function of the extent $\lambda$ that violation of the fairness criterion is penalized when the \textbf{age} group is considered as the sensitive attribute for prediction of \textbf{ICU length of stay greater than 7 days} in the \textbf{MIMIC-III} database. Results shown are the mean $\pm$ SD for the area under the ROC curve (AUROC), average precision (Avg. Prec), the cross entropy loss (CE Loss), the absolute calibration error (ACE), the signed absolute calibration error (Sign. ACE), and cross group ranking performance (xAUC; $\textrm{xAUC}_k^1$ is indicated by (y=1) and $\textrm{xAUC}_k^0$ by (y=0)) for each group for objectives that penalize violation of threshold-free Demographic Parity (Uncond. MMD and Mean), Equalized Odds (Cond. MMD and Mean), and Equal Opportunity (Pos. Cond. MMD and Mean) with MMD- and mean-based penalties. Dashed lines correspond to the mean result for the unpenalized training procedure.
	}
	\label{fig:supplement/mimic/all_performance/los_icu_7days/age_group}
\end{figure}

\begin{figure}[!htb]
	\centering
	\includegraphics[width=0.9\linewidth]{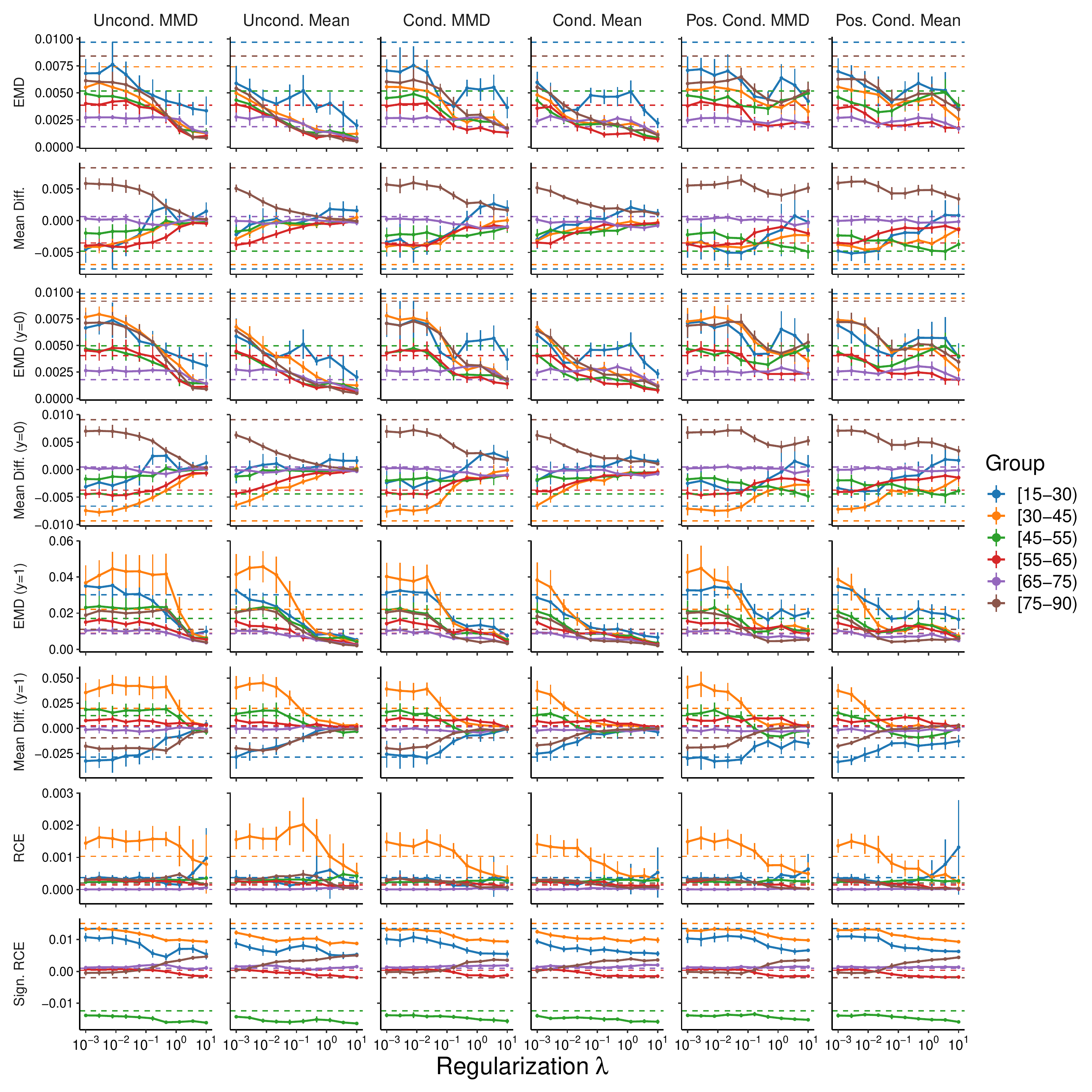}
	\caption{
	    Fairness metrics as a function of the extent $\lambda$ that violation of the fairness criterion is penalized when the \textbf{age} group is considered as the sensitive attribute for prediction of \textbf{ICU length of stay greater than 7 days} in the \textbf{MIMIC-III} database. Results shown are the mean $\pm$ SD for decomposed group-level metrics that assess conditional prediction parity (EMD and Mean Diff.) and relative calibration error (RCE and Sign. RCE) for objectives that penalize violation of threshold-free Demographic Parity (Uncond. MMD and Mean), Equalized Odds (Cond. MMD and Mean), and Equal Opportunity (Pos. Cond. MMD and Mean) on the basis of MMD- and mean-based penalties.  Measures of conditional prediction parity are separately assessed in the whole population and in the strata for which the outcome is and is not observed (suffixed with (y=1) and (y=0), respectively). 
	    Dashed lines correspond to the mean result for the unpenalized training procedure.
	}
	\label{fig:supplement/mimic/all_fairness/los_icu_7days/age_group}
\end{figure}

\begin{figure}[!htb]
	\centering
	\includegraphics[width=0.9\linewidth]{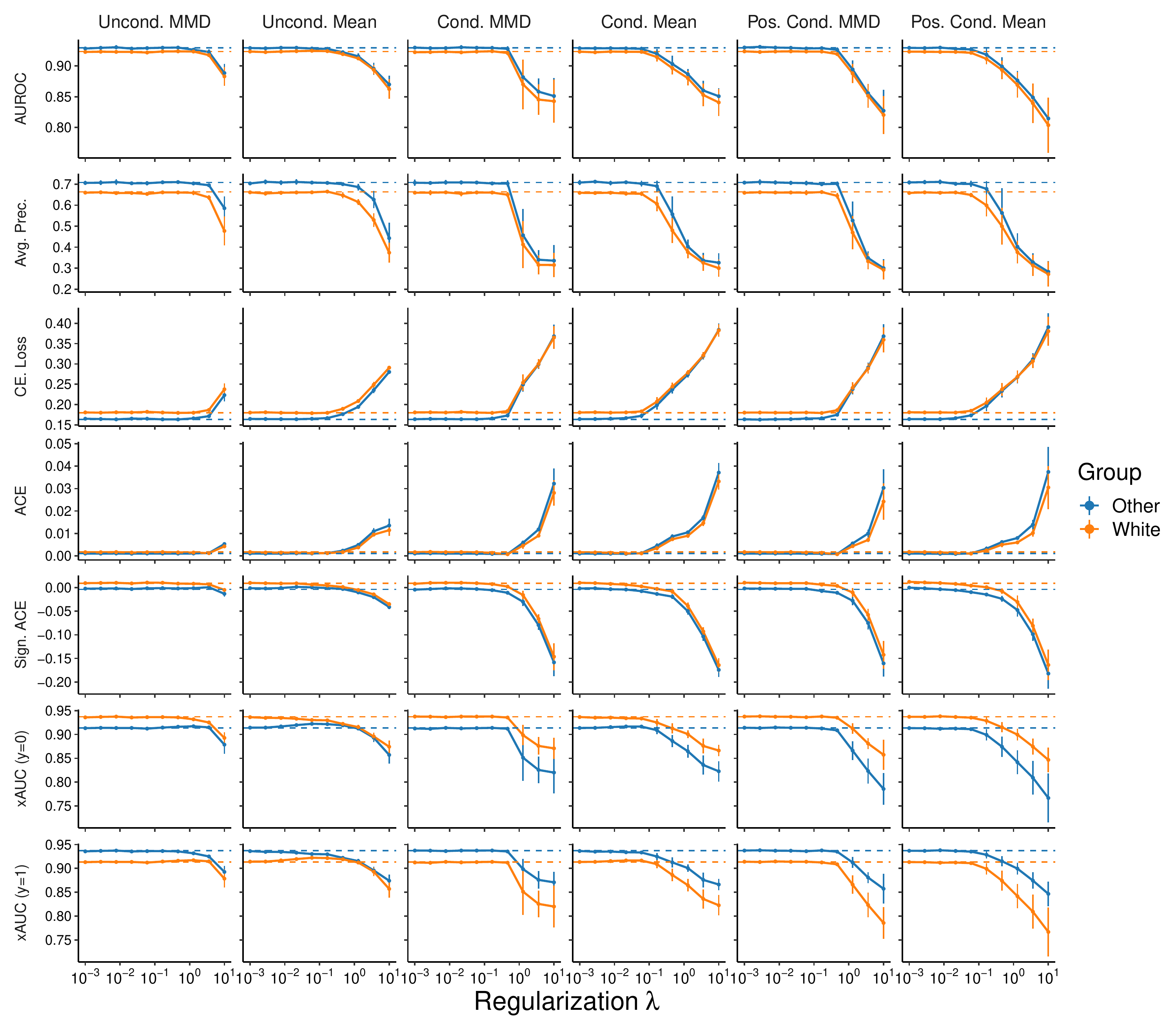}
	\caption{
	    Group-level model performance measures as a function of the extent $\lambda$ that violation of the fairness criterion is penalized when the \textbf{race and ethnicity} category is considered as the sensitive attribute for prediction of \textbf{hospital mortality} in the \textbf{MIMIC-III} database. Results shown are the mean $\pm$ SD for the area under the ROC curve (AUROC), average precision (Avg. Prec), the cross entropy loss (CE Loss), the absolute calibration error (ACE), the signed absolute calibration error (Sign. ACE), and cross group ranking performance (xAUC; $\textrm{xAUC}_k^1$ is indicated by (y=1) and $\textrm{xAUC}_k^0$ by (y=0)) for each group for objectives that penalize violation of threshold-free Demographic Parity (Uncond. MMD and Mean), Equalized Odds (Cond. MMD and Mean), and Equal Opportunity (Pos. Cond. MMD and Mean) with MMD- and mean-based penalties. Dashed lines correspond to the mean result for the unpenalized training procedure.
	}
	\label{fig:supplement/mimic/all_performance/mortality_hospital/race_eth}
\end{figure}

\begin{figure}[!htb]
	\centering
	\includegraphics[width=0.9\linewidth]{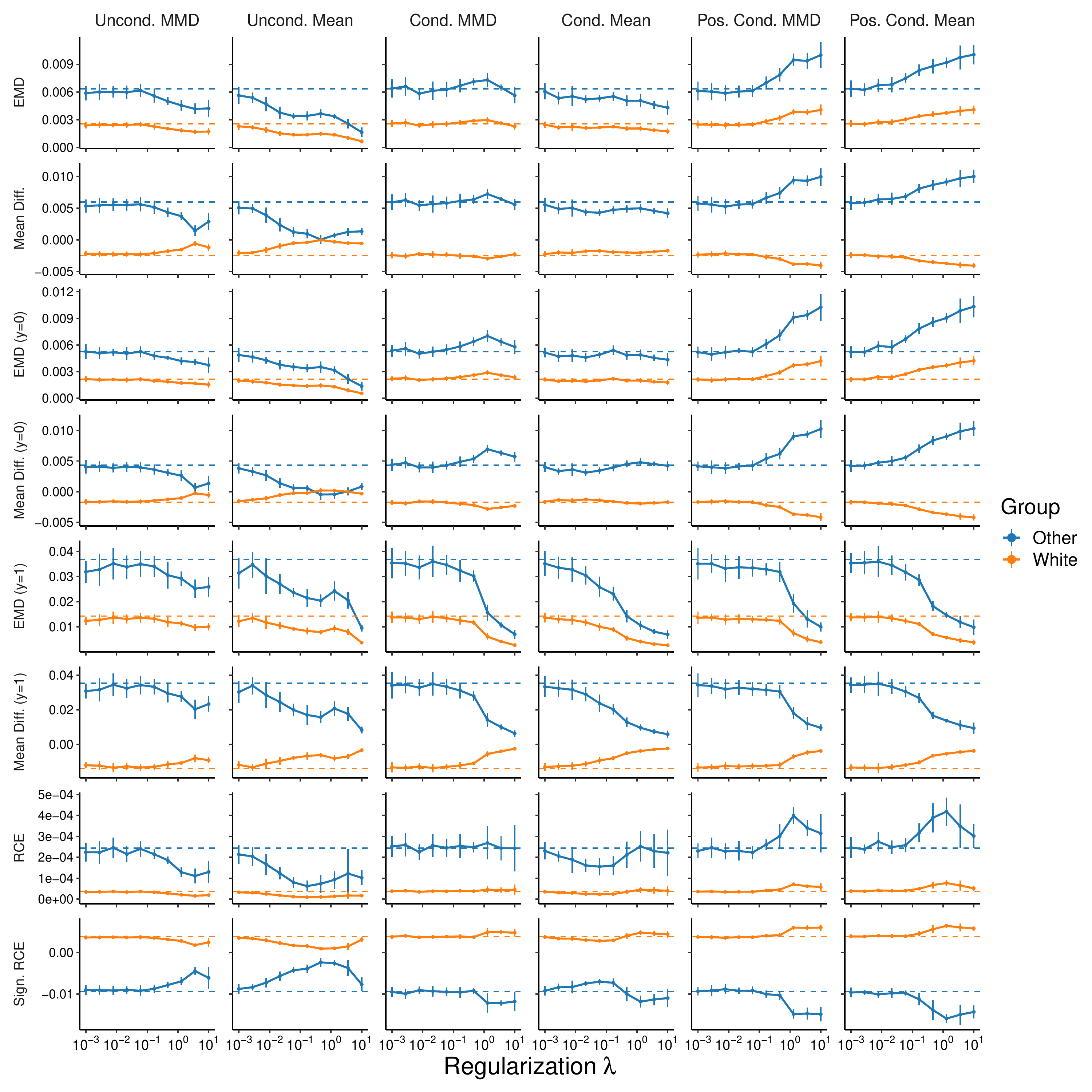}
	\caption{
	    Fairness metrics as a function of the extent $\lambda$ that violation of the fairness criterion is penalized when the \textbf{race and ethnicity} category is considered as the sensitive attribute for prediction of \textbf{hospital mortality} in the \textbf{MIMIC-III} database. Results shown are the mean $\pm$ SD for decomposed group-level metrics that assess conditional prediction parity (EMD and Mean Diff.) and relative calibration error (RCE and Sign. RCE) for objectives that penalize violation of threshold-free Demographic Parity (Uncond. MMD and Mean), Equalized Odds (Cond. MMD and Mean), and Equal Opportunity (Pos. Cond. MMD and Mean) on the basis of MMD- and mean-based penalties.  Measures of conditional prediction parity are separately assessed in the whole population and in the strata for which the outcome is and is not observed (suffixed with (y=1) and (y=0), respectively). 
	    Dashed lines correspond to the mean result for the unpenalized training procedure.
	}
	\label{fig:supplement/mimic/all_fairness/mortality_hospital/race_eth}
\end{figure}

\begin{figure}[!htb]
	\centering
	\includegraphics[width=0.9\linewidth]{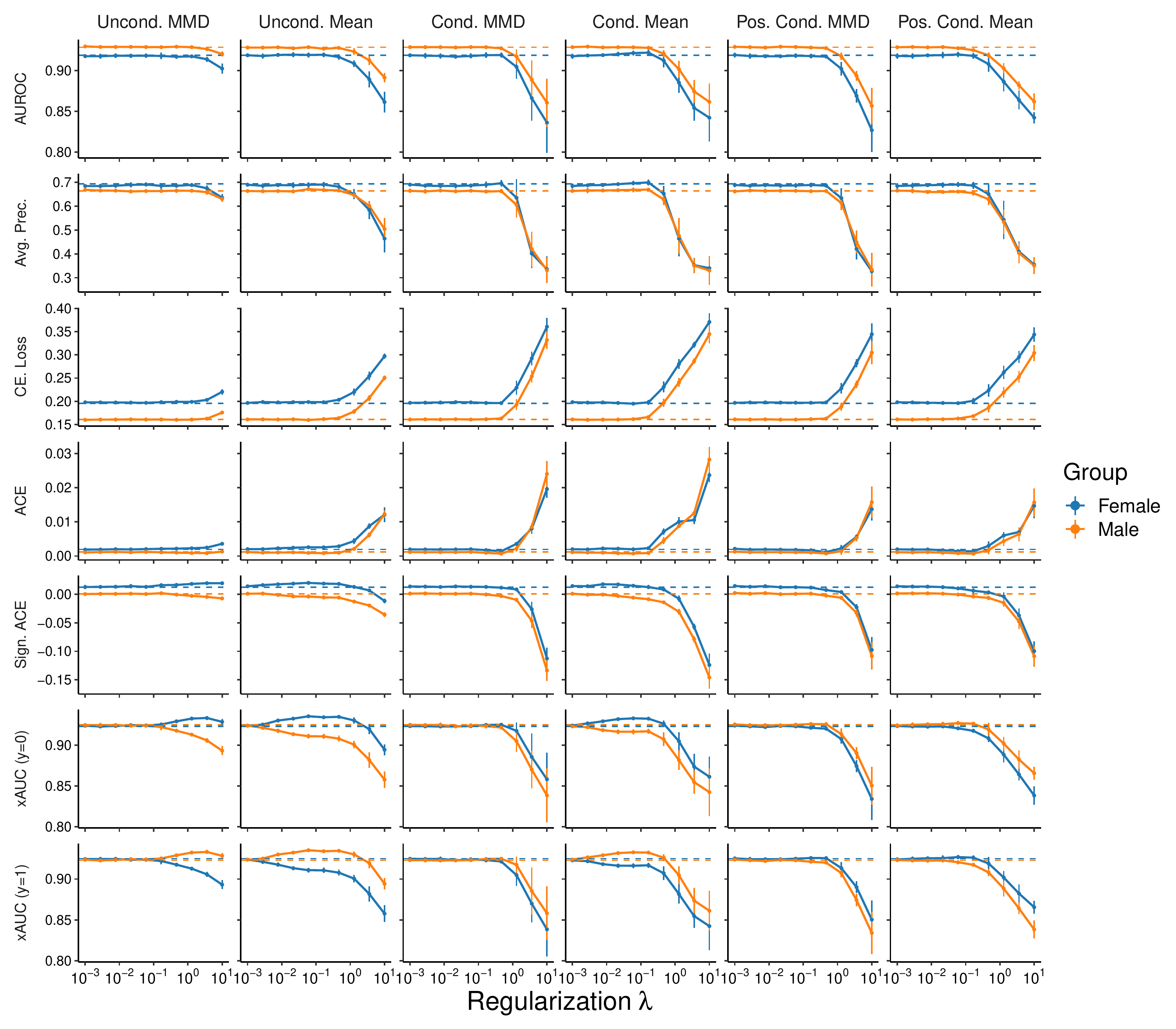}
	\caption{
	    Group-level model performance measures as a function of the extent $\lambda$ that violation of the fairness criterion is penalized when \textbf{sex} is considered as the sensitive attribute for prediction of \textbf{hospital mortality} in the \textbf{MIMIC-III} database. Results shown are the mean $\pm$ SD for the area under the ROC curve (AUROC), average precision (Avg. Prec), the cross entropy loss (CE Loss), the absolute calibration error (ACE), the signed absolute calibration error (Sign. ACE), and cross group ranking performance (xAUC; $\textrm{xAUC}_k^1$ is indicated by (y=1) and $\textrm{xAUC}_k^0$ by (y=0)) for each group for objectives that penalize violation of threshold-free Demographic Parity (Uncond. MMD and Mean), Equalized Odds (Cond. MMD and Mean), and Equal Opportunity (Pos. Cond. MMD and Mean) with MMD- and mean-based penalties. Dashed lines correspond to the mean result for the unpenalized training procedure.
	}
	\label{fig:supplement/mimic/all_performance/mortality_hospital/gender_concept_name}
\end{figure}

\begin{figure}[!htb]
	\centering
	\includegraphics[width=0.9\linewidth]{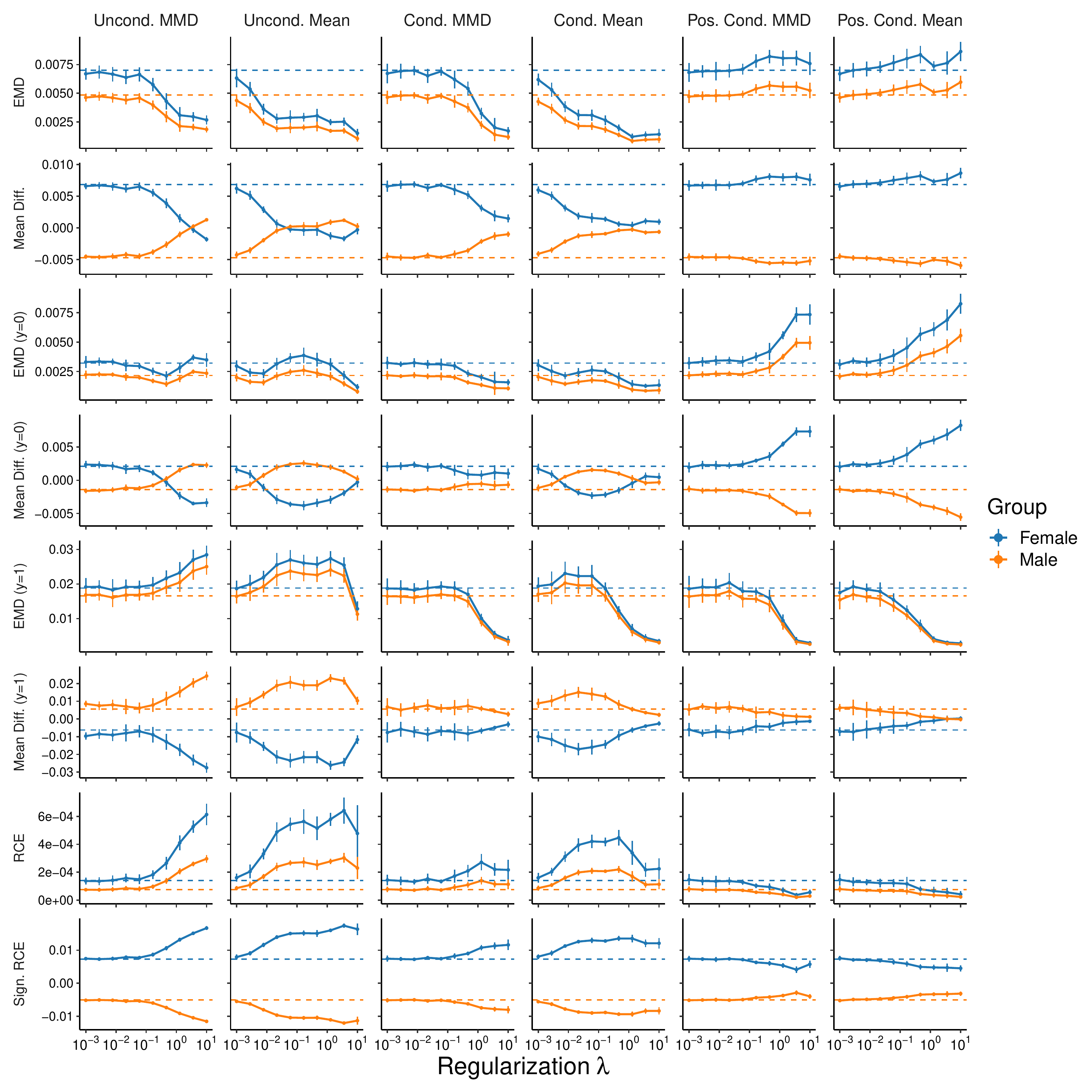}
	\caption{
	    Fairness metrics as a function of the extent $\lambda$ that violation of the fairness criterion is penalized when \textbf{sex} is considered as the sensitive attribute for prediction of \textbf{hospital mortality} in the \textbf{MIMIC-III} database. Results shown are the mean $\pm$ SD for decomposed group-level metrics that assess conditional prediction parity (EMD and Mean Diff.) and relative calibration error (RCE and Sign. RCE) for objectives that penalize violation of threshold-free Demographic Parity (Uncond. MMD and Mean), Equalized Odds (Cond. MMD and Mean), and Equal Opportunity (Pos. Cond. MMD and Mean) on the basis of MMD- and mean-based penalties.  Measures of conditional prediction parity are separately assessed in the whole population and in the strata for which the outcome is and is not observed (suffixed with (y=1) and (y=0), respectively). 
	    Dashed lines correspond to the mean result for the unpenalized training procedure.
	}
	\label{fig:supplement/mimic/all_fairness/mortality_hospital/gender_concept_name}
\end{figure}

\begin{figure}[!htb]
	\centering
	\includegraphics[width=0.9\linewidth]{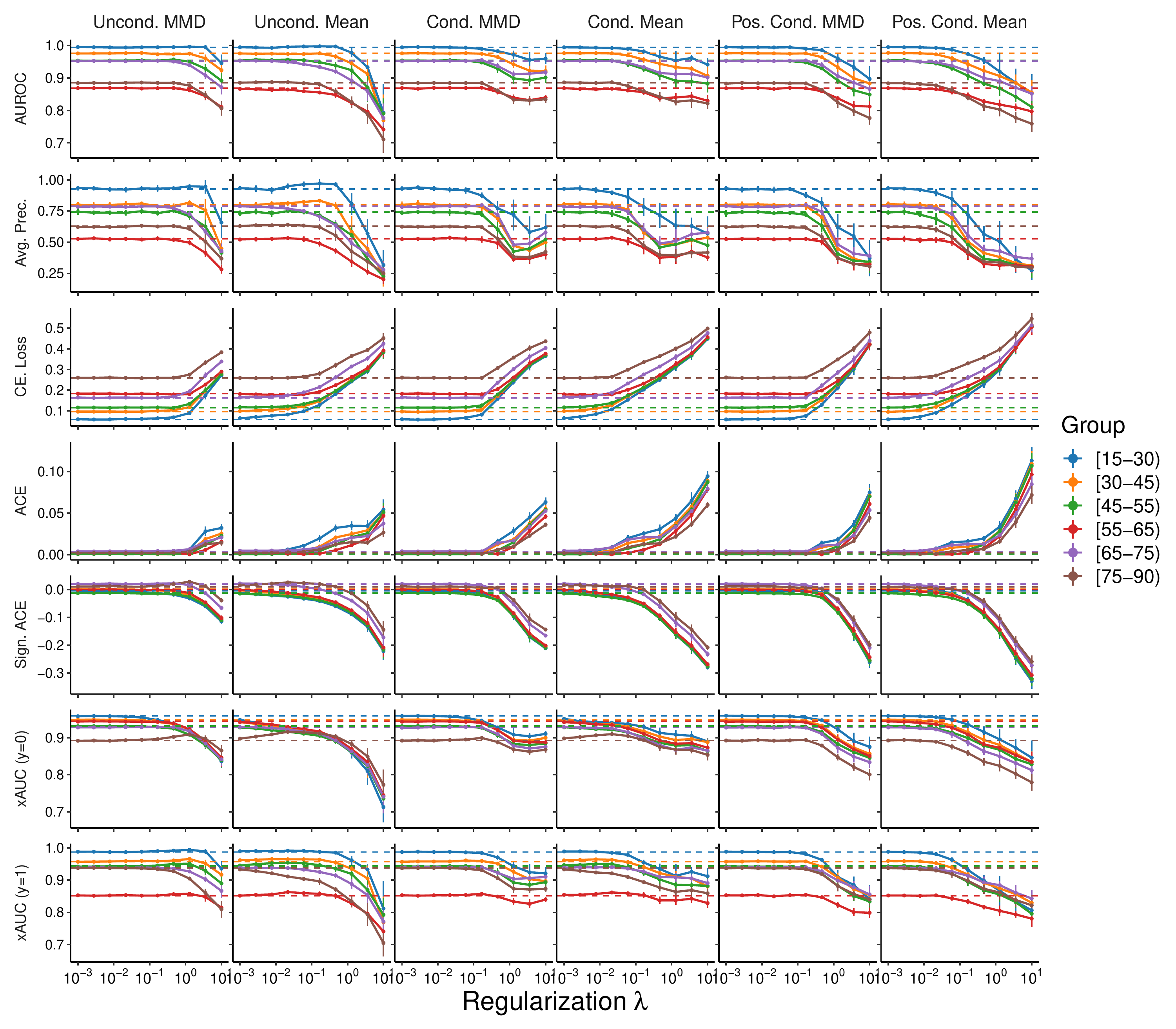}
	\caption{
	    Group-level model performance measures as a function of the extent $\lambda$ that violation of the fairness criterion is penalized when the \textbf{age} group is considered as the sensitive attribute for prediction of \textbf{hospital mortality} in the \textbf{MIMIC-III} database. Results shown are the mean $\pm$ SD for the area under the ROC curve (AUROC), average precision (Avg. Prec), the cross entropy loss (CE Loss), the absolute calibration error (ACE), the signed absolute calibration error (Sign. ACE), and cross group ranking performance (xAUC; $\textrm{xAUC}_k^1$ is indicated by (y=1) and $\textrm{xAUC}_k^0$ by (y=0)) for each group for objectives that penalize violation of threshold-free Demographic Parity (Uncond. MMD and Mean), Equalized Odds (Cond. MMD and Mean), and Equal Opportunity (Pos. Cond. MMD and Mean) with MMD- and mean-based penalties. Dashed lines correspond to the mean result for the unpenalized training procedure.
	}
	\label{fig:supplement/mimic/all_performance/mortality_hospital/age_group}
\end{figure}

\begin{figure}[!htb]
	\centering
	\includegraphics[width=0.9\linewidth]{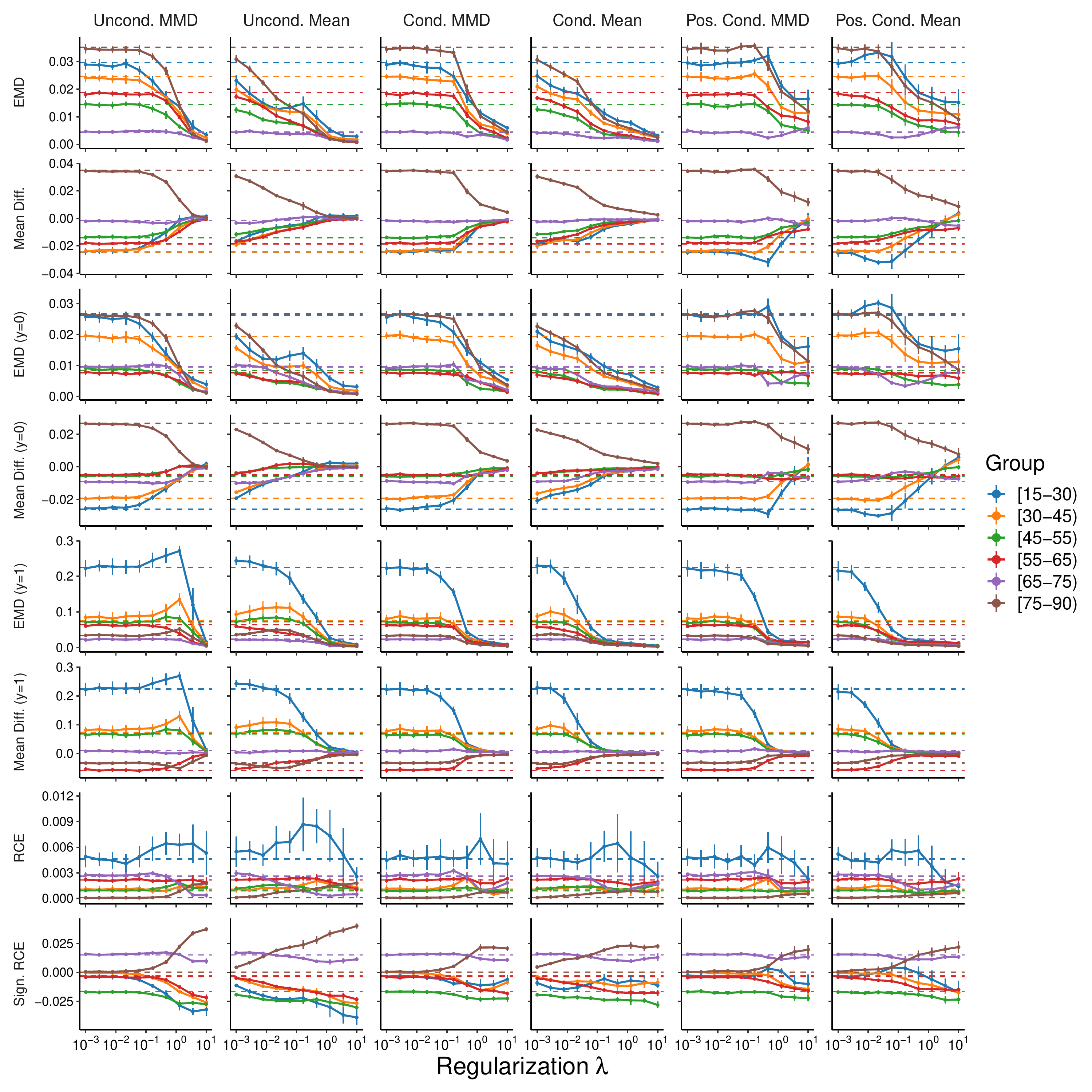}
	\caption{
	    Fairness metrics as a function of the extent $\lambda$ that violation of the fairness criterion is penalized when the \textbf{age} group is considered as the sensitive attribute for prediction of \textbf{hospital mortality} in the \textbf{MIMIC-III} database. Results shown are the mean $\pm$ SD for decomposed group-level metrics that assess conditional prediction parity (EMD and Mean Diff.) and relative calibration error (RCE and Sign. RCE) for objectives that penalize violation of threshold-free Demographic Parity (Uncond. MMD and Mean), Equalized Odds (Cond. MMD and Mean), and Equal Opportunity (Pos. Cond. MMD and Mean) on the basis of MMD- and mean-based penalties.  Measures of conditional prediction parity are separately assessed in the whole population and in the strata for which the outcome is and is not observed (suffixed with (y=1) and (y=0), respectively). 
	    Dashed lines correspond to the mean result for the unpenalized training procedure.
	}
	\label{fig:supplement/mimic/all_fairness/mortality_hospital/age_group}
\end{figure}

\begin{figure}[!htb]
	\centering
	\includegraphics[width=0.9\linewidth]{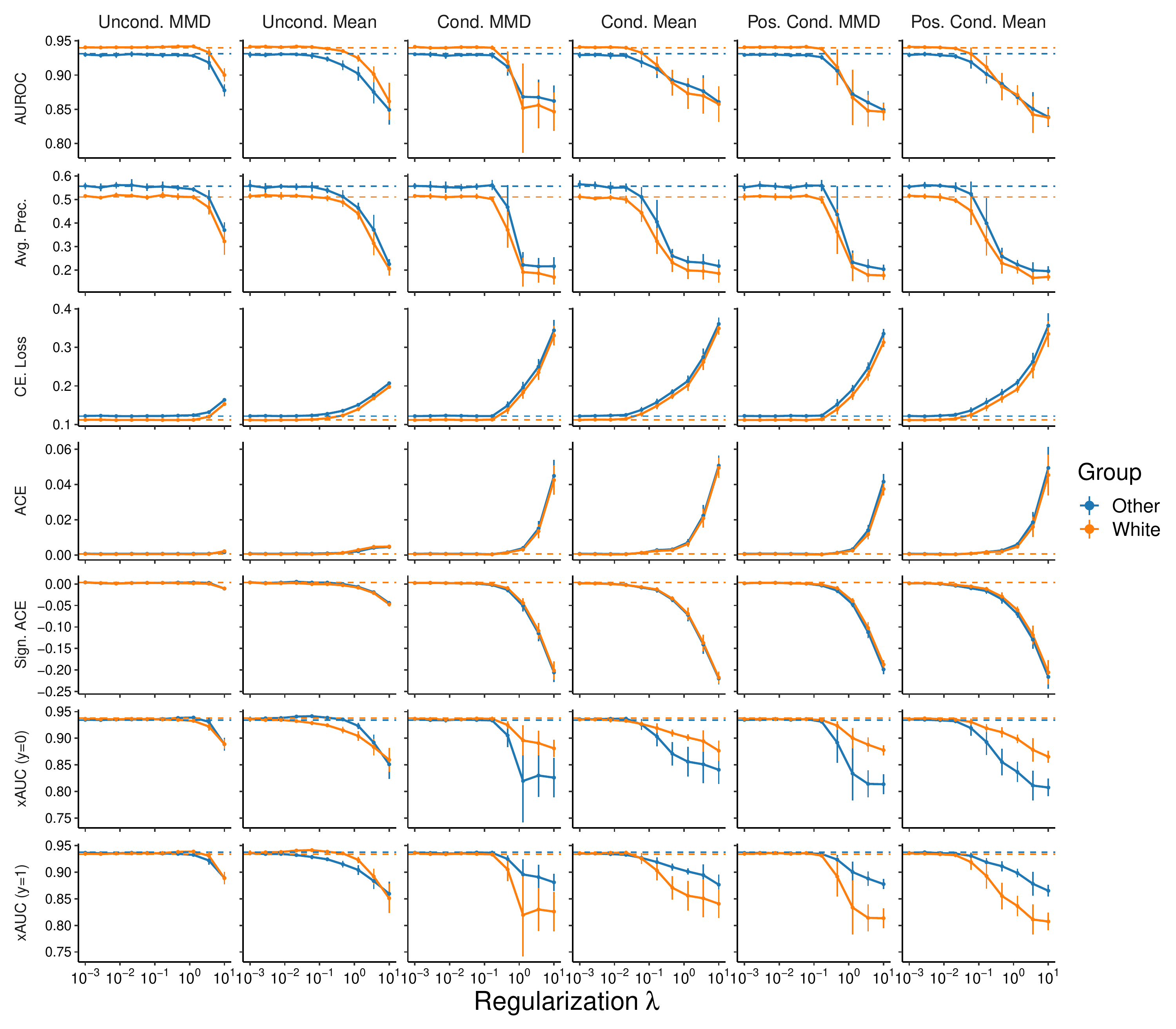}
	\caption{
	    Group-level model performance measures as a function of the extent $\lambda$ that violation of the fairness criterion is penalized when the \textbf{race and ethnicity} category is considered as the sensitive attribute for prediction of \textbf{ICU mortality} in the \textbf{MIMIC-III} database. Results shown are the mean $\pm$ SD for the area under the ROC curve (AUROC), average precision (Avg. Prec), the cross entropy loss (CE Loss), the absolute calibration error (ACE), the signed absolute calibration error (Sign. ACE), and cross group ranking performance (xAUC; $\textrm{xAUC}_k^1$ is indicated by (y=1) and $\textrm{xAUC}_k^0$ by (y=0)) for each group for objectives that penalize violation of threshold-free Demographic Parity (Uncond. MMD and Mean), Equalized Odds (Cond. MMD and Mean), and Equal Opportunity (Pos. Cond. MMD and Mean) with MMD- and mean-based penalties. Dashed lines correspond to the mean result for the unpenalized training procedure.
	}
	\label{fig:supplement/mimic/all_performance/mortality_icu/race_eth}
\end{figure}

\begin{figure}[!htb]
	\centering
	\includegraphics[width=0.9\linewidth]{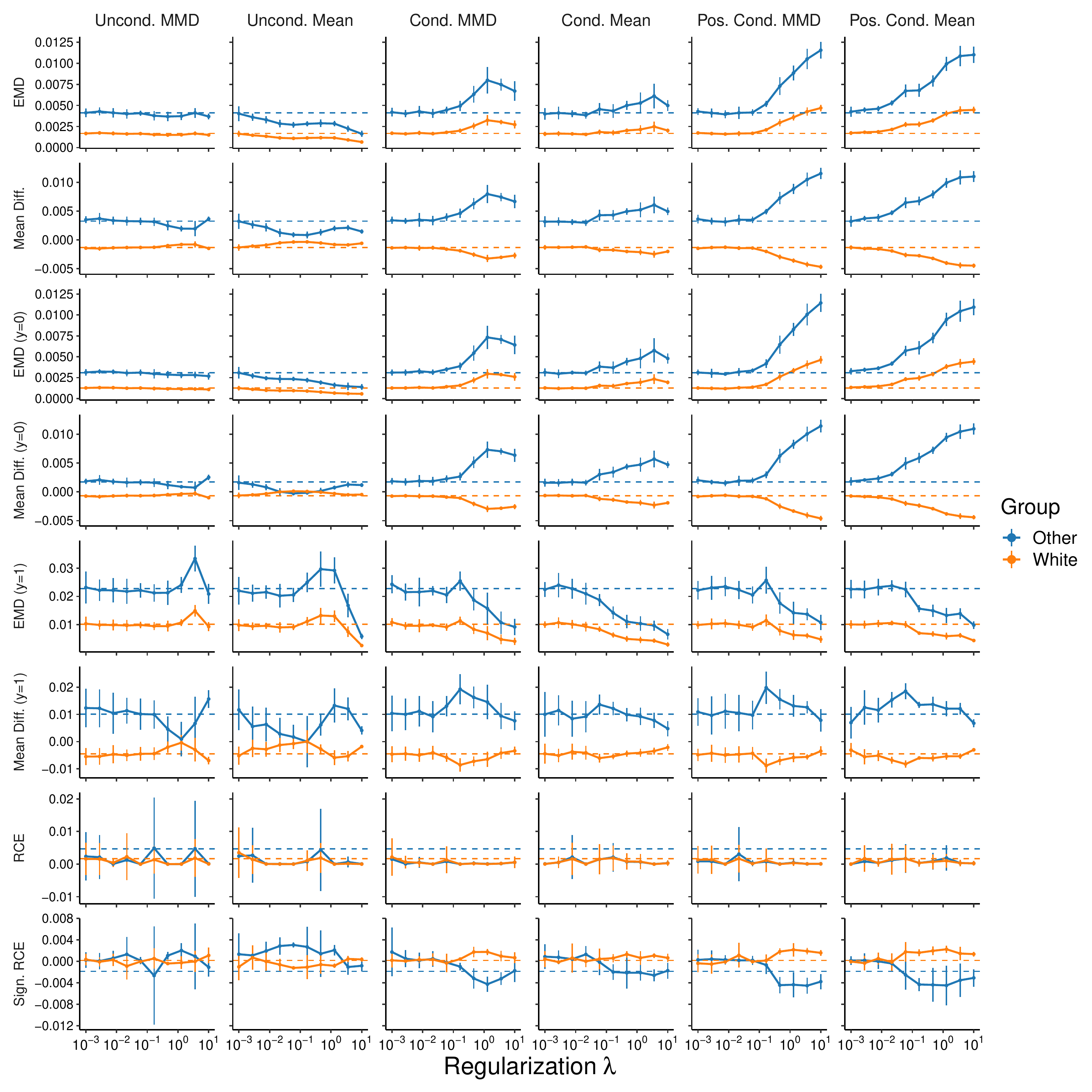}
	\caption{
	    Fairness metrics as a function of the extent $\lambda$ that violation of the fairness criterion is penalized when the \textbf{race and ethnicity} category is considered as the sensitive attribute for prediction of \textbf{ICU mortality} in the \textbf{MIMIC-III} database. Results shown are the mean $\pm$ SD for decomposed group-level metrics that assess conditional prediction parity (EMD and Mean Diff.) and relative calibration error (RCE and Sign. RCE) for objectives that penalize violation of threshold-free Demographic Parity (Uncond. MMD and Mean), Equalized Odds (Cond. MMD and Mean), and Equal Opportunity (Pos. Cond. MMD and Mean) on the basis of MMD- and mean-based penalties.  Measures of conditional prediction parity are separately assessed in the whole population and in the strata for which the outcome is and is not observed (suffixed with (y=1) and (y=0), respectively). 
	    Dashed lines correspond to the mean result for the unpenalized training procedure.
	}
	\label{fig:supplement/mimic/all_fairness/mortality_icu/race_eth}
\end{figure}

\begin{figure}[!htb]
	\centering
	\includegraphics[width=0.9\linewidth]{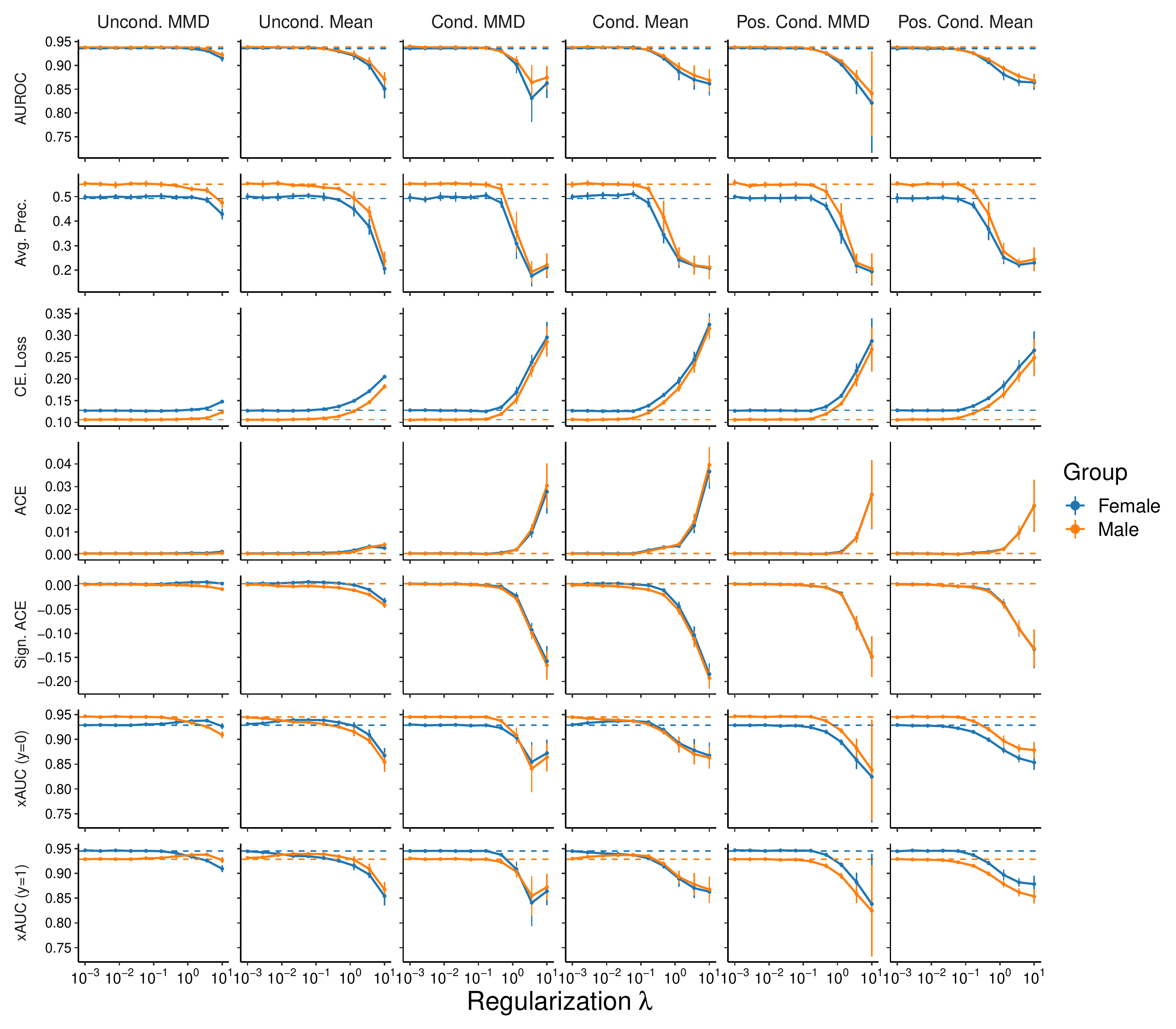}
	\caption{
	    Group-level model performance measures as a function of the extent $\lambda$ that violation of the fairness criterion is penalized when \textbf{sex} is considered as the sensitive attribute for prediction of \textbf{ICU mortality} in the \textbf{MIMIC-III} database. Results shown are the mean $\pm$ SD for the area under the ROC curve (AUROC), average precision (Avg. Prec), the cross entropy loss (CE Loss), the absolute calibration error (ACE), the signed absolute calibration error (Sign. ACE), and cross group ranking performance (xAUC; $\textrm{xAUC}_k^1$ is indicated by (y=1) and $\textrm{xAUC}_k^0$ by (y=0)) for each group for objectives that penalize violation of threshold-free Demographic Parity (Uncond. MMD and Mean), Equalized Odds (Cond. MMD and Mean), and Equal Opportunity (Pos. Cond. MMD and Mean) with MMD- and mean-based penalties. Dashed lines correspond to the mean result for the unpenalized training procedure.
	}
	\label{fig:supplement/mimic/all_performance/mortality_icu/gender_concept_name}
\end{figure}

\begin{figure}[!htb]
	\centering
	\includegraphics[width=0.9\linewidth]{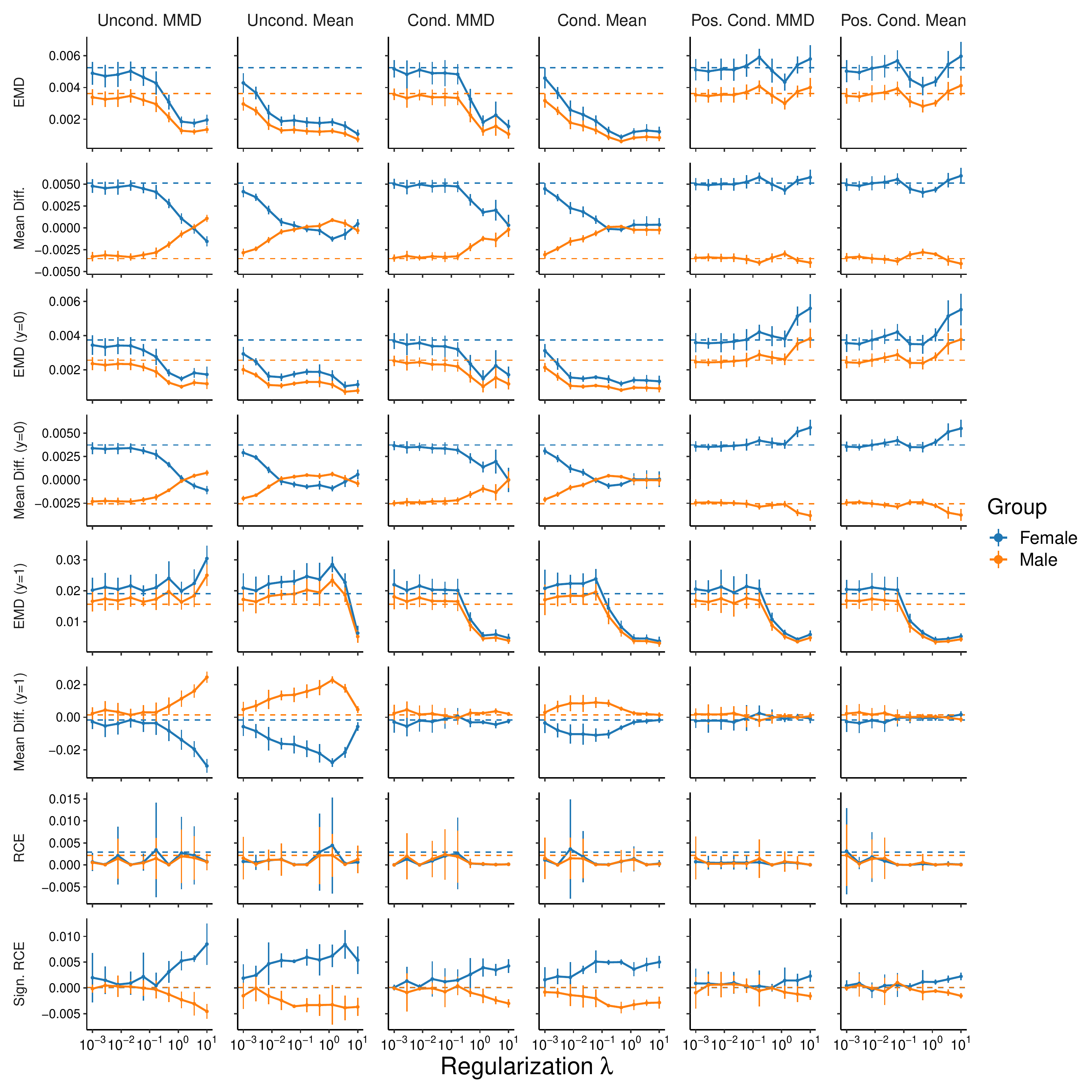}
	\caption{
	    Fairness metrics as a function of the extent $\lambda$ that violation of the fairness criterion is penalized when \textbf{sex} is considered as the sensitive attribute for prediction of \textbf{ICU mortality} in the \textbf{MIMIC-III} database. Results shown are the mean $\pm$ SD for decomposed group-level metrics that assess conditional prediction parity (EMD and Mean Diff.) and relative calibration error (RCE and Sign. RCE) for objectives that penalize violation of threshold-free Demographic Parity (Uncond. MMD and Mean), Equalized Odds (Cond. MMD and Mean), and Equal Opportunity (Pos. Cond. MMD and Mean) on the basis of MMD- and mean-based penalties.  Measures of conditional prediction parity are separately assessed in the whole population and in the strata for which the outcome is and is not observed (suffixed with (y=1) and (y=0), respectively). 
	    Dashed lines correspond to the mean result for the unpenalized training procedure.
	}
	\label{fig:supplement/mimic/all_fairness/mortality_icu/gender_concept_name}
\end{figure}

\begin{figure}[!htb]
	\centering
	\includegraphics[width=0.9\linewidth]{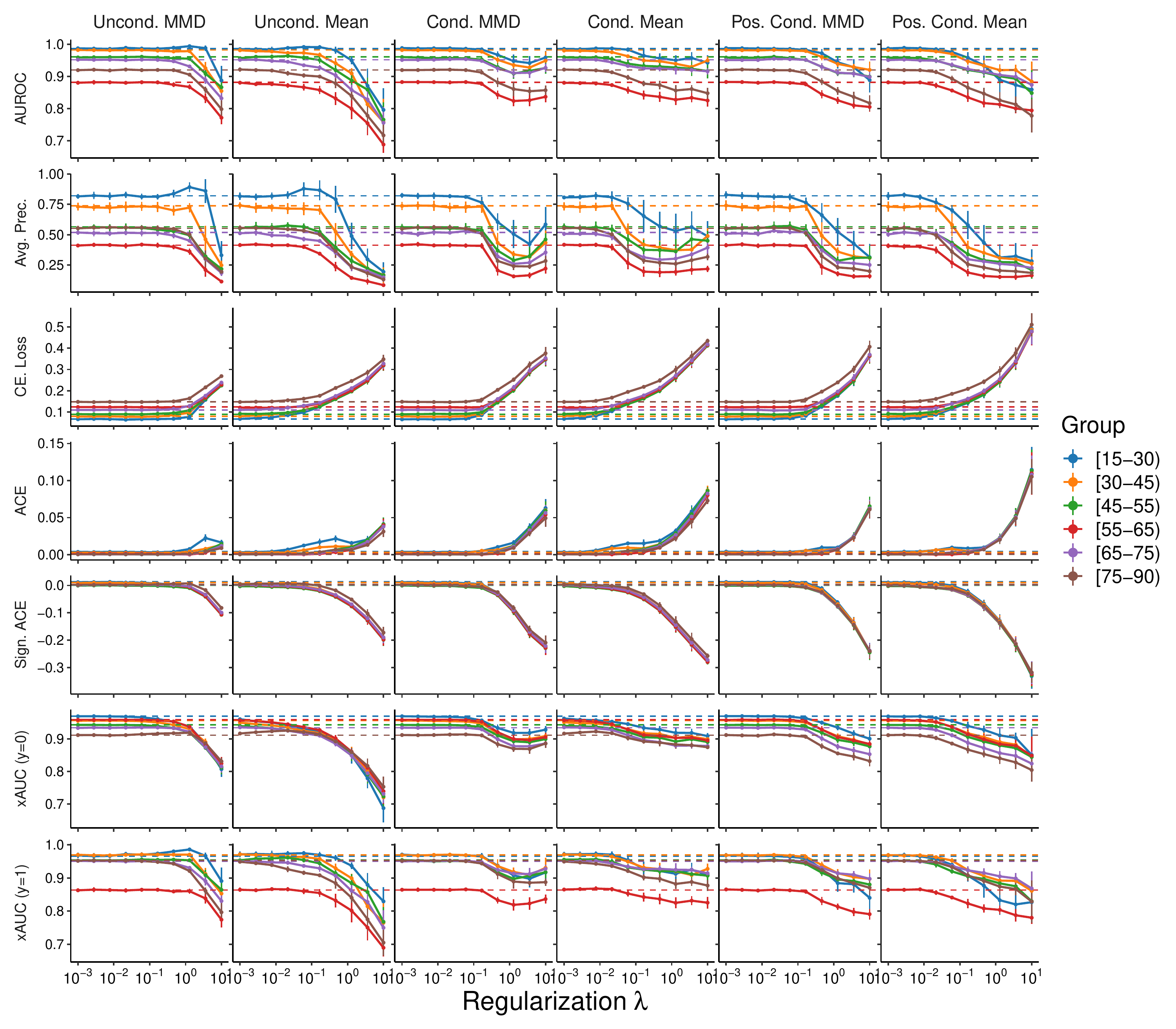}
	\caption{
	    Group-level model performance measures as a function of the extent $\lambda$ that violation of the fairness criterion is penalized when the \textbf{age} group is considered as the sensitive attribute for prediction of \textbf{ICU mortality} in the \textbf{MIMIC-III} database. Results shown are the mean $\pm$ SD for the area under the ROC curve (AUROC), average precision (Avg. Prec), the cross entropy loss (CE Loss), the absolute calibration error (ACE), the signed absolute calibration error (Sign. ACE), and cross group ranking performance (xAUC; $\textrm{xAUC}_k^1$ is indicated by (y=1) and $\textrm{xAUC}_k^0$ by (y=0)) for each group for objectives that penalize violation of threshold-free Demographic Parity (Uncond. MMD and Mean), Equalized Odds (Cond. MMD and Mean), and Equal Opportunity (Pos. Cond. MMD and Mean) with MMD- and mean-based penalties. Dashed lines correspond to the mean result for the unpenalized training procedure.
	}
	\label{fig:supplement/mimic/all_performance/mortality_icu/age_group}
\end{figure}

\begin{figure}[!htb]
	\centering
	\includegraphics[width=0.9\linewidth]{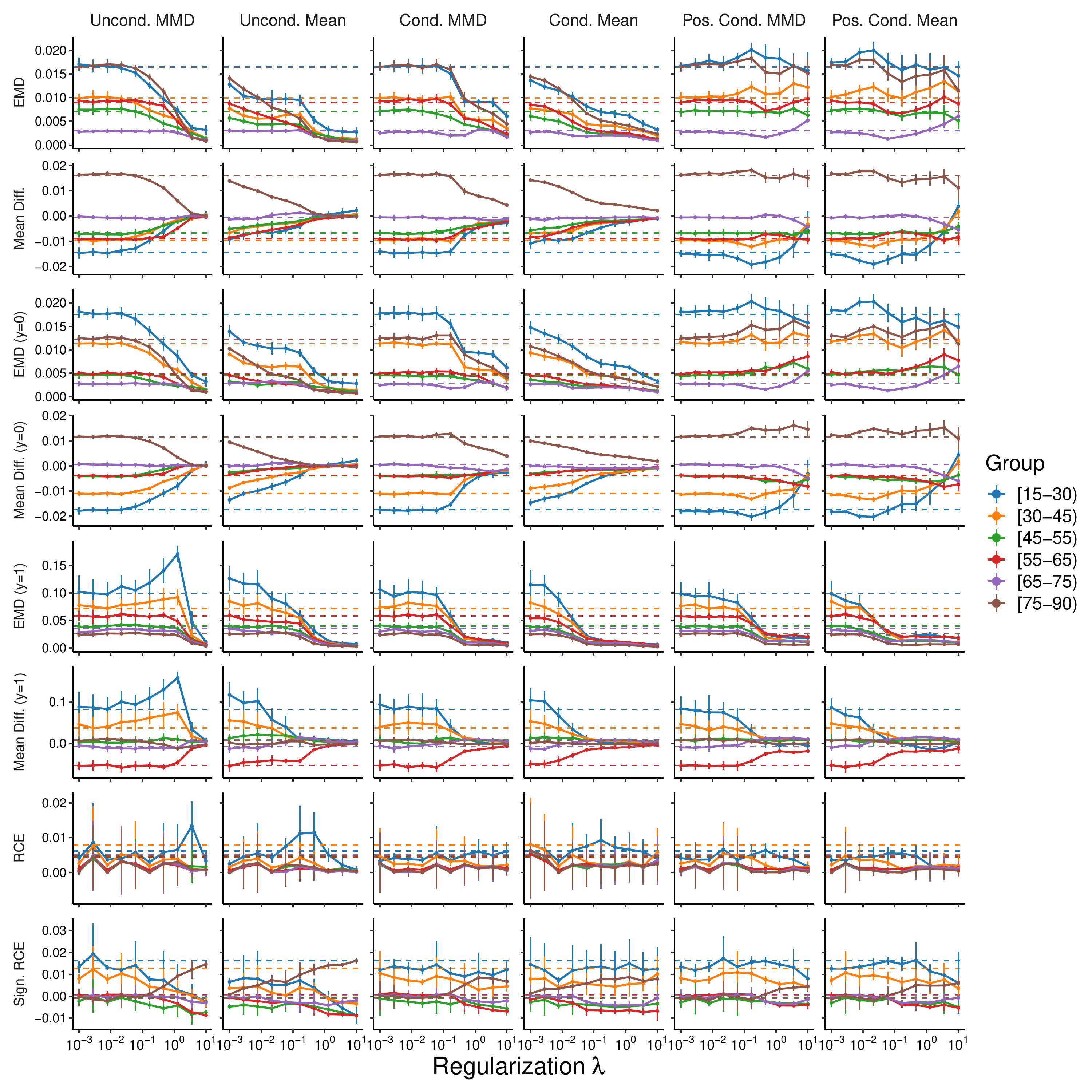}
	\caption{
	    Fairness metrics as a function of the extent $\lambda$ that violation of the fairness criterion is penalized when the \textbf{age} group is considered as the sensitive attribute for prediction of \textbf{ICU mortality} in the \textbf{MIMIC-III} database. Results shown are the mean $\pm$ SD for decomposed group-level metrics that assess conditional prediction parity (EMD and Mean Diff.) and relative calibration error (RCE and Sign. RCE) for objectives that penalize violation of threshold-free Demographic Parity (Uncond. MMD and Mean), Equalized Odds (Cond. MMD and Mean), and Equal Opportunity (Pos. Cond. MMD and Mean) on the basis of MMD- and mean-based penalties.  Measures of conditional prediction parity are separately assessed in the whole population and in the strata for which the outcome is and is not observed (suffixed with (y=1) and (y=0), respectively). 
	    Dashed lines correspond to the mean result for the unpenalized training procedure.
	}
	\label{fig:supplement/mimic/all_fairness/mortality_icu/age_group}
\end{figure}

\end{document}